%% file: main.tex
\documentclass{article} 
\usepackage{iclr2027_conference,times}
\usepackage{koopcell_preprint}

\input{math_commands.tex}

\usepackage{hyperref}
\usepackage{url}
\usepackage{etoc} 

\usepackage[utf8]{inputenc} 
\usepackage[T1]{fontenc}    
\usepackage{hyperref}       
\usepackage{url}            
\usepackage{graphicx}       
\usepackage{wrapfig}        
\usepackage{booktabs}       
\usepackage[table]{xcolor} 
\usepackage{arydshln}       
\usepackage{multirow}       
\usepackage{amsfonts}       
\usepackage{nicefrac}       
\usepackage{microtype}      
\definecolor{bestresultcolor}{HTML}{C6EFCE}
\definecolor{secondresultcolor}{HTML}{DDEBF7}
\definecolor{interpolationlabelcolor}{HTML}{EDE7F6}
\definecolor{extrapolationlabelcolor}{HTML}{FCE8D5}
\definecolor{algpretrain}{HTML}{235B82}
\definecolor{algalternating}{HTML}{237267}
\definecolor{algcomment}{HTML}{56616A}
\newcommand{\bestresult}[1]{\cellcolor{bestresultcolor}$\mathbf{#1}$}
\newcommand{\secondresult}[1]{\cellcolor{secondresultcolor}$#1$}
\usepackage{amsmath}
\usepackage{amssymb}
\usepackage{mathrsfs}
\usepackage{algorithm}
\usepackage{algpseudocode}
\usepackage[normalem]{ulem}
\usepackage{booktabs}

\usepackage{amsthm}
\theoremstyle{plain}
\newtheorem{theorem}{Theorem}[section]
\newtheorem*{theorem*}{Theorem}
\newtheorem{proposition}[theorem]{Proposition}
\newtheorem{lemma}[theorem]{Lemma}

\theoremstyle{definition}

\newtheorem{example}{Example}[section]
\newtheorem{assumption}[theorem]{Assumption}
\theoremstyle{remark}

\newtheorem*{remark*}{Remark}

\title{KoopCell: Koopman-Based Generative Model\\for Learning Single-Cell Dynamics\\from Distribution Snapshots}

\author{%
\begin{minipage}[t]{\dimexpr\textwidth-2\tabcolsep\relax}
\raggedright\normalfont
{\normalsize\bfseries
Wanfeng Lu\textsuperscript{1,2,}\thanks{Equal contribution.}\hspace{0.6em}%
Yutong Zhang\textsuperscript{2,}\footnotemark[1]\hspace{0.6em}%
Keyi Zhou\textsuperscript{2}\hspace{0.6em}%
Chenxin Ge\textsuperscript{2}\hspace{0.6em}%
Wei Lin\textsuperscript{1,2,3,4}\hspace{0.6em}%
Qunxi Zhu\textsuperscript{2,3,4,}\thanks{Corresponding author: \href{mailto:qxzhu@fudan.edu.cn}{\texttt{qxzhu@fudan.edu.cn}}.}%
}\\[4pt]
{\small
\textsuperscript{1}School of Mathematical Sciences, SCMS, SCAM, and CCSB, Fudan University, China\\
\textsuperscript{2}Research Institute of Intelligent Complex Systems, Fudan University, China\\
\textsuperscript{3}State Key Laboratory of Medical Neurobiology and MOE Frontiers Center for Brain Science,\\
\hspace*{0.6em}Institutes of Brain Science, Fudan University, China\\
\textsuperscript{4}Shanghai Artificial Intelligence Laboratory, China
}
\end{minipage}%
}

\hypersetup{
  hidelinks,
  pdftitle={KoopCell: Koopman-Based Generative Model for Learning Single-Cell Dynamics from Distribution Snapshots},
  pdfauthor={Wanfeng Lu, Yutong Zhang, Keyi Zhou, Chenxin Ge, Wei Lin, Qunxi Zhu},
  pdfsubject={Learning single-cell population dynamics from distribution snapshots},
  pdfkeywords={Koopman operator, Mori--Zwanzig, single-cell dynamics, generative modeling}
}
\begin{document}

\maketitle
\etocdepthtag.toc{main}

\begin{abstract}
Learning population dynamics from temporally sparse, unpaired distribution
snapshots is a fundamental challenge in developmental biology.
Recent approaches based on neural differential equations and flow matching can interpolate between observed population snapshots, but may struggle to extrapolate beyond the training 
horizon and often lack an explicit mechanism for modeling developmental branching.
We propose KoopCell, a unified generative framework based on Koopman--Mori--Zwanzig
theory that jointly learns representations and predictive linear latent dynamics.
Theoretically, using the weak continuity equation, we derive a closed-form least-squares estimator for the
Koopman generator from distribution snapshots and establish convergence
guarantees under suitable assumptions. To model branching dynamics, we 
further develop KoopCell-M, which incorporates non-Markovian memory into the 
latent Koopman dynamics through a Markovian embedding.
Experiments on synthetic systems and three scRNA-seq datasets
demonstrate the ability of our framework to recover Koopman spectra,
model branching through memory, and scale to predicting high-dimensional
gene expression distributions, achieving state-of-the-art performance
among the evaluated methods.
\end{abstract}

\section{Introduction}
Inferring population dynamics from temporally sparse, unpaired distribution snapshots is a 
fundamental challenge in machine learning and the natural sciences, with applications in 
biology~\citep{schiebinger2019optimal,moon2019visualizing}, ecology~\citep{ayala1973competition}, 
and epidemiology~\citep{wang2021literature,kosma2023neural}. In particular, advances in high-throughput 
single-cell RNA sequencing (scRNA-seq)~\citep{hwang2018single,chen2019single} have made this problem 
increasingly important in single-cell biology. By profiling population-level gene expression at multiple 
time points, scRNA-seq enables the study of cellular developmental dynamics, the identification of key 
driver genes, and the prediction of future states. However, due to the destructive nature of scRNA-seq measurements, 
only a cross-sectional sample from the population-level gene-expression distribution is available, without cell-level correspondence across time. Inferring the underlying continuous population dynamics from such observations therefore 
remains a substantial computational challenge.

Neural dynamical systems have been widely used to model cellular dynamics from population snapshots. 
\citet{huguet2022manifold,zhang2024scnode} use neural ordinary differential equations (ODEs)~\citep{chen2018neural}, whereas \citet{yeo2021generative,10.1093/bioinformatics/btae400} employ neural stochastic differential equations (SDEs) to capture biological stochasticity in latent space. More recently, \citet{zhang2025learning,wang2026joint,ling2026cellstream} have introduced unbalanced dynamical optimal transport to model cell proliferation and death. Although these methods perform well under their respective dynamical assumptions and can interpolate missing distributions within the observed time horizon, they do not effectively extrapolate beyond the observed time points or explicitly account for branching dynamics. Both capabilities are essential for predicting future cell-population 
states and representing the lineage differentiation that is crucial for understanding cellular development.

In this work, we propose KoopCell, a Koopman-based generative model 
for population dynamics. 
Koopman theory \citep{koopman1931hamiltonian, koopman1932dynamical} provides a unifying perspective 
for modeling nonlinear dynamics by lifting the system state 
into a space of observables (functions of the system 
state), in which the evolution is governed by 
a linear operator.
Neural Koopman methods \citep{nc_deep_koopman}
approximate these observables with an encoder and represent their 
evolution using a finite-dimensional linear operator, but they 
generally rely on paired state transitions or trajectories. 
In contrast, KoopCell jointly learns the observable representation 
and its linear dynamics directly from temporally sparse, unpaired 
distribution snapshots. We theoretically formulate the estimation of the latent 
linear propagator through the weak continuity equation \citep{lu2024weak} and integrate 
it with a generative VAE model in an alternating 
learning procedure, enabling population prediction both within and 
beyond the observed time horizon. A single Markovian linear flow, 
however, may be insufficient to represent branching processes such 
as the differentiation of stem cells into multiple specialized cell 
types. We therefore extend KoopCell using the Mori--Zwanzig 
formalism, which accounts for the influence of unresolved variables 
through non-Markovian memory \citep{mori1965transport,zwanzig1973nonlinear}. By embedding the suggested memory 
dynamics into an augmented latent space \citep{Hijon2010, iLED}, we develop tractable 
variational inference for the hidden memory and a corresponding alternating learning algorithm for the 
memory-augmented model, KoopCell-M. This memory mechanism enables KoopCell-M to model the divergence of a population into distinct developmental branches.
Our main contributions are summarized as follows:
\begin{itemize}
  \item We develop a theoretical and computational framework for identifying a finite-dimensional Koopman representation from unpaired distribution snapshots. Based on the weak continuity equation, we characterize the closed-form estimator, establish its convergence properties, and jointly learn the associated observable representation within a generative model.
  \item Furthermore, we introduce KoopCell-M, a Mori--Zwanzig-inspired extension that represents non-Markovian memory in an augmented latent space, providing a mechanistic formulation and a practical learning algorithm for population dynamics with branching.
  \item We evaluate KoopCell and KoopCell-M on synthetic toy systems and three high-dimensional scRNA-seq datasets. The experiments validate Koopman spectral recovery and the ability to model branching dynamics, while demonstrating state-of-the-art extrapolation performance in predicting cellular population distributions from sparse, unpaired measurements.
\end{itemize}

\section{Preliminaries}
\label{sec:preliminaries}

\textbf{Problem Formulation and Notation.}
We consider a time series of data distributions $\rho_t$
($t\in\mathcal T\subset\mathbb R^+$), from each of which we observe
$n_t$ independent samples (e.g., cells)
$\{\bm{x}_{t,i}\}_{i=1}^{n_t}$, with
$\bm{x}_{t,i}\sim\rho_t$ and
$\bm{x}_{t,i}\in\mathcal X=\mathbb R^G$.
We aim to learn the population dynamics that explain the evolution of
these distribution snapshots. A typical application is to recover
cellular developmental dynamics from time-series scRNA-seq data, where
cells sampled from $\rho_t$ form a gene expression matrix
$\bm{X}_t\in\mathbb R^{n_t\times G}$, and $G$ is the number of genes.
Given the high dimensionality of the observations, a common approach is
to model temporal dynamics in a $D$-dimensional latent space
$\mathcal Z=\mathbb R^D$, with $D\ll G$.
Rather than specifying a heuristic dynamical structure in the latent space,
we use Koopman theory to guide the learning of latent coordinates with approximately linear evolution.
Throughout, we use uppercase
state symbols for random variables and lowercase symbols for their realizations.

\textbf{Koopman Theory and Koopman Learning.}
For an autonomous system $\dot{\bm{x}}=\bm{f}(\bm{x})$ with flow $F_t$, an observable
is a function $g:\mathcal X\to\mathbb R$ of the system state.
\textcolor{black}{For non-autonomous systems, augmenting the state with time ($\dot t=1$) yields an autonomous formulation.}
The Koopman operator evolves observables by
$[U(t)g](\bm{x})=g(F_t \bm{x})$ and is linear even when the state-space dynamics are
nonlinear~\citep{koopman1931hamiltonian,koopman1932dynamical}.
We assume that the operators $U(t):\mathcal H\to\mathcal H$ form a
strongly continuous semigroup on a Hilbert space $\mathcal H$ of
observables, with generator $L$ and domain $D(L)$.
For sufficiently smooth $g\in D(L)$, $Lg=\bm{f}\cdot\nabla g$ and
$\frac{\mathrm{d}}{\mathrm{d}t}U(t)g=U(t)Lg$.
Let $\bm{X}_t=F_t\bm{X}_0$ with $\bm{X}_0\sim\rho_0$.
If $\mathcal H_D=\operatorname{span}\{g_1,\ldots,g_D\}\subset D(L)$ is
\textit{invariant} under $L$, then $L\bm{g}=\bm{A}\bm{g}$ for a matrix $\bm{A}$, so
$\bm{Z}_t=\bm{g}(\bm{X}_t)$ satisfies $\dot{\bm{Z}}_t=\bm{A}\bm{Z}_t$, where
$\bm{g}=(g_1,\ldots,g_D)^\top$.
With the observable map $\bm{g}$ fixed, extended dynamic mode decomposition
(EDMD)~\citep{edmd} estimates a propagator from paired states
$(\bm{x}_i,\bm{y}_i)$, $\bm{y}_i=F_\Delta \bm{x}_i$, by solving a regression problem
$\min_{\bm{K}}\sum_i\|\bm{g}(\bm{y}_i)-\bm{K}\bm{g}(\bm{x}_i)\|_2^2$.
Koopman representation learning extends EDMD by learning the observable
map together with the latent dynamics. Specifically, an encoder
$\bm{x}\mapsto \bm{z}=\bm{g}(\bm{x})\in\mathbb R^D$ is jointly trained with a linear
latent propagator and a decoder to obtain coordinates that support both
approximately linear evolution and state
reconstruction~\citep{otto2019linearly,nc_deep_koopman}.
Our work extends this framework to learning generative latent linear
dynamics from temporally sparse, unpaired distribution snapshots.

\textbf{Mori--Zwanzig Formalism and Memory Embedding.}
The reduced finite-dimensional linear dynamics above are exact when
the span of the chosen observables is invariant under the Koopman generator $L$.
When this linear closure fails, the evolution of the resolved coordinates
$\bm{Z}_t=\bm{g}(\bm{X}_t)\in\mathbb R^D$ generally involves unresolved observables
outside the chosen observable subspace $\mathcal H_D$.
The Mori--Zwanzig formalism, originally developed
in statistical mechanics, accounts for their influence through memory
and forcing terms~\citep{mori1965transport,zwanzig1973nonlinear}.
Specifically, under suitable regularity, Mori's linear projection onto
$\mathcal H_D$ yields a generalized Langevin equation~\citep{lin2021datamori}:
\[
 \dot{\bm{Z}}_t=\bm{A}\bm{Z}_t+\int_0^t \bm{K}(t-s)\bm{Z}_s\,\mathrm{d}s+\bm{\eta}_t.
\]
Here $\bm{A}$ describes the instantaneous linear contribution, $\bm{K}$ is the
memory kernel, and $\bm{\eta}_t$ carries the influence of unresolved initial
conditions.
A tractable closure approximates the kernel by
$\bm{K}(u)\approx \bm{B}e^{\bm{\Lambda} u}\bm{C}$ and represents the unresolved influence
through auxiliary hidden memory variables $\bm{H}_t\in\mathbb R^{d_h}$
\citep{lei2016parameterization,iLED}.
This leads to the augmented dynamics
$\mathrm{d}\bm{Z}_t=(\bm{A}\bm{Z}_t+\bm{B}\bm{H}_t)\,\mathrm{d}t$ and
$\mathrm{d}\bm{H}_t=(\bm{C}\bm{Z}_t+\bm{\Lambda} \bm{H}_t)\,\mathrm{d}t+\bm{\Sigma}_h\,\mathrm{d}\bm{W}_t$,
where $\bm{W}_t$ is standard Brownian motion and $\bm{\Sigma}_h$ controls the
hidden noise. Eliminating $\bm{H}_t$ yields the kernel $\bm{K}(u)=\bm{B}e^{\bm{\Lambda} u}\bm{C}$
and a temporally correlated forcing determined by $\bm{H}_0$ and the hidden
noise. The augmented state $(\bm{Z}_t,\bm{H}_t)$ thus evolves as a Markovian system,
while $\bm{Z}_t$ alone generally retains memory.
This Markovian embedding provides the underlying mechanism for modeling
branching dynamics in our framework.

\textbf{Wasserstein Flow and Optimal Transport.}
Time series of distribution snapshots are often modeled as probability
flows in Wasserstein space.
We write $\mathcal P_2(\mathbb R^D)$ for the space of probability measures with finite
second moment and $W_2$ for the Wasserstein-2 distance, defined by
$W_2^2(\mu,\nu)=\inf_{\pi\in\Pi(\mu,\nu)}
\int\|\bm{z}-\bm{y}\|^2\mathrm{d}\pi(\bm{z},\bm{y})$, where $\Pi(\mu,\nu)$ is the set of couplings.
We use $\mu[\psi]=\int\psi(\bm{z})\,\mathrm{d}\mu(\bm{z})$ and write $F_\#\mu$ for the
distribution of $F(\bm{Z})$ when $\bm{Z}\sim\mu$.
The Wasserstein flow $t\mapsto\mu_t$, with
$\mu_\cdot\in AC^2([0,T];\mathcal P_2(\mathbb R^D))$, is an absolutely
continuous curve in Wasserstein space that describes the evolution of
probability measures over time.
For almost every $t$, there exists a unique canonical velocity
$\bm{v}_t\in T_{\mu_t}\mathcal P_2(\mathbb R^D)$\footnote{Here,
$T_\mu\mathcal P_2=\overline{\{\nabla\varphi:
\varphi\in C_c^\infty(\mathbb R^D)\}}^{L^2(\mu)}$ is the tangent space; the overline denotes closure in $L^2(\mu)$.}
satisfying $\partial_t\mu_t+\nabla\cdot(\bm{v}_t\mu_t)=0$ in the
distributional sense~\citep{ambrosioGradientFlowsMetric2008}.
Among all velocity fields generating the same measure curve, $\bm{v}_t$ has
minimum $L^2(\mu_t)$ norm
$\bigl(\int\|\bm{v}_t\|^2\mathrm{d}\mu_t\bigr)^{1/2}$.
We denote by $P_\mu:L^2(\mu;\mathbb R^D)\to
T_\mu\mathcal P_2(\mathbb R^D)$ the $L^2(\mu)$-orthogonal projection
onto this tangent space.
For a zero-mass distribution $\sigma$, we define the negative Sobolev norm
$\|\sigma\|_{-1,\mu}
=\sup\limits_{\int\|\nabla\varphi\|^2\mathrm{d}\mu\le1}
\langle\sigma,\varphi\rangle$.
Here $\langle\sigma,\varphi\rangle$ denotes the duality pairing of a
distribution with a test function.
Equivalently, when finite, $\|\sigma\|_{-1,\mu}$ is the minimum
$L^2(\mu)$ norm of a tangent field $\bm{w}$ satisfying
$\sigma=-\nabla\cdot(\bm{w}\mu)$; see Lemma~\ref{lem:min-flux-v6}
in Appendix~\ref{app:wc-geometry-v6}.

\section{Methods}
\subsection{Koopman-based VAE for Generative Modeling of Population Dynamics}
\label{sec:linear-comparison}

Motivated by Koopman theory, we jointly learn an encoder that maps cellular
states to observables with approximately linear evolution, a linear
propagator governing their dynamics, and a decoder reconstructing the
cellular states from latent representations. Specifically, we use a VAE~\citep{Kingma2014},
widely used for single-cell gene expression
modeling~\citep{gronbech2020scvae},
with encoder $q_{\bm{\theta}}(\cdot\mid \bm{x},t)=\mathcal N(\bm{m}_{\bm{\theta}}(\bm{x},t),\bm{S}_{\bm{\theta}}(\bm{x},t))$
and decoder $\operatorname{Dec}_{\bm{\phi}}$, where $\bm{z}\in\mathbb R^D$.
\textcolor{black}{We concatenate $t$ with $\bm{x}$ to encode the augmented state.}
Including the constant observable $1$ yields affine dynamics
$\dot{\bm{Z}}_t=\bm{A}\bm{Z}_t+\bm{b}_z$. For simplicity, we use $\dot{\bm{Z}}_t=\bm{A}\bm{Z}_t$ below and
describe the detailed implementation in Appendix~\ref{app:linear-method}.

Unlike traditional settings where the Koopman model is
learned from densely sampled, paired
trajectories, we are given only temporally sparse, unpaired distribution
snapshots. We address the challenge of
\dotuline{learning the Koopman operator from snapshots}
through the following two alternating steps:
(1) with the observables (latent coordinates) fixed, we estimate the linear
matrix that best fits the evolution of the latent distributions,
providing an analogue of EDMD estimation for population snapshots;
(2) with the estimated linear propagator fixed, we refine the latent
representation to better fit the dynamics. By alternating between propagator estimation and representation refinement,
we ultimately learn the Koopman representation and its linear dynamics generator simultaneously.

\textbf{Estimation of the linear propagator.}
We fix a set of observables $\{g_1,\ldots,g_D\}$, which induces a
Wasserstein flow of latent distributions
$\mu_t=\bm{g}_\#\rho_t\in\mathcal P_2(\mathbb R^D)$, $t\in[0,T]$.
We aim to find a linear generator $\bm{A}$ such that
$(e^{\bm{A}\Delta t})_\#\mu_t\approx\mu_{t+\Delta t}$.
Exact propagation under this linear flow satisfies the continuity equation
$\partial_t\mu_t+\nabla_{\bm{z}}\cdot(\bm{A}\bm{z}\,\mu_t)=0$.
Accordingly, we introduce the \textit{weak population residual}
$r_{\bm{A}}(t)=\partial_t\mu_t+\nabla_{\bm{z}}\cdot(\bm{A}\bm{z}\,\mu_t)$ to quantify the
mismatch between the linear dynamics and the evolution of $\mu_t$.
We then seek an $\bm{A}$ minimizing $\mathcal J_\mu(\bm{A})=\frac12\int_0^T
\|r_{\bm{A}}(t)\|_{-1,\mu_t}^2\mathrm{d}t$.
However, directly minimizing the residual is intractable, as it requires the density derivative $\partial_t \mu_t$.
To this end, we use the equivalent min--max formulation in the dual test
space, as established in Theorem~\ref{thm:dual-geometry-v6}
(Appendix~\ref{app:wc-geometry-v6}):
\begin{flalign}\label{eq:linear-comparison-residual}
    &\begin{aligned}
    & \bm{A}^* = \operatorname{arg}\min_{\bm{A}} \mathcal{J}_{\mu}(\bm{A}) = \operatorname{arg}\min_{\bm{A}} \sup_{\varphi\in\mathbb V_D}
  \left\{
    \mathcal R_D(\bm{A};\varphi)
    -\frac12\|\varphi\|_{\mathbb V_D}^2
  \right\}, \\
  \text{where } {}\quad & \mathcal R_D(\bm{A};\varphi)
  ={}\mu_T[\varphi(T,\cdot)]-\mu_0[\varphi(0,\cdot)]-\int_0^T\mu_t\!\left[
    \partial_t\varphi(t,\bm{z})
    +\nabla_{\bm{z}}\varphi(t,\bm{z})^\top \bm{A}\bm{z}
  \right]\mathrm{d}t.
    \end{aligned} &&
\end{flalign}
Here $\mathbb V_D=L^2(0,T;\dot H^1(\mu_t))$ is the Hilbert space with
$\|\varphi\|_{\mathbb V_D}^2=\int_0^T\mu_t[\|\nabla_{\bm{z}}\varphi\|^2]\mathrm{d}t$.
{The following theorem shows that minimizing $\mathcal J_\mu$
gives the best approximation of the canonical Wasserstein velocity $\bm{v}_t$ by
projected linear fields in $L^2(\mathrm{d}t\,\mathrm{d}\mu_t)$.
Furthermore, restricting the test space $\mathbb V_D$ to finitely many functions yields a
least-squares problem which admits a closed-form solution.}

\begin{theorem}
\label{thm:main-finite-test}
Assume $\mu_\cdot\in AC^2([0,T];\mathcal P_2(\mathbb R^D))$, and let $\bm{v}_t$
be its canonical velocity. Then, for every $\bm{A}\in\mathbb R^{D\times D}$,
\begin{equation}
 \mathcal J_\mu(\bm{A})
 =\sup_{\varphi\in\mathbb V_D}
   \left\{\mathcal R_D(\bm{A};\varphi)-\tfrac12\|\varphi\|_{\mathbb V_D}^2\right\}
 =\frac12\int_0^T\|\bm{v}_t-P_{\mu_t}(\bm{A}\bm{z})\|_{L^2(\mu_t)}^2\mathrm{d}t.
 \label{eq:linear-comparison-geometry}
\end{equation}
Hence the minimization gives the orthogonal best approximation of the
canonical velocity by projected linear velocity fields.
Furthermore, restrict the supremum to
$\mathbb V_{D,m}=\operatorname{span}\{\varphi_1,\ldots,\varphi_m\}$ and define
\begin{equation}
 \begin{gathered}
 y_j=\mu_T[\varphi_j(T)]-\mu_0[\varphi_j(0)]
       -\int_0^T\mu_t[\partial_t\varphi_j(t)]\mathrm{d}t,\\
 \bm{H}_j=\int_0^T\mu_t[\nabla_{\bm{z}}\varphi_j(t,\bm{z})\bm{z}^\top]\mathrm{d}t,\qquad
 M_{ij}=\int_0^T\mu_t[\nabla_{\bm{z}}\varphi_i(t,\bm{z})^\top
                            \nabla_{\bm{z}}\varphi_j(t,\bm{z})]\mathrm{d}t.
 \end{gathered}
 \label{eq:linear-comparison-finite-statistics}
\end{equation}
Let $\bm{a}=\operatorname{vec}(\bm{A})$, let row $j$ of $\bm{G}$ be
$\operatorname{vec}(\bm{H}_j)^\top$, and let $\bm{y}=(y_1,\ldots,y_m)^\top$.
If $\bm{M}\succ0$, restricting \Eqref{eq:linear-comparison-residual} to
$\mathbb V_{D,m}$ gives the weighted least-squares objective
$\mathcal J_{D,m}(\bm{A})=\frac12(\bm{y}-\bm{G}\bm{a})^\top \bm{M}^{-1}(\bm{y}-\bm{G}\bm{a})$.
If $\bm{G}$ has full column rank, its unique minimizer is
$\operatorname{vec}(\widehat{\bm{A}}_{D,m})
=(\bm{G}^\top \bm{M}^{-1}\bm{G})^{-1}\bm{G}^\top \bm{M}^{-1}\bm{y}$.
Finally, under the test-enrichment and uniqueness assumptions in
Appendix~\ref{app:finite-tests-v6}, these minimizers converge to $\bm{A}^*$, the unique
population minimizer of \Eqref{eq:linear-comparison-residual} as $m\to\infty$.
\end{theorem}

We provide the proof of Theorem~\ref{thm:main-finite-test} in
Appendices~\ref{app:wc-geometry-v6} and~\ref{app:finite-tests-v6}.
In practice, we use random Fourier tests to construct $\mathbb V_{D,m}$,
reducing the min--max optimization to a weighted least-squares problem.
For irregular snapshots at $0=t_0<t_1<\cdots<t_N=T$, we formulate weak
equations on each adjacent interval $I_k=[t_k,t_{k+1}]$ and approximate
the integrals by the trapezoidal rule,
$\int_{I_k}f(t)\mathrm{d}t\simeq\frac{t_{k+1}-t_k}{2}
\{f(t_k)+f(t_{k+1})\}$.
The integrands $f(\cdot) = \mu_\cdot [\cdot]$ are expectations under the latent distributions, approximated 
by empirical averages over the encoded snapshots. Finally, equations
over different intervals are combined to form the overall regression
problem, with ridge regularization used to stabilize the solve.
Implementation details are provided in Appendix~\ref{app:linear-weak-regression}.
We further show in a Koopman-invariant RKHS that the minimum weak population
residual vanishes as $D\to\infty$ (Appendix~\ref{sec:weak-residual-convergence}),
derive Wasserstein prediction bounds (Appendix~\ref{sec:weak-residual-prediction}),
and prove locally uniform strong semigroup convergence
(Appendix~\ref{sec:strong-semigroup-convergence}), under the stated
approximation, stability, and identifiability assumptions.
We also discuss theoretical limitations in Appendix~\ref{sec:strong-semigroup-convergence}.

\textbf{Updating the representation under the linear flow.}
Given the estimated propagator $\bm{A}=\widehat{\bm{A}}_{D,m}$, we update the
representation to better fit the linear dynamics. To regularize the VAE
representation, we use a prior transported by the linear flow: the initial
prior $p_0=\mathcal N(\bm{0},\bm{I}_D)$ evolves into
$p_t=\mathcal N(\bm{0},e^{\bm{A}t}e^{\bm{A}^\top t})$.
We thus combine squared reconstruction error with the KL divergence~\citep{csiszar1975divergence}
between the encoder posterior and this dynamic prior:
\begin{equation}
 \mathcal L_{\mathrm{VAE}}
 =\mathbb E_{t,\bm{x}}\left[
 \frac{1}{{2\sigma_x^2d_x}}{\mathbb E_{\bm{Z}\sim q_{\bm{\theta}}(\cdot\mid \bm{x},t)}
          \|\bm{x}-\operatorname{Dec}_{\bm{\phi}}(\bm{Z})\|^2}
 +\beta\operatorname{KL}\bigl(q_{\bm{\theta}}(\cdot\mid \bm{x},t)\,\|\,p_t\bigr)\right].
 \label{eq:linear-comparison-vae}
\end{equation}
Here $d_x$ is the expression dimension, $\sigma_x^2$ is the fixed
observation variance, and $\mathbb E_{t,\bm{x}}$ averages over training times
and cells.
To encourage accurate population prediction in both latent and expression space,
we augment $\mathcal L_{\mathrm{VAE}}$
with the population-matching loss
$\mathcal L_{\mathrm{pop}}
=\mathbb E_{s<t\in\mathcal T}[
\lambda_z\mathcal S_\varepsilon(\mu_t,\widehat\mu_{s\to t})
+\lambda_x\mathcal S_\varepsilon(\rho_t,\widehat\rho_{s\to t})]$,
where $\mu_t=\int q_{\bm{\theta}}(\cdot\mid \bm{x},t)\mathrm{d}\rho_t(\bm{x})$ is the encoder-induced
population, $\widehat\mu_{s\to t}=(e^{\bm{A}(t-s)})_\#\mu_s$ is the latent distribution
propagated from $\mu_s$ to time $t$, and
$\widehat\rho_{s\to t}=(\operatorname{Dec}_{\bm{\phi}})_\#\widehat\mu_{s\to t}$
is the decoded prediction.
The expectation averages over all forward pairs of training times, and
$\mathcal S_\varepsilon$ is the debiased Sinkhorn divergence with quadratic
cost.
We further regularize the encoder with the weak population residual loss
$\mathcal L_{\mathrm{weak}}=\mathcal J_{D,m}(\widehat{\bm{A}}_{D,m})$,
which penalizes violations of the weak continuity equation and encourages
latent coordinates in which population evolution is better approximated
by the estimated linear dynamics
(Appendix~\ref{app:linear-weak-regression}).
The linear flow is evaluated by matrix exponentials, enabling prediction
at arbitrary times without costly numerical ODE integration.

\textbf{Training.}
Following~\citet{zhang2024scnode}, we pretrain the VAE with a standard-normal Gaussian
prior, then alternate between generator estimation and representation
learning. During each representation update, we hold $\bm{A}$ fixed and minimize
$\mathcal L_{\mathrm{KC}}=\mathcal L_{\mathrm{VAE}}
+\mathcal L_{\mathrm{pop}}+\lambda_{\mathrm{weak}}\mathcal L_{\mathrm{weak}}$
over the VAE parameters $\bm{\theta},\bm{\phi}$.
Algorithm~\ref{alg:linear-training} summarizes the training procedure of KoopCell.

\subsection{KoopCell-M: Memory-Augmented Koopman Dynamics for Branching}
\label{subsec:memory}

Learning a linear latent flow is appealing, but it constrains the
modeling of branching during cellular differentiation, in which an
initially homogeneous population differentiates into multiple cell types.
Specifically, an ODE with a Lipschitz vector field cannot exactly transport
a source distribution supported on $M$ disjoint, compact, path-connected
components to a target with $N\ne M$ such components~\citep[Proposition~3.3]{zhao2026delay}.
Thus, when the initial population has connected latent support,
the linear dynamics followed by a Lipschitz decoder
cannot produce fate populations with disconnected support.
Moreover, the linear flow itself cannot create new latent
density modes.
Although nonlinear decoding can introduce additional modes,
bounding the decoder Lipschitz constant limits the amplification
of latent separation
(Appendix~\ref{app:memory-branching}).
To model branching through an explicit dynamical mechanism,
\citet{zhao2026delay} introduce non-Markovian delay dynamics with distinct
initial histories for the same state, allowing trajectories to diverge.
Such non-Markovian dynamics also arise within Koopman theory through the
Mori--Zwanzig formalism (Section~\ref{sec:preliminaries}), which describes
how unresolved observables contribute memory to the resolved linear dynamics.
In cellular differentiation, such unresolved influences may reflect
heritable fate biases that are not fully captured by
scRNA-seq~\citep{weinreb2020lineage}.
This connection motivates us to introduce memory into KoopCell, using
non-Markovian dynamics to model branching within the same Koopman framework.

\textbf{Memory embedding and branching.}
To obtain a tractable formulation of this non-Markovian model, we use the Markovian embedding
from Section~\ref{sec:preliminaries} to evolve the augmented latent state $(\bm{Z}_t,\bm{H}_t)$:
\begin{equation}
  \mathrm{d}\bm{Z}_t=(\bm{A}\bm{Z}_t+\bm{B}\bm{H}_t+\bm{b}_z)\,\mathrm{d}t,\qquad
  \mathrm{d}\bm{H}_t=(\bm{C}\bm{Z}_t+\bm{\Lambda} \bm{H}_t + \bm{b}_h)\,\mathrm{d}t+\bm{\Sigma}_h\,\mathrm{d}\bm{W}_t.
  \label{eq:memory-embedding}
\end{equation}
Here $\bm{Z}_t\in\mathbb R^D$ is the \emph{visible latent state} encoded
from cellular gene expression,
and $\bm{H}_t\in\mathbb R^{d_h}$ contains auxiliary memory variables.
A joint Gaussian initialization of $(\bm{Z}_0,\bm{H}_0)$ yields Gaussian
visible $\bm{Z}$-marginals. To enable branching in the latent distribution,
we parameterize hidden initial distributions by a discrete variable $R\in\mathscr R$.
Each value of $R$ indexes a candidate root-to-terminal path in a directed
acyclic graph (DAG) of coarse biological states.
We detail the state annotations and graph construction in
Appendix~\ref{app:memory-dynamics} and show the resulting DAG for the mouse
cell reprogramming dataset (SC; \citealp{schiebinger2019optimal}) in
Figure~\ref{fig:sc-candidate-dag}.
With $R$ specifying the developmental path, we define a generative model
with latent variables $(R,\bm{Z}_t,\bm{H}_t)$.
We specify the initial distribution of the augmented dynamics in
\Eqref{eq:memory-embedding} as
\begin{equation}
 R\sim\operatorname{Cat}(\bm{\pi}),\qquad
 \bm{Z}_0\mid R=r\sim\mathcal N(\bm{0},\bm{I}_D),\qquad
 \bm{H}_0\mid \bm{Z}_0=\bm{z},R=r\sim\mathcal N(\bm{F}_r\bm{z}+\bm{a}_r,\bm{S}_r).
 \label{eq:memory-shared-initial}
\end{equation}
Under this parametrization, the augmented process is Gaussian conditional on $R$.
The marginal distribution of $\bm{Z}_t$ is therefore a Gaussian mixture,
$p_t(\bm{z})=\sum_{r\in\mathscr R}\pi_r
\mathcal N(\bm{z};\bm{m}^z_{t,r},\bm{V}^{zz}_{t,r})$, with propagated means
$\bm{m}^z_{t,r}$ and covariances $\bm{V}^{zz}_{t,r}$
(Appendix~\ref{app:memory-gaussian-inference}).
Although all paths share the same visible initial distribution $p_0$, coupling
through $\bm{B}\bm{H}_t$ can turn hidden heterogeneity into visible separation,
producing multiple density modes when component means become sufficiently
separated relative to their spread.
We construct a toy example in Section~\ref{subsec:toy-branching}
to illustrate how introducing memory enables the dynamics to
generate distinct branches from an initially unimodal population
(Figure~\ref{fig:branching-rollouts}).

\textbf{Learning the representation and memory dynamics.}
As in KoopCell, we first pretrain the VAE, then alternate between dynamics
estimation and representation learning.
With the representation and memory statistics fixed, we estimate $\bm{A},\bm{b}_z$
in closed form using the same least-squares formulation, after
subtracting the memory contribution from the weak continuity equations
(Appendix~\ref{app:memory-weak-regression}).

With $\bm{A},\bm{b}_z$ fixed, we jointly update the representation and remaining
memory parameters using the distribution-matching objective $\mathcal L_{\mathrm{pop}}$ over all
training pairs $s<t$, as in KoopCell.
Prediction from a snapshot at $s$ additionally requires inferring $R$ and $\bm{H}_s$.
Our generative model supports tractable variational inference: given
$\bm{z}\sim q_{\bm{\theta}}(\cdot\mid \bm{x},s)$, we compute the path probabilities
$q_s(r\mid \bm{z},\bm{x})$ and Gaussian memory conditionals
$p(\bm{H}_s\mid \bm{Z}_s=\bm{z},R=r)$ analytically
(Appendix~\ref{app:memory-gaussian-inference}).
We propagate the inferred augmented population under \Eqref{eq:memory-embedding},
with transition means and covariances evaluated by matrix exponentials.
The predicted $\bm{Z}_t$-marginal is $\widehat\mu_{s\to t}$, and decoding gives
$\widehat\rho_{s\to t}=(\operatorname{Dec}_{\bm{\phi}})_\#\widehat\mu_{s\to t}$.
To encourage rollouts to follow the biological pathway specified by $R$,
we introduce a state classifier $c_{\bm{\omega}}(c\mid \bm{z})$.
We jointly train the VAE and classifier using the snapshot objective
$\mathcal L_{\mathrm{snap}}$, which combines reconstruction, KL
regularization, and supervised classification on annotated training cells
(Appendix~\ref{app:memory-dynamics}).
We then evaluate $c_{\bm{\omega}}$ along each rollout and use the predicted state
probabilities to compute the path consistency loss $\mathcal L_{\mathrm{path}}$,
which encourages progression through biological states in the order specified by $R$
(\Eqref{eq:memory-path-loss} in Appendix~\ref{app:memory-objectives}).
The overall training objective is
$\mathcal L_{\mathrm{snap}}+\mathcal L_{\mathrm{pop}}
+\lambda_R\mathcal L_{\mathrm{path}}+\lambda_{\mathrm{weak}}\mathcal L_{\mathrm{weak}}$.
Appendix~\ref{app:memory-objectives} derives these objectives, and
Algorithm~\ref{alg:memory-training} summarizes the alternating learning procedure.

\section{Experiments}
We evaluate our models on synthetic systems and three
single-cell RNA-sequencing datasets. Through these experiments, we answer the following three questions:
\textbf{Q1.} Can KoopCell recover the underlying Koopman spectrum from
sparse, unpaired snapshots, and does this recovery support stable extrapolation?
\textbf{Q2.} How does incorporating memory improve the modeling of
branching dynamics?
\textbf{Q3.} Can our models scale to high-dimensional scRNA-seq data and
learn from sparse, irregular snapshots to accurately predict gene expression
distributions within and beyond the training horizon?

\begin{table*}[t]
\centering
\caption{Hard-task Wasserstein-2 distance ($\downarrow$) at held-out time points
(mean $\pm$ sample standard deviation over three training seeds). Shaded labels
distinguish interpolation from extrapolation within each dataset.
The best and second-best means are shaded green (bold) and blue, respectively.}
\label{tab:wasserstein-hard-tasks}
\scriptsize
\setlength{\tabcolsep}{1.4pt}
\renewcommand{\arraystretch}{1.08}
\resizebox{\textwidth}{!}{%
\begin{tabular}{*{11}{c}}
\toprule
Prediction & Data & $t$ & scNODE & LGP-OT & MIOFlow & PRESCIENT & PI-SDE & VGFM & KoopCell & KoopCell-M \\
\midrule
\cellcolor{interpolationlabelcolor} & ZB & 3 & $577.71\pm3.69$ & $577.91\pm0.64$ & $594.71\pm32.03$ & $597.49\pm1.76$ & \bestresult{567.30\pm1.41} & $575.86\pm3.22$ & \secondresult{572.32\pm2.14} & $588.91\pm6.04$ \\
\cellcolor{interpolationlabelcolor} & ZB & 5 & $511.58\pm5.63$ & $504.45\pm1.73$ & $519.32\pm7.48$ & $540.72\pm1.52$ & \secondresult{492.43\pm2.19} & $499.66\pm0.81$ & $497.80\pm3.83$ & \bestresult{488.84\pm9.27} \\
\cellcolor{interpolationlabelcolor} & ZB & 7 & $440.65\pm6.47$ & \secondresult{411.20\pm3.07} & $452.30\pm0.77$ & $476.68\pm1.16$ & $416.80\pm3.36$ & $434.99\pm1.37$ & $412.99\pm1.50$ & \bestresult{408.79\pm1.57} \\
\cellcolor{interpolationlabelcolor}\multirow{-4}{*}{Interpolation} & ZB & 9 & $514.24\pm5.97$ & $475.65\pm2.60$ & $539.92\pm11.44$ & $527.09\pm1.25$ & \secondresult{474.41\pm3.51} & $522.60\pm2.83$ & $474.92\pm1.55$ & \bestresult{468.53\pm2.72} \\
\noalign{\vskip 1.5pt}
\cdashline{1-11}
\noalign{\vskip 1.5pt}
\cellcolor{extrapolationlabelcolor} & ZB & 11 & $653.56\pm8.98$ & $605.14\pm1.78$ & $654.64\pm2.17$ & $656.41\pm4.17$ & $612.74\pm3.51$ & $635.96\pm2.73$ & \secondresult{595.82\pm0.48} & \bestresult{588.73\pm0.77} \\
\cellcolor{extrapolationlabelcolor}\multirow{-2}{*}{Extrapolation} & ZB & 12 & $719.28\pm12.36$ & $665.78\pm4.77$ & $720.67\pm17.25$ & $700.68\pm4.58$ & $679.61\pm4.11$ & $720.54\pm10.97$ & \secondresult{648.91\pm2.47} & \bestresult{639.78\pm0.72} \\
\midrule
\cellcolor{interpolationlabelcolor} & DR & 3 & $443.40\pm5.78$ & $433.15\pm2.43$ & $446.21\pm8.45$ & $475.42\pm1.27$ & $432.23\pm2.64$ & $455.79\pm6.96$ & \bestresult{429.70\pm3.54} & \secondresult{430.41\pm3.72} \\
\cellcolor{interpolationlabelcolor} & DR & 5 & $468.39\pm0.68$ & $465.22\pm1.10$ & $468.02\pm1.97$ & $492.41\pm1.06$ & \bestresult{457.57\pm2.85} & $466.01\pm0.99$ & \secondresult{463.32\pm0.60} & $466.04\pm1.46$ \\
\cellcolor{interpolationlabelcolor}\multirow{-3}{*}{Interpolation} & DR & 7 & $534.69\pm8.30$ & $519.13\pm0.49$ & $533.38\pm1.03$ & $541.43\pm0.77$ & \secondresult{519.12\pm0.67} & $526.89\pm0.67$ & $522.50\pm2.41$ & \bestresult{518.49\pm1.41} \\
\noalign{\vskip 1.5pt}
\cdashline{1-11}
\noalign{\vskip 1.5pt}
\cellcolor{extrapolationlabelcolor} & DR & 9 & $602.99\pm5.41$ & $567.25\pm0.94$ & $617.05\pm5.86$ & $604.33\pm0.80$ & $577.97\pm0.83$ & $592.17\pm5.07$ & \secondresult{558.89\pm2.21} & \bestresult{556.80\pm1.27} \\
\cellcolor{extrapolationlabelcolor} & DR & 10 & $593.05\pm2.68$ & $570.52\pm5.15$ & $673.08\pm24.93$ & $565.45\pm0.37$ & $603.87\pm2.94$ & $653.24\pm19.74$ & \secondresult{563.86\pm3.26} & \bestresult{557.68\pm3.56} \\
\cellcolor{extrapolationlabelcolor}\multirow{-3}{*}{Extrapolation} & DR & 11 & $726.16\pm5.89$ & $713.37\pm4.70$ & $860.26\pm87.00$ & $716.82\pm0.96$ & $751.16\pm5.70$ & $851.07\pm22.16$ & \bestresult{694.53\pm4.21} & \secondresult{699.49\pm4.56} \\
\midrule
\cellcolor{interpolationlabelcolor} & SC & 6 & $54.44\pm0.28$ & \secondresult{48.66\pm0.47} & $53.87\pm4.29$ & $64.70\pm0.88$ & \bestresult{47.35\pm0.04} & $48.94\pm0.34$ & $50.20\pm1.58$ & $49.49\pm0.63$ \\
\cellcolor{interpolationlabelcolor} & SC & 8 & $58.45\pm0.94$ & \bestresult{52.79\pm0.22} & $63.28\pm3.71$ & $66.35\pm0.52$ & \secondresult{53.03\pm0.26} & $53.68\pm0.80$ & $54.00\pm0.68$ & $53.44\pm0.28$ \\
\cellcolor{interpolationlabelcolor} & SC & 10 & $103.23\pm1.21$ & \secondresult{96.40\pm0.28} & $110.39\pm5.17$ & $108.01\pm0.22$ & \bestresult{95.14\pm1.39} & $108.46\pm1.37$ & $96.72\pm0.35$ & $96.99\pm0.32$ \\
\cellcolor{interpolationlabelcolor}\multirow{-4}{*}{Interpolation} & SC & 12 & $142.30\pm1.36$ & $128.31\pm1.26$ & $156.91\pm7.55$ & $152.00\pm2.76$ & $137.35\pm0.78$ & $166.69\pm3.40$ & \secondresult{127.47\pm0.92} & \bestresult{127.33\pm0.37} \\
\noalign{\vskip 1.5pt}
\cdashline{1-11}
\noalign{\vskip 1.5pt}
\cellcolor{extrapolationlabelcolor} & SC & 16 & $128.25\pm3.90$ & \secondresult{108.53\pm1.37} & $160.34\pm21.35$ & $134.07\pm0.64$ & $125.81\pm1.99$ & $159.34\pm1.88$ & $109.78\pm1.73$ & \bestresult{105.61\pm1.48} \\
\cellcolor{extrapolationlabelcolor} & SC & 17 & $142.40\pm4.78$ & \secondresult{114.45\pm1.06} & $195.20\pm44.48$ & $142.04\pm1.57$ & $147.12\pm3.60$ & $161.57\pm2.74$ & $118.54\pm4.86$ & \bestresult{111.87\pm0.64} \\
\cellcolor{extrapolationlabelcolor} & SC & 18 & $131.79\pm5.62$ & $118.72\pm3.29$ & $203.72\pm60.02$ & $131.42\pm1.30$ & $140.50\pm5.75$ & $169.55\pm4.90$ & \secondresult{109.89\pm5.23} & \bestresult{108.69\pm2.28} \\
\cellcolor{extrapolationlabelcolor}\multirow{-4}{*}{Extrapolation} & SC & 19 & $145.91\pm8.78$ & $132.99\pm4.67$ & $243.97\pm92.21$ & $136.76\pm1.76$ & $160.55\pm7.53$ & $181.55\pm9.57$ & \secondresult{121.87\pm8.65} & \bestresult{117.16\pm3.84} \\
\bottomrule
\end{tabular}
}
\end{table*}

\subsection{Spectral Recovery from Unpaired Snapshots}
\label{subsec:toy-spectral-recovery}

To test whether KoopCell can recover the underlying Koopman operator from
unpaired snapshots, and whether the recovered spectral structure translates
into more accurate temporal prediction, we consider two toy systems with
known Koopman spectra that are widely studied in the literature:
\begin{enumerate}
  \item[(Toy I)] Following \citet{ONDMD}, we consider
  $\dot x_1=-0.2x_1$ and $\dot x_2=-(x_2-x_1^2)$.  The system has the
  attracting slow manifold $x_2=\frac{5}{3}x_1^2$ and admits the exact
  Koopman observables $\Psi_1=x_1$,
  $\Psi_2=x_2-\frac{5}{3}x_1^2$, and $\Psi_3=x_1^2$.  Consequently,
  $\dot{\bm{\Psi}}=\bm{A}\bm{\Psi}$ with
  $\bm{A}=\operatorname{diag}(-0.2,-1,-0.4)$.
  \item[(Toy II)] Following \citet{otto2019linearly}, we consider the
  unforced Duffing equation
  $\ddot q=-\delta\dot q-q(\beta+\alpha q^2)$ with
  $\delta=0.5$, $\beta=-1$, and $\alpha=1$.  The system has two stable
  equilibria at $q=\pm1$, with the conjugate intrawell eigenpair
  $\lambda_{1,2}=(-1\pm\sqrt{31}i)/4$.  Across the two wells, it also
  admits an eigenfunction with zero eigenvalue that separates the invariant basins.
\end{enumerate}

We generate 2D unpaired snapshots by solving these equations over short
training horizons during which the populations remain far from equilibrium.
For Toy I, training comprises 9 snapshots at
$t\in[0,1.2]$, and extrapolation is evaluated
at $0.2$-time-unit intervals from $t=1.4$ to $4.0$. For Toy II, training
comprises 10 snapshots at $t=0,0.5,\ldots,4.5$, and extrapolation is evaluated at
$t\in\{5,6,8,10,12\}$. We also synthesize a 64D version of each dataset using
$\bm{y}=\operatorname{softplus}(\bm{W}\bm{x}+\bm{c})+\bm{\epsilon}\in\mathbb{R}^{64}$, with
$\bm{x}\in\mathbb R^2$ and Gaussian observation noise $\bm{\epsilon}$. \textcolor{black}{Since the underlying systems are autonomous, we omit $t$ from the encoder input.}
 We compare KoopCell with
scNODE~\citep{zhang2024scnode}, which parameterizes the latent dynamics with a
neural ODE. Both methods use the same latent dimension of $D=3$ across all
experiments. After training, we roll out each model from the observed initial
distribution beyond the training horizon and report the mean sliced
Wasserstein distance (SWD) to the ground-truth snapshots. Complete protocols
and full results are provided in Appendix~\ref{app:toy-models}.

\begin{figure}[!ht]
  \centering
  \setlength{\abovecaptionskip}{2pt}
  \setlength{\belowcaptionskip}{0pt}
  \begin{minipage}[c]{0.533\textwidth}
    \centering
    \includegraphics[width=\linewidth]{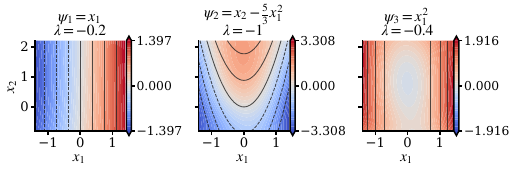}
    \vspace{-1.2em}

    {\scriptsize (a) Toy I: learned eigenfunctions aligned with $\{x_1, x_2-\frac53 x_1^2, x_1^2\}$.\par}
    \vspace{0.15em}
    \includegraphics[width=\linewidth]{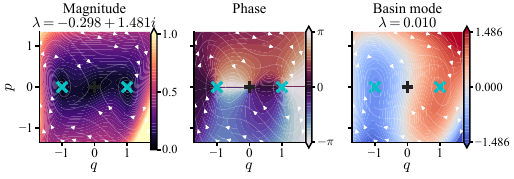}
    \vspace{-1.2em}

    {\scriptsize (b) Toy II: learned oscillatory and basin eigenmodes.\par}
  \end{minipage}
  \hfill
  \begin{minipage}[c]{0.452\textwidth}
    \centering
    \makeatletter
    \def\@captype{table}
    \makeatother
    \caption{Spectral recovery (MAE) and extrapolation comparison with scNODE (SWD $\downarrow$).}
    \label{tab:toy-main}
    \vspace{0.35em}
    \scriptsize
    \setlength{\tabcolsep}{1.0pt}
    \renewcommand{\arraystretch}{1.10}
    \begin{tabular*}{\linewidth}{@{\extracolsep{\fill}}cc@{}}
      \toprule
      Toy I & Toy II \\
      \midrule
      \multicolumn{2}{c}{Recovered Spectrum} \\
      $\{-0.195,-0.423,-0.977\}$ & $\{0.010,-0.298\pm1.481i\}$ \\
      \midrule
      \multicolumn{2}{c}{Spectrum MAE} \\
      $0.0167$ & $0.0710$ \\
      \midrule
    \end{tabular*}
    \vspace{-0.65em}
    \begin{tabular*}{\linewidth}{@{\extracolsep{\fill}}cccccc@{}}
          & \multirow[c]{2}{*}{Model}
                  & \multicolumn{2}{c}{Toy I} & \multicolumn{2}{c}{Toy II} \\
          &       & \textbf{2D} & \textbf{64D} & \textbf{2D} & \textbf{64D} \\
      \cmidrule(lr){3-6}
      \smash{\raisebox{-0.55\baselineskip}{SWD}}
          & scNODE   & $0.1936$ & $0.2011$ & $0.1076$ & $0.0277$ \\
          & KoopCell & $\mathbf{0.0665}$ & $\mathbf{0.0641}$ &
                       $\mathbf{0.0592}$ & $\mathbf{0.0175}$ \\
      \bottomrule
    \end{tabular*}
  \end{minipage}
  \vspace{0.4em}
  \caption{Koopman eigenmodes learned from unpaired 2D snapshots.  In (a),
  colors show the learned modes and black contours show the analytic
  eigenfunctions.  In (b), the complex mode separates intrawell amplitude and
  phase, whereas the near-zero mode separates the two attraction basins; sink
  and saddle markers are included only for visualization.}
  \label{fig:toy-spectral-recovery}
  \label{fig:toy1-observables}
  \label{fig:toy2-eigenmodes}
\end{figure}

\textbf{Spectral recovery.} Table~\ref{tab:toy-main}
reports the recovered continuous-time spectrum of the learned matrix $\bm{A}$, while
Figure~\ref{fig:toy-spectral-recovery} shows the corresponding eigenfunctions.
KoopCell consistently recovers the Koopman spectrum with low MAE, and, for
Toy I, the learned observables reproduce the level-set geometry of the
ground-truth observables. For Toy II, we visualize the learned complex
eigenfunction through its magnitude $|\Psi|$ and phase $\angle\Psi$
(Figure~\ref{fig:toy-spectral-recovery}(b), left and middle). The magnitude
decays toward zero at both attractors, $q=\pm1$, while the phase reveals the oscillatory relaxation within each basin.
Meanwhile, the eigenmode associated with the zero eigenvalue acts as an
indicator that separates the two basins of attraction. Overall, these results
demonstrate KoopCell's ability to recover the underlying spectral structure
from unpaired, temporally sparse distribution snapshots. 

{\textbf{Spectral structure supports stable extrapolation.}
Across all settings, KoopCell achieves lower extrapolation SWD than scNODE
(Table~\ref{tab:toy-main}). Together with the spectral recovery results,
this suggests that learning the underlying Koopman spectral structure helps
capture the long-term evolution of the system and supports more stable
extrapolation beyond the training horizon.}

\subsection{Modeling Branching Dynamics with KoopCell-M}
\label{subsec:toy-branching}
\begin{figure}[!ht]
  \centering
  \setlength{\abovecaptionskip}{2pt}
  \setlength{\belowcaptionskip}{0pt}
  \begin{minipage}[c]{0.515\textwidth}
    \centering
    \includegraphics[width=\linewidth]{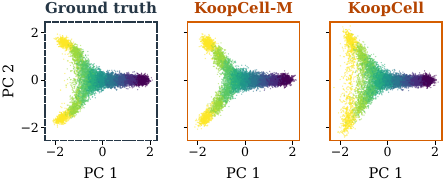}
  \end{minipage}\hfill
  \begin{minipage}[c]{0.425\textwidth}
    \centering
    \makeatletter\def\@captype{table}\makeatother
    \caption{Distributional prediction errors averaged over held-out times (mean $\pm$ SD over three optimization seeds).}
    \label{tab:branching-main}
    \scriptsize
    \setlength{\tabcolsep}{1pt}
    \renewcommand{\arraystretch}{1.03}
    \input{figures/branching/table_main.tex}
  \end{minipage}
  \caption{Population rollouts at all nine training times in a shared PCA
  plane (1,024 cells per time). Colors progress from purple to yellow with
  time; all panels use identical axes.}
  \label{fig:branching-rollouts}
\end{figure}

We compare KoopCell and KoopCell-M on a stochastic pitchfork system in which
an initially unimodal population separates into two branches. A fixed nonlinear
map produces 64D noisy observations at nine training times.
We initialize both models with the same pretrained autoencoder and jointly
train their autoencoders and dynamics using the same population-matching objective.
We impose the same Lipschitz bound on both decoders to limit the
amplification of latent separation during decoding.
Figure~\ref{fig:branching-rollouts} overlays the population rollouts, while
Table~\ref{tab:branching-main} summarizes SWD and branch-coordinate $W_1$
over all held-out times; definitions of both metrics and experimental details
are provided in Appendix~\ref{app:toy-branching}.
KoopCell-M more closely reproduces the separation
of the two branches and achieves lower errors on both metrics, as expected.

\subsection{Modeling High-dimensional Single-Cell Population Dynamics}\label{subsec: realworld_cellular_dynamics}

To assess the scalability of KoopCell and KoopCell-M, we evaluate them on
three high-dimensional single-cell RNA-sequencing datasets:
zebrafish embryogenesis (ZB)~\citep{doi:10.1126/science.aar3131},
Drosophila embryogenesis (DR)~\citep{doi:10.1126/science.abn5800}, and mouse
iPSC reprogramming (SC)~\citep{schiebinger2019optimal}.
Dataset descriptions, preprocessing, and visualizations are provided in
Appendix~\ref{app:single-cell-data}.

\begin{wrapfigure}{R}{0.36\textwidth}
  \centering
  \includegraphics[width=\linewidth]{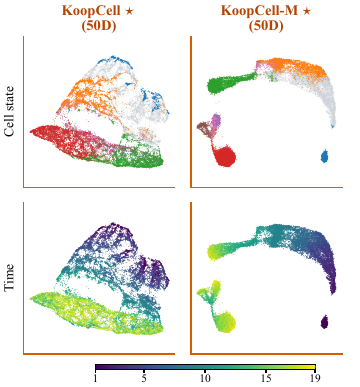}
  \setlength{\abovecaptionskip}{2pt}
  \caption{Latent representations of SC cells colored by reference state
  (top; stromal: green, iPSC: red) and snapshot index (bottom).}
  \label{fig:latent-representations}
\end{wrapfigure}

\textbf{Tasks and evaluation protocol.}
Following \citet{zhang2024scnode}, we evaluate
our models on three tasks of increasing difficulty to
assess their ability to predict gene expression distributions at held-out
time points, both within and beyond the training horizon. For each task,
we train on the populations at the retained time points. At evaluation,
we sample cells from the initial snapshot, propagate their encoded states
forward, and decode the predicted populations at the held-out times.
Generated and observed distributions are compared in gene-expression space
using the Wasserstein distance $W_2$. The tasks differ in which snapshots
are withheld: \textbf{(1) Easy (interpolation)} removes selected intermediate
snapshots to assess prediction within the training horizon;
\textbf{(2) Medium (extrapolation)} removes several final snapshots to assess
prediction beyond that horizon; and \textbf{(3) Hard (interpolation and
extrapolation)} removes both intermediate and several final snapshots to
assess the ability to recover population dynamics from highly sparse,
unpaired distribution snapshots.
Figure~\ref{fig:single-cell-task-splits} shows the training and held-out time points.

\textbf{Baselines.} We compare against several representative models for temporal dynamics of single-cell populations, including \textbf{scNODE} \citep{zhang2024scnode}, \textbf{LGP-OT} \citep{balik2026modeling}, \textbf{MIOFlow} \citep{huguet2022manifold}, \textbf{PRESCIENT} \citep{yeo2021generative}, \textbf{PI-SDE} \citep{10.1093/bioinformatics/btae400}, and \textbf{VGFM} \citep{wang2026joint}.
KoopCell(-M), scNODE and LGP-OT jointly learn an informative latent representation and the latent dynamics, while other models learn dynamics in a fixed latent space obtained by dimensionality reduction.
We tune all models to their optimal hyperparameters (Appendix \ref{appendixsubsec: implementation_baselines})
and report the mean and std over three training seeds.

\textbf{Interpolation and extrapolation performance.}
KoopCell achieves strong predictive performance across all three datasets
and tasks
(Table~\ref{tab:wasserstein-hard-tasks}; Appendix
Tables~\ref{tab:wasserstein-easy} and~\ref{tab:wasserstein-medium}).
While some baselines attain comparable or slightly lower errors at individual
interpolation times, KoopCell achieves lower average $W_2$ distances than
all baselines on both the medium task and the extrapolation portion of
the hard task.
Incorporating memory further enhances predictive performance compared
to the linear model (KoopCell). Relative to the strongest baseline for each dataset,
KoopCell-M reduces the average error by $3.6$--$7.3\%$ on the medium task
and $2.0$--$6.6\%$ on the extrapolation portion of the hard task.
These results highlight the ability of our models to learn population
dynamics from highly sparse, unpaired snapshots and extrapolate reliably
beyond the training horizon.
Appendix Figures~\ref{fig:easy-populations}--\ref{fig:hard-populations}
provide the corresponding visualization comparisons of predicted cell distributions.

\textbf{Dynamics Shapes Representation.}
Figure~\ref{fig:latent-representations} compares latent representations learned
by KoopCell and KoopCell-M on the SC dataset under the hard-task setting.
The complete comparison of latent representations, including those from
PCA, scNODE, and MIOFlow, appears in Appendix
Figure~\ref{fig:latent-representations-full}.
We match cells to published Waddington-OT reference states and label the
initial population as mouse embryonic fibroblasts (MEFs), leaving
unmatched or ambiguous cells gray.
Reprogramming involves divergence toward stromal states or
mesenchymal-to-epithelial transition (MET), from which iPSCs and alternative
neural- and trophoblast-like states emerge~\citep{schiebinger2019optimal}.
KoopCell-M more clearly separates stromal and iPSC populations with
distinct temporal progressions, whereas these late populations overlap
substantially in KoopCell. These qualitative patterns are consistent with
the branching structure of the memory-augmented model.

\textbf{Biological Downstream Analysis.}
Following \citet{zhang2024scnode}, we examine
lineage-associated genes in ZB using KoopCell-M fitted to all snapshots,
focusing on Presomitic Mesoderm (PSM) and Hindbrain.
Using the driver-gene statistic from CellRank~\citep{weiler2024cellrank},
we correlate preterminal gene expression with predicted fate probabilities and recover
\textit{TBX16}/\textit{MSGN1} for PSM and \textit{SOX3}/\textit{SOX19A} for
Hindbrain (Appendix Figure~\ref{fig:biological-markers}).
We further characterize dynamic gene-expression programs along model-predicted
trajectories over 8--12 hours post fertilization (hpf). Gene modules with
similar predicted expression-rate profiles are enriched for muscle and neural
development in the PSM and Hindbrain lineages, respectively.
Appendix~\ref{app:biological-downstream} provides
methods and additional results.
Together, these analyses support the biological interpretability of our model.

\section{Related Work}
\textbf{Learning scRNA-seq dynamics from distribution snapshots.}
Several approaches have been proposed to infer temporal dynamics from unpaired distributions.
TrajectoryNet~\citep{tong2020trajectorynet} fits continuous normalizing flows
regularized by dynamic OT in fixed PCA coordinates, while PRESCIENT and
PI-SDE~\citep{yeo2021generative,10.1093/bioinformatics/btae400} model
potential-driven and Hamilton--Jacobi-regularized SDEs, respectively.
MIOFlow~\citep{huguet2022manifold} fits neural ODEs in a pretrained
geodesic embedding; scNODE~\citep{zhang2024scnode} jointly trains a VAE
and latent neural ODE through dynamic regularization, and
LGP-OT~\citep{balik2026modeling} learns a latent Gaussian
process and a decoder using an OT objective. Additionally, proliferation
and death are accommodated through unbalanced
OT~\citep{zhang2025learning, sha2024reconstructing, sun2025variational}.
VGFM~\citep{wang2026joint} and WFR-FM~\citep{peng2026wfrfm} further combine
unbalanced OT with flow matching~\citep{lipman2023flow, tong2024improving, tong2023simulation} for scalable
joint velocity and growth estimation. 

\textbf{Koopman learning and memory.}
Koopman theory~\citep{koopman1931hamiltonian} describes nonlinear dynamics
through the linear evolution of observables, motivating the joint
learning of autoencoders and latent linear propagators for dynamical prediction
and time-series generation
\citep{otto2019linearly,nc_deep_koopman,liu2023koopa,naiman2023generative}.
While these methods learn from paired time-series data, we develop
theory and algorithms for generative Koopman learning
from unpaired distribution snapshots. Finding observables with exact linear
closure remains challenging. Recent work uses the Mori--Zwanzig
formalism~\citep{mori1965transport,zwanzig1961memory} to learn
non-Markovian memory closures to account for unresolved variables
and improve dynamical prediction
\citep{lin2021datamori,iLED,gupta2025mori,lu2026interpretable}.
Building on this principle, we learn a memory closure through an
augmented Markovian system to model branching population dynamics
from snapshots.

\section{Conclusions}
We introduced KoopCell, a unified framework for learning Koopman generative
dynamics from unpaired distribution snapshots. The weak population residual
in Wasserstein space yields a closed-form least-squares estimator of the
linear generator with convergence guarantees, while a Markovian memory
embedding further enables KoopCell-M to capture branching dynamics.
Our theoretical and empirical findings demonstrate the ability to recover Koopman spectra and improve
extrapolation of gene-expression distributions from sparse, unpaired measurements,
supporting quantitative analysis of developmental dynamics beyond the observed time window.
Future directions include applications to multi-omics data and learning
stochastic Koopman operators from distribution snapshots.

\subsection*{AI use statement}
In this work, we used generative AI tools to assist with code implementation
and debugging, including baseline implementations. We also used these tools
to help verify the correctness and technical details of our mathematical
arguments and algorithms, and to support synthetic data generation,
experimental visualizations, and language polishing. We did not use
generative AI tools to generate research ideas, develop the theoretical
framework, formulate the main mathematical results, or design the core
algorithms. We reviewed all AI-assisted work and take full responsibility
for the final content, including the mathematical arguments, code, data,
figures, and text.

\subsection*{Ethics statement}

This work develops computational methods for modeling cellular population
dynamics. Our experiments use synthetic data and publicly available single-cell
datasets from zebrafish, Drosophila, and mouse, and involve no new biological
sample collection or human participants. We anticipate no foreseeable adverse
societal effects from this work.

\subsection*{Reproducibility statement}

We facilitate reproducibility by documenting the assumptions and proofs of
our theoretical results in Appendix~\ref{app:theory}, and the estimation
and training procedures for KoopCell and KoopCell-M, including pseudocode,
in Appendix~\ref{app:algorithmic-details}.
Synthetic data generation and experimental protocols are described in
Appendices~\ref{app:toy-models} and~\ref{app:toy-branching}.
Single-cell dataset sources and preprocessing
are provided in Appendix~\ref{app:single-cell-data}, with training and
held-out time points, evaluation metrics in
Appendix~\ref{app:single-cell-protocol}.
Baseline configurations and model-selection procedures are documented in
Appendix~\ref{appendixsubsec: implementation_baselines}, and the biological
downstream analyses are detailed in Appendix~\ref{app:biological-downstream}.
We will release our code as open source upon acceptance of the paper.

\bibliography{references}
\bibliographystyle{iclr2027_conference}

\clearpage
\appendix
\etocdepthtag.toc{appendix}
\begingroup
  \etocsettagdepth{main}{none}
  \etocsettagdepth{appendix}{subsubsection}
  \etocobeydepthtags
  \etocsettocstyle{\section*{Appendix Contents}}{}
  \setlength{\parskip}{0pt}
  \hypersetup{linktoc=all}
  \pdfbookmark[1]{Appendix Contents}{appendix.contents}
  \tableofcontents
\endgroup
\clearpage
\section{Theory}
\label{app:theory}

This appendix develops the theoretical foundations of estimating the
linear generator based on the weak population residual, introduced in Section~\ref{sec:linear-comparison}.
Sections~\ref{app:wc-geometry-v6}--\ref{app:finite-tests-v6} prove
Theorem~\ref{thm:main-finite-test}: minimizing the {weak population residual} gives
the best projected linear approximation to the canonical Wasserstein velocity,
and finite tests yield a least-squares regression problem whose minimizers converge
under dense test enrichment and uniqueness. We then consider increasing the observable
dimension in a Koopman-invariant RKHS. Section~\ref{app:operator-consistency-v6}
shows that the minimum weak population residual vanishes as $D\to\infty$;
Section~\ref{app:koopman-consequences-v6} relates this consistency to
Wasserstein prediction bounds and locally uniform strong convergence of
the learned propagators, under the respective approximation, stability,
and identifiability assumptions. Algorithmic details and pseudocode are
provided in Appendix~\ref{app:algorithmic-details}.

\subsection{Dual formulation and geometry of the weak population residual}
\label{app:wc-geometry-v6}

Throughout Sections~\ref{app:wc-geometry-v6}--\ref{app:finite-tests-v6}, the
observable dimension $D$ is fixed. Following
Section~\ref{sec:linear-comparison}, we write
$\mu_t\in\mathcal P_2(\mathbb R^D)$ for the latent distribution induced by
the chosen observables at time $t$.
We work over real Hilbert spaces; complex-valued tests can be handled by
separating real and imaginary parts.

\subsubsection{Canonical velocity and the space--time test space}

Assume
\begin{equation}\label{eq:AC2-assumption-v6}
  \mu_\cdot\in AC^2([0,T];\mathcal P_2(\mathbb R^D)).
\end{equation}
Then there is a unique minimum-norm velocity
$\bm{v}_t\in T_{\mu_t}\mathcal P_2$ such that
\begin{equation}\label{eq:canonical-velocity-app-v6}
  \partial_t\mu_t+\nabla\!\cdot(\bm{v}_t\mu_t)=0,
  \qquad
  \int_0^T\|\bm{v}_t\|_{L^2(\mu_t)}^2\mathrm{d}t<\infty.
\end{equation}
Here
\begin{equation}\label{eq:tangent-app-v6}
  T_\mu\mathcal P_2
  =\overline{\{\nabla\phi:\phi\in C_c^\infty(\mathbb R^D)\}}^{L^2(\mu)},
\end{equation}
and $P_\mu$ denotes the $L^2(\mu)$-orthogonal projection onto this tangent
space.  We collect these spaces along the curve in
\begin{equation}\label{eq:direct-integral-v6}
  \mathscr T_\mu
  =L^2\!\left(0,T;T_{\mu_t}\mathcal P_2\right),
  \qquad
  \|\bm{q}\|_{\mathscr T_\mu}^2
  =\int_0^T\int\|\bm{q}_t(\bm{z})\|^2\mathrm{d}\mu_t(\bm{z})\mathrm{d}t.
\end{equation}

Let $\mathscr D$ consist of $C^1$ space--time potentials compactly
supported in space uniformly in $t$. We identify two potentials whose spatial gradients agree
in $L^2(\mathrm{d}t\,\mathrm{d}\mu_t)$ and define $\mathbb V_D$ as the completion under
\begin{equation}\label{eq:V-norm-app-v6}
  \|\varphi\|_{\mathbb V_D}^2
  =\int_0^T\int
  \|\nabla_{\bm{z}}\varphi(t,\bm{z})\|^2\mathrm{d}\mu_t(\bm{z})\mathrm{d}t.
\end{equation}

\begin{lemma}[Density of smooth space--time gradients]
\label{lem:density-gradients-v6}
The map $\varphi\mapsto\nabla_{\bm{z}}\varphi$ extends to an isometric isomorphism
from $\mathbb V_D$ onto $\mathscr T_\mu$.  In particular, finite sums
\begin{equation}\label{eq:separable-tests-v6}
  \sum_{\ell=1}^L\eta_\ell(t)\nabla\psi_\ell(\bm{z}),
  \qquad
  \eta_\ell\in C^1([0,T]),\quad
  \psi_\ell\in C_c^\infty(\mathbb R^D),
\end{equation}
are dense in $\mathscr T_\mu$.
\end{lemma}

\begin{proof}
The gradient map is an isometry from the quotient of $\mathscr D$ into
$\mathscr T_\mu$: spatial mollification approximates each $C_c^1$
potential by smooth compactly supported potentials with uniformly
convergent gradients. To prove density, let $\bm{q}\in\mathscr T_\mu$ be orthogonal
to every field in \Eqref{eq:separable-tests-v6}.  For fixed
$\psi\in C_c^\infty$, set
\begin{equation*}
  F_\psi(t)=\int \bm{q}_t(\bm{z})^\top\nabla\psi(\bm{z})\mathrm{d}\mu_t(\bm{z}).
\end{equation*}
Cauchy--Schwarz and boundedness of $\nabla\psi$ imply $F_\psi\in L^1(0,T)$.
Orthogonality to $\eta(t)\nabla\psi(\bm{z})$ for every smooth $\eta$ therefore
implies $F_\psi(t)=0$ for almost every $t$.

For each integer $m\ge1$, choose a countable $C^1$-norm dense subset of
$C_c^\infty(B_m(0))$, viewing its members as functions on $\mathbb R^D$,
and enumerate their union as $\{\psi_n\}$. Such subsets exist by
separability of $C^1(\overline{B_m(0)})$.
Every $\psi\in C_c^\infty(\mathbb R^D)$ can then be approximated by
members supported in one fixed ball, with
$\|\nabla\psi_n-\nabla\psi\|_\infty\to0$ along a suitable sequence.
Since $\mu_t$ is a probability measure,
$\|\nabla\psi_n-\nabla\psi\|_{L^2(\mu_t)}
\le\|\nabla\psi_n-\nabla\psi\|_\infty$.
Thus the same countable family generates a dense gradient class for
every $t$. Removing the union of the null sets
obtained above, we have, for almost every $t$,
\begin{equation*}
  \int \bm{q}_t^\top\nabla\psi_n\,\mathrm{d}\mu_t=0
  \qquad\text{for every }n.
\end{equation*}
Since $\bm{q}_t\in T_{\mu_t}\mathcal P_2$, density forces $\bm{q}_t=0$.  Hence the
orthogonal complement of the range is trivial, proving density and the
claimed isomorphism after completion.
\end{proof}
This lemma allows optimization over completed test potentials to be
identified with optimization over all square-integrable tangent velocity
fields, as used in Theorem~\ref{thm:dual-geometry-v6} below.

\subsubsection{Residual functional and minimum-flux geometry}

For $\bm{A}\in\mathbb R^{D\times D}$ define
\begin{equation}\label{eq:residual-app-v6}
  r_{\bm{A}}(t)=\partial_t\mu_t+\nabla\!\cdot(\bm{A}\bm{z}\,\mu_t).
\end{equation}
For a smooth test $\varphi\in \mathscr{D}$, define
$\mathcal R_D(\bm{A};\varphi):=\langle r_{\bm{A}},\varphi\rangle$.
Integration by parts gives
\begin{equation}
  \mathcal R_D(\bm{A};\varphi)
  =\mu_T[\varphi(T)]-\mu_0[\varphi(0)]
  -\int_0^T\mu_t\!\left[
    \partial_t\varphi(t)+\nabla_{\bm{z}}\varphi(t)^\top \bm{A}\bm{z}
  \right]\mathrm{d}t.
  \label{eq:residual-pair-app-v6}
\end{equation}
Using \Eqref{eq:canonical-velocity-app-v6}, the same quantity is
\begin{equation}\label{eq:residual-velocity-pair-v6}
  \mathcal R_D(\bm{A};\varphi)
  =\int_0^T\int
  \nabla_{\bm{z}}\varphi(t,\bm{z})^\top(\bm{v}_t(\bm{z})-\bm{A}\bm{z})\mathrm{d}\mu_t(\bm{z})\mathrm{d}t.
\end{equation}
Thus it extends continuously to $\mathbb V_D$.

For a zero-mass distribution $\sigma$, define
\begin{equation}\label{eq:weighted-Hminus-v6}
  \|\sigma\|_{-1,\mu}
  =\sup_{\phi\in C_c^\infty(\mathbb R^D)}\left\{
    \langle\sigma,\phi\rangle:
    \int\|\nabla\phi\|^2\mathrm{d}\mu\leq1
  \right\}.
\end{equation}

\begin{lemma}[Minimum-flux representation]
\label{lem:min-flux-v6}
If $\bm{w}\in L^2(\mu;\mathbb R^D)$ and
$\sigma=-\nabla\!\cdot(\bm{w}\mu)$, then
\begin{equation}\label{eq:min-flux-equality-v6}
  \|\sigma\|_{-1,\mu}=\|P_\mu \bm{w}\|_{L^2(\mu)}.
\end{equation}
Moreover, $P_\mu \bm{w}$ is the unique min-norm vector field among all
$L^2(\mu)$ fields satisfying $-\nabla\cdot(\bm{q}\mu)=\sigma$.
Conversely, if $\|\sigma\|_{-1,\mu}<\infty$, a tangent field satisfying
this constraint exists.
\end{lemma}

\begin{proof}
For every smooth $\phi$,
\begin{align*}
  \langle\sigma,\phi\rangle
  &=\int\nabla\phi^\top \bm{w}\,\mathrm{d}\mu
   =\int\nabla\phi^\top P_\mu \bm{w}\,\mathrm{d}\mu,
\end{align*}
because $\bm{w}-P_\mu \bm{w}$ is orthogonal to the tangent space.  Cauchy--Schwarz
proves the upper bound in \Eqref{eq:min-flux-equality-v6}.  Conversely,
if $P_\mu \bm{w}\ne0$, choose gradients converging in $L^2(\mu)$ to
$P_\mu \bm{w}$ and normalize them.
Their pairings converge to $\|P_\mu \bm{w}\|$, proving equality.
If $P_\mu \bm{w}=0$, all pairings vanish and equality is immediate.

If $\bm{q}$ has the same weighted divergence, then
$\bm{q}-\bm{w}$ is orthogonal to all smooth gradients, hence
$P_\mu \bm{q}=P_\mu \bm{w}$.  Orthogonal decomposition gives
\begin{equation*}
  \|\bm{q}\|_{L^2(\mu)}^2
  =\|P_\mu \bm{w}\|_{L^2(\mu)}^2
   +\|(I-P_\mu)\bm{q}\|_{L^2(\mu)}^2.
\end{equation*}
The minimum is therefore attained uniquely at $\bm{q}=P_\mu \bm{w}$.

For the converse, finiteness of $\|\sigma\|_{-1,\mu}$ makes
$\nabla\phi\mapsto\langle\sigma,\phi\rangle$ a well-defined bounded
linear functional on smooth gradients. It extends continuously to
$T_\mu\mathcal P_2$, so the Riesz representation theorem gives a unique
$\bm{q}\in T_\mu\mathcal P_2$ with
$\langle\sigma,\phi\rangle=\int \bm{q}^\top\nabla\phi\,\mathrm{d}\mu$
for every $\phi\in C_c^\infty(\mathbb R^D)$.
Thus $\sigma=-\nabla\cdot(\bm{q}\mu)$, and the preceding argument applies.
\end{proof}

\begin{theorem}[Dual representation, exact geometry, and population optimum]
\label{thm:dual-geometry-v6}
For every $\bm{A}\in\mathbb R^{D\times D}$,
\begin{align}
  \mathcal J_D(\bm{A})
  &:=\frac{1}{2}\int_0^T\|r_{\bm{A}}(t)\|_{-1,\mu_t}^2\mathrm{d}t
  \label{eq:J-def-app-v6}\\
  &=\sup_{\varphi\in\mathbb V_D}
  \left\{
    \mathcal R_D(\bm{A};\varphi)
    -\frac{1}{2}\|\varphi\|_{\mathbb V_D}^2
  \right\}
  \label{eq:J-dual-app-v6}\\
  &=\frac{1}{2}\int_0^T
  \|\bm{v}_t-P_{\mu_t}(\bm{A}\bm{z})\|_{L^2(\mu_t)}^2\mathrm{d}t.
  \label{eq:J-geometry-app-v6}
\end{align}
A minimizer $\bm{A}_D^\dagger$ exists.  It satisfies
\begin{equation}\label{eq:normal-matrix-v6}
  \int_0^T\int
  \bigl(\bm{v}_t(\bm{z})-P_{\mu_t}(\bm{A}_D^\dagger \bm{z})\bigr)\bm{z}^\top
  \mathrm{d}\mu_t(\bm{z})\mathrm{d}t=0.
\end{equation}
If
\begin{equation}\label{eq:fixedD-identifiability-v6}
  P_{\mu_t}(\bm{B}\bm{z})=0\text{ in }L^2(\mu_t)\text{ for a.e. }t
  \quad\Longrightarrow\quad \bm{B}=0,
\end{equation}
then the minimizing matrix is unique. Every minimizer satisfies
\begin{equation}\label{eq:pythagorean-v6}
  \mathcal J_D(\bm{A})
  =\mathcal J_D(\bm{A}_D^\dagger)
  +\frac{1}{2}\int_0^T
  \|P_{\mu_t}((\bm{A}-\bm{A}_D^\dagger)\bm{z})\|_{L^2(\mu_t)}^2\mathrm{d}t.
\end{equation}
\end{theorem}

\begin{proof}
The canonical equation gives
$r_{\bm{A}}=-\nabla\!\cdot((\bm{v}_t-\bm{A}\bm{z})\mu_t).$
Lemma~\ref{lem:min-flux-v6} and $P_{\mu_t}\bm{v}_t=\bm{v}_t$ yield, for almost every
$t$,
$\|r_{\bm{A}}(t)\|_{-1,\mu_t}
  =\|\bm{v}_t-P_{\mu_t}(\bm{A}\bm{z})\|_{L^2(\mu_t)}.$
This proves \Eqref{eq:J-geometry-app-v6}.
To obtain \Eqref{eq:J-dual-app-v6}, set $\bm{h}_t=\bm{v}_t-P_{\mu_t}(\bm{A}\bm{z})$.
For smooth $\varphi$, \Eqref{eq:residual-velocity-pair-v6} and orthogonal
projection give
$\mathcal R_D(\bm{A};\varphi)=\langle \bm{h},\nabla_{\bm{z}}\varphi\rangle_{\mathscr T_\mu}$,
where the inner product integrates over both space and time.
This identity extends continuously to $\mathbb V_D$.
Lemma~\ref{lem:density-gradients-v6} makes the correspondence
$\varphi\mapsto \bm{q}=\nabla_{\bm{z}}\varphi$ an isometric isomorphism onto
$\mathscr T_\mu$. Thus $\|\varphi\|_{\mathbb V_D}=\|\bm{q}\|_{\mathscr T_\mu}$,
and taking the supremum over $\varphi\in\mathbb V_D$ is equivalent to
taking it over all $\bm{q}\in\mathscr T_\mu$. Completing the square gives
\[
 \langle \bm{h},\bm{q}\rangle_{\mathscr T_\mu}-\tfrac12\|\bm{q}\|_{\mathscr T_\mu}^2
 =\tfrac12\|\bm{h}\|_{\mathscr T_\mu}^2
  -\tfrac12\|\bm{q}-\bm{h}\|_{\mathscr T_\mu}^2.
\]
The maximum is attained at $\bm{q}=\bm{h}$ and equals
$\tfrac12\|\bm{h}\|_{\mathscr T_\mu}^2=\mathcal J_D(\bm{A})$ by
\Eqref{eq:J-geometry-app-v6}, proving \Eqref{eq:J-dual-app-v6}.

Define
$\mathsf S_D \bm{B}=[t\mapsto P_{\mu_t}(\bm{B}\bm{z})]\in\mathscr T_\mu.$
This is a bounded linear map because
$\|\mathsf S_D\bm{B}\|_{\mathscr T_\mu}^2
  \leq\|\bm{B}\|_F^2\int_0^T\int\|\bm{z}\|^2\mathrm{d}\mu_t\mathrm{d}t.$
Its range is finite dimensional and closed, and
$\mathcal J_D(\bm{A})=\tfrac12\|\bm{v}-\mathsf S_D\bm{A}\|_{\mathscr T_\mu}^2$.
Projecting $\bm{v}$ onto this range therefore gives a minimizing matrix
$\bm{A}_D^\dagger$, whose projected velocity $\mathsf S_D\bm{A}_D^\dagger$ is unique.
Orthogonality to every $\mathsf S_D\bm{B}$ gives
\begin{equation*}
  \int_0^T\int
  (\bm{v}_t-P_{\mu_t}(\bm{A}_D^\dagger \bm{z}))^\top P_{\mu_t}(\bm{B}\bm{z})\mathrm{d}\mu_t\mathrm{d}t=0.
\end{equation*}
The first factor is tangent, so $P_{\mu_t}(\bm{B}\bm{z})$ may be replaced by $\bm{B}\bm{z}$.
Since the last identity holds for every $\bm{B}$, it is equivalent to
\Eqref{eq:normal-matrix-v6}.  The Pythagorean theorem gives
\Eqref{eq:pythagorean-v6}; injectivity in
\Eqref{eq:fixedD-identifiability-v6} makes the matrix representative unique.
\end{proof}

\subsection{Finite tests, closed-form regression, and convergence under test enrichment}
\label{app:finite-tests-v6}

Let
\begin{equation}\label{eq:nested-tests-v6}
  \mathbb V_{D,1}\subset\mathbb V_{D,2}\subset\cdots\subset\mathbb V_D,
  \qquad
  \overline{\bigcup_{m\geq1}\mathbb V_{D,m}}=\mathbb V_D,
\end{equation}
where
$\mathbb V_{D,m}=\operatorname{span}\{\varphi_1,\ldots,\varphi_m\}$. 
We choose the test potentials $\varphi_j$ from the class $\mathscr D$
defined above; such a dense family exists by
Lemma~\ref{lem:density-gradients-v6}.

Define
\begin{align}
  y_j={}&\mu_T[\varphi_j(T)]-\mu_0[\varphi_j(0)]
  -\int_0^T\mu_t[\partial_t\varphi_j(t)]\mathrm{d}t,
  \label{eq:y-finite-v6}\\
  \bm{H}_j={}&\int_0^T
  \mu_t[\nabla_{\bm{z}}\varphi_j(t,\bm{z})\bm{z}^\top]\mathrm{d}t,
  \label{eq:H-finite-v6}\\
  M_{ij}={}&\int_0^T
  \mu_t[\nabla_{\bm{z}}\varphi_i(t,\bm{z})^\top\nabla_{\bm{z}}\varphi_j(t,\bm{z})]\mathrm{d}t.
  \label{eq:M-finite-v6}
\end{align}
Let $\bm{a}=\operatorname{vec}(\bm{A})$, let row $j$ of $\bm{G}$ be
$\operatorname{vec}(\bm{H}_j)^\top$, and set $\bm{y}=(y_1,\ldots,y_m)^\top$.

\begin{lemma}
\label{lem:finite-dual-v6}
If $\bm{M}\succ0$, then
\begin{equation}\label{eq:finite-dual-v6}
  \sup_{\varphi\in\mathbb V_{D,m}}
  \left\{\mathcal R_D(\bm{A};\varphi)-\frac{1}{2}\|\varphi\|_{\mathbb V_D}^2\right\}
  =\frac{1}{2}(\bm{y}-\bm{G}\bm{a})^\top \bm{M}^{-1}(\bm{y}-\bm{G}\bm{a}).
\end{equation}
The maximizing test is
\begin{equation}\label{eq:max-test-v6}
  \varphi_{\bm{A},m}^\star
  =\sum_{j=1}^m[\bm{M}^{-1}(\bm{y}-\bm{G}\bm{a})]_j\varphi_j.
\end{equation}
If $\bm{G}$ has full column rank, the unique matrix minimizer is
\begin{equation}\label{eq:closed-form-app-v6}
  \operatorname{vec}(\bm{A}_{D,m})
  =(\bm{G}^\top \bm{M}^{-1}\bm{G})^{-1}\bm{G}^\top \bm{M}^{-1}\bm{y}.
\end{equation}
\end{lemma}

\begin{proof}
Write $\varphi=\sum_{j=1}^m c_j\varphi_j$, with
$\bm{c}=(c_1,\ldots,c_m)^\top$. For each basis function, the definitions of
$y_j$ and $\bm{H}_j$ give
\begin{align*}
  \mathcal R_D(\bm{A};\varphi_j)
  &=
  y_j-\int_0^T
  \mu_t\!\left[\nabla_{\bm{z}}\varphi_j(t,\bm{z})^\top \bm{A}\bm{z}\right]\mathrm{d}t\\
  &=
  y_j-\langle \bm{H}_j,\bm{A}\rangle_F
  =
  y_j-\operatorname{vec}(\bm{H}_j)^\top \bm{a}
  =
  y_j-(\bm{G}\bm{a})_j.
\end{align*}
Therefore, by linearity of $\mathcal R_D(\bm{A};\cdot)$,
\begin{equation*}
  \mathcal R_D(\bm{A};\varphi)
  =
  \sum_{j=1}^m c_j\mathcal R_D(\bm{A};\varphi_j)
  =
  \bm{c}^\top(\bm{y}-\bm{G}\bm{a}).
\end{equation*}
Moreover, since $\bm{M}$ is the Gram matrix of
$\{\varphi_j\}_{j=1}^m$ under the $\mathbb V_D$ inner product,
\begin{align*}
  \|\varphi\|_{\mathbb V_D}^2
  =
  \sum_{i,j=1}^m c_i c_j
  \int_0^T
  \mu_t\!\left[
    \nabla_{\bm{z}}\varphi_i(t,\bm{z})^\top
    \nabla_{\bm{z}}\varphi_j(t,\bm{z})
  \right]\mathrm{d}t =\bm{c}^\top \bm{M}\bm{c}.
\end{align*}
Completing the square gives
\begin{align*}
  \bm{c}^\top(\bm{y}-\bm{G}\bm{a})-\frac{1}{2}\bm{c}^\top \bm{M}\bm{c}
  ={}&-\frac{1}{2}
  (\bm{c}-\bm{M}^{-1}(\bm{y}-\bm{G}\bm{a}))^\top \bm{M}(\bm{c}-\bm{M}^{-1}(\bm{y}-\bm{G}\bm{a}))\\
  &+\frac{1}{2}(\bm{y}-\bm{G}\bm{a})^\top \bm{M}^{-1}(\bm{y}-\bm{G}\bm{a}).
\end{align*}
The first term is nonpositive and vanishes exactly at the coefficients in
\Eqref{eq:max-test-v6}, proving \Eqref{eq:finite-dual-v6}.  Differentiating
the remaining quadratic in $\bm{a}$ gives
$\bm{G}^\top \bm{M}^{-1}\bm{G}\bm{a}=\bm{G}^\top \bm{M}^{-1}\bm{y}$.  Full column rank makes the normal matrix
positive definite and yields \Eqref{eq:closed-form-app-v6}.
\end{proof}

\begin{remark*}[Weak population residuals as regression residuals]
\phantomsection\label{rem:weak-residual-regression}
For fixed $\bm{A}$, $\mathcal R_D(\bm{A};\cdot)$ is the continuous linear
functional obtained by testing the continuity-equation residual $r_{\bm{A}}$
in \Eqref{eq:residual-app-v6}.
Its value $\mathcal R_D(\bm{A};\varphi_j)$ measures the signed discrepancy
between the moment evolution of $\mu_t$ and the contribution of the
candidate drift $\bm{A}\bm{z}$, as expressed in \Eqref{eq:residual-pair-app-v6}.
With the moments defined above,
\[
 \mathcal R_D(\bm{A};\varphi_j)
 =y_j-\langle \bm{H}_j,\bm{A}\rangle_F
 =y_j-(\bm{G}\bm{a})_j,\qquad j=1,\ldots,m.
\]
Thus $\bm{y}-\bm{G}\bm{a}$ collects these scalar weak residuals, and $\bm{y}=\bm{G}\bm{a}$ means that
the continuity equation is satisfied against every test in
$\mathbb V_{D,m}$.
When $\bm{M}\succ0$, Lemma~\ref{lem:finite-dual-v6} gives the weighted
least-squares objective $\tfrac12(\bm{y}-\bm{G}\bm{a})^\top \bm{M}^{-1}(\bm{y}-\bm{G}\bm{a})$ for
estimating $\bm{A}$.
In Appendix~\ref{app:linear-weak-regression}, we implement this regression by
approximating $\bm{y}$, $\bm{G}$, and $\bm{M}$ with empirical sample averages and temporal
quadrature.
\end{remark*}

Lemma~\ref{lem:finite-dual-v6} gives the estimator for a fixed finite test
space. We next show that, as the nested test spaces become dense in
$\mathbb V_D$, these estimators converge to the unique minimizer of the
full weak population residual objective. To compare the two objectives,
we use the Riesz representation of the continuous functional
$\mathcal R_D(\bm{A};\cdot)$: for each $\bm{A}$, there is a unique
$h_{\bm{A}}\in\mathbb V_D$ satisfying
$\mathcal R_D(\bm{A};\varphi)=\langle h_{\bm{A}},\varphi\rangle_{\mathbb V_D}$
for every $\varphi\in\mathbb V_D$.

Writing $\bm{a}=\operatorname{vec}(\bm{A})$ and letting $h_0$ correspond to $\bm{A}=0$,
define the linear operator $\mathsf T_D:\mathbb R^{D^2}\to\mathbb V_D$ by
\begin{equation}\label{eq:T-definition-v6}
  \langle\mathsf T_D\bm{a},\varphi\rangle_{\mathbb V_D}
  =\int_0^T\mu_t[\nabla_{\bm{z}}\varphi(t,\bm{z})^\top \bm{A}\bm{z}]\mathrm{d}t,
  \qquad \varphi\in\mathbb V_D.
\end{equation}
Cauchy--Schwarz and the finite integrated second moment make this
definition well posed by the Riesz theorem, with
$\|\mathsf T_D\bm{a}\|_{\mathbb V_D}
\leq(\int_0^T\mu_t[\|\bm{z}\|^2]\mathrm{d}t)^{1/2}\|\bm{a}\|_2$.
The residual is affine in $\bm{A}$, so $h_{\bm{A}}=h_0-\mathsf T_D\bm{a}$.
Under the gradient identification of
Lemma~\ref{lem:density-gradients-v6},
$\nabla_{\bm{z}} h_0(t,\cdot)=\bm{v}_t$,
$\nabla_{\bm{z}}(\mathsf T_D\bm{a})(t,\cdot)=P_{\mu_t}(\bm{A}\bm{z})$, and thus
$\nabla_{\bm{z}} h_{\bm{A}}(t,\cdot)=\bm{v}_t-P_{\mu_t}(\bm{A}\bm{z})$, with these identities understood in
$\mathscr T_\mu$.

The Riesz representation allows us to complete the square in the dual
objective:
\[
  \mathcal R_D(\bm{A};\varphi)-\tfrac12\|\varphi\|_{\mathbb V_D}^2
  =\tfrac12\|h_{\bm{A}}\|_{\mathbb V_D}^2
   -\tfrac12\|\varphi-h_{\bm{A}}\|_{\mathbb V_D}^2.
\]
The supremum over $\mathbb V_D$ is therefore attained at $\varphi=h_{\bm{A}}$.
For the finite-test problem, let
$Q_m:\mathbb V_D\to\mathbb V_{D,m}$ be the orthogonal projection.
Since $h_{\bm{A}}-Q_mh_{\bm{A}}$ is orthogonal to $\mathbb V_{D,m}$, every
$\varphi\in\mathbb V_{D,m}$ satisfies
$\langle h_{\bm{A}},\varphi\rangle_{\mathbb V_D}
=\langle Q_mh_{\bm{A}},\varphi\rangle_{\mathbb V_D}$.
The same calculation with $Q_mh_{\bm{A}}$ in place of $h_{\bm{A}}$ shows that the
restricted supremum is attained at $\varphi=Q_mh_{\bm{A}}$, which belongs to
$\mathbb V_{D,m}$. Consequently,
\begin{equation}\label{eq:full-finite-objectives-v6}
  \mathcal J_D(\bm{a})=\tfrac12\|h_{\bm{A}}\|_{\mathbb V_D}^2,
  \qquad
  \mathcal J_{D,m}(\bm{a})=\tfrac12\|Q_mh_{\bm{A}}\|_{\mathbb V_D}^2.
\end{equation}
Thus increasing the test space approximates the full residual
representative by its projections $Q_mh_{\bm{A}}$. To compare the minimizers,
write $\bm{C}=\mathsf T_D^*\mathsf T_D$ and $\bm{d}=\mathsf T_D^*h_0$ for the
normal-equation matrix and vector of the full problem, and
$\bm{C}_m=\mathsf T_D^*Q_m\mathsf T_D$ and $\bm{d}_m=\mathsf T_D^*Q_mh_0$ for
their finite-test counterparts.

\begin{theorem}[Convergence under test-space enrichment]
\label{thm:test-enrichment-v6}
Suppose that the nested test spaces satisfy
\Eqref{eq:nested-tests-v6} and that the population minimizer
$\bm{A}_D^\dagger$ is unique. Let $\bm{A}_{D,m}$ minimize, over
$\bm{A}\in\mathbb R^{D\times D}$, the objective obtained by restricting the
dual supremum to $\mathbb V_{D,m}$, and set
$\bm{a}^\dagger=\operatorname{vec}(\bm{A}_D^\dagger)$ and
$\bm{a}_m=\operatorname{vec}(\bm{A}_{D,m})$.
Then $\bm{C}\succ0$, and, for all sufficiently large $m$, $\bm{C}_m\succ0$ and the
finite-test minimizer is unique. For these $m$,
\begin{equation}\label{eq:test-error-v6}
  \|\bm{a}_m-\bm{a}^\dagger\|_2
  \leq
  \frac{2}{\lambda_{\min}(\bm{C})}
  \left(
    \|\bm{d}_m-\bm{d}\|_2
    +
    \|\bm{C}_m-\bm{C}\|_{\mathrm{op}}\|\bm{a}^\dagger\|_2
  \right).
\end{equation}
Consequently,
\begin{equation}\label{eq:test-convergence-v6}
  \bm{A}_{D,m}\longrightarrow \bm{A}_D^\dagger
  \qquad\text{as }m\to\infty.
\end{equation}
\end{theorem}

\begin{proof}
For each $m$, the range of $Q_m\mathsf T_D$ is finite dimensional and
hence closed. Projecting $Q_mh_0$ onto this range therefore yields a
finite-test minimizer.
The quadratic representations above yield the normal equations
\begin{equation}\label{eq:full-finite-normal-v6}
  \bm{C}\bm{a}^\dagger=\bm{d},
  \qquad
  \bm{C}_m\bm{a}_m=\bm{d}_m.
\end{equation}
Density in \Eqref{eq:nested-tests-v6} gives $Q_mq\to q$ for every
$q\in\mathbb V_D$.  Since the domain of $\mathsf T_D$ is finite dimensional,
this convergence is uniform after composition with $\mathsf T_D$.  To see it
directly, let $\bm{e}_1,\ldots,\bm{e}_{D^2}$ be an orthonormal basis of
$\mathbb R^{D^2}$.  For $\bm{a}=\sum_qa_q\bm{e}_q$ with $\|\bm{a}\|_2=1$,
\begin{align*}
  \|(I-Q_m)\mathsf T_D\bm{a}\|
  &\leq\sum_q|a_q|\,\|(I-Q_m)\mathsf T_D\bm{e}_q\|\\
  &\leq D\max_q\|(I-Q_m)\mathsf T_D\bm{e}_q\|\longrightarrow0.
\end{align*}
Consequently,
\begin{align}
  \|\bm{C}_m-\bm{C}\|_{\mathrm{op}}
  &=\|\mathsf T_D^*(Q_m-I)\mathsf T_D\|_{\mathrm{op}}\to0,
  \label{eq:C-conv-v6}\\
  \|\bm{d}_m-\bm{d}\|_2
  &=\|\mathsf T_D^*(Q_m-I)h_0\|_2\to0.
  \label{eq:d-conv-v6}
\end{align}
Uniqueness of $\bm{A}_D^\dagger$ is equivalent to injectivity of $\mathsf T_D$.
Because its domain is finite dimensional,
$\bm{C}=\mathsf T_D^*\mathsf T_D\succ0$.  Put
$\kappa=\lambda_{\min}(\bm{C})$.  For all sufficiently large $m$,
$\|\bm{C}_m-\bm{C}\|\leq\kappa/2$, and Weyl's inequality gives
$\bm{C}_m\succeq(\kappa/2)\bm{I}$.  Thus $\bm{C}_m$ is invertible and
$\|\bm{C}_m^{-1}\|\leq2/\kappa$.

Subtracting the equations in \Eqref{eq:full-finite-normal-v6} yields
\begin{equation*}
  \bm{C}_m(\bm{a}_m-\bm{a}^\dagger)
  =(\bm{d}_m-\bm{d})-(\bm{C}_m-\bm{C})\bm{a}^\dagger.
\end{equation*}
Multiplying by $\bm{C}_m^{-1}$ proves \Eqref{eq:test-error-v6}.  Its right-hand
side tends to zero by \Eqref{eq:C-conv-v6}--\Eqref{eq:d-conv-v6}, which
proves \Eqref{eq:test-convergence-v6}.
\end{proof}

\subsection{Convergence of the weak population residual}
\label{app:operator-consistency-v6}

Our goal in this section is to prove that as the number of
compatible observables increases, \textcolor{black}{the \emph{minimum  weak population residual}
converges to zero.} This is different from Section~\ref{app:finite-tests-v6}, where the observable dimension $D$
is fixed and only the test space is enlarged. As a beginning, we first define the
observable space and the infinite-dimensional feature coordinates in which its dynamics become linear.

\subsubsection{Preliminaries: Koopman-invariant RKHS and linear feature dynamics}
\label{sec:rkhs-preliminaries}

Koopman operators on reproducing-kernel spaces have been studied from both
operator-theoretic and statistical perspectives, including boundedness,
spectral approximation, and learning from data
\citep{ikeda2022koopman,kostic2022learning}.
From now on, let $\mathcal X$ be a separable metric space with a flow $F_t$,
and let $\mathcal H$ be a separable real RKHS with kernel $k$
\citep{aronszajn1950theory}.
The feature map $\Phi(\bm{x}):=k(\bm{x},\cdot)$ satisfies
$f(\bm{x})=\langle f,\Phi(\bm{x})\rangle_{\mathcal H}$ for every $f\in\mathcal H$;
we assume that $\Phi$ is strongly measurable.
Suppose the Koopman operators $U(t)f=f\circ F_t$ preserve $\mathcal H$
and form a $C_0$-semigroup with generator $L$.

Wendland native spaces provide a concrete example of Koopman invariance:
these RKHSs are norm-equivalent to Sobolev spaces on bounded Lipschitz
domains, and composition preserves them when the state map is a sufficiently
smooth diffeomorphism with bounded derivatives and Jacobian determinant
bounded away from zero \citep{koehne2025linfty}.

Furthermore, write $\mathcal Z=\mathcal H$ for the feature space. Since $\mathcal H$ is
reflexive, $T(t):=U(t)^*$ is also a $C_0$-semigroup, with generator
$\mathsf A=L^*$ on its adjoint domain, possibly unbounded.
The reproducing identity gives
$\langle f,T(t)\Phi(\bm{x})\rangle_{\mathcal H}
=\langle U(t)f,\Phi(\bm{x})\rangle_{\mathcal H}=f(F_t \bm{x})$.
As this holds for every $f\in\mathcal H$, the features evolve linearly, following:
\[
    \Phi(F_t \bm{x})=T(t)\Phi(\bm{x}).
\]

For an orthonormal basis $\{e_j\}_{j\ge1}$ of $\mathcal H$, let $P_D$ be the projector
onto $\mathcal Z_D=\operatorname{span}\{e_1,\ldots,e_D\}$.
Since $\langle e_j,\Phi(\bm{x})\rangle_{\mathcal H}=e_j(\bm{x})$,
\[
    P_D\Phi(\bm{x})=\sum_{j=1}^D e_j(\bm{x})e_j
    \quad\longleftrightarrow\quad
    \bm{g}_D(\bm{x})=\bigl(e_1(\bm{x}),\ldots,e_D(\bm{x})\bigr)^\top.
\]
Thus we may view the  $D$-dimensional encoder as a projection
of the same feature map $\Phi(\bm{x})$. For a random initial state $\bm{X}_0$, define
$\mu_t:=\operatorname{Law}(\Phi(F_t\bm{X}_0))$ and
$\mu_t^D:=(P_D)_\#\mu_t$.
Here $\mu_t$ denotes the full feature distribution, while $\mu_t^D$,
identified with a measure on $\mathbb R^D$, corresponds to the latent
distribution denoted by $\mu_t$ in Section~\ref{sec:linear-comparison}
and Sections~\ref{app:wc-geometry-v6}--\ref{app:finite-tests-v6}.
The superscript makes its dependence on the observable dimension explicit.
These projected laws can be compared in the common space $\mathcal Z$
as $D$ increases, with $T(t)$ providing the exact linear evolution for
the subsequent convergence analysis.

\subsubsection{Convergence of the weak population residual}
\label{sec:weak-residual-convergence}

Fix $\tau>0$ and $\bm{X}_0\sim\rho_0$ on $(\Omega,\mathbb P)$. For
$t\in[0,2\tau]$, set $\bm{Z}_0=\Phi(\bm{X}_0)$, $\bm{Z}_t=T(t)\bm{Z}_0=\Phi(F_t\bm{X}_0)$,
$\mu_t=\operatorname{Law}(\bm{Z}_t)$, and $\mu_t^D=(P_D)_\#\mu_t$.
Assume that $\bm{Z}_0\in D(\mathsf A)$ almost surely and
$\mathbb E(\|\bm{Z}_0\|^2+\|\mathsf A\bm{Z}_0\|^2)<\infty$.
We write $\|\cdot\|_{L^2}$ for the norm on $L^2(\Omega;\mathcal Z)$ and
$\|\cdot\|_{\mathbb L}$ for the norm on
$\mathbb L=L^2([0,2\tau]\times\Omega;\mathcal Z)$.

The semigroup preserves $D(\mathsf A)$ and satisfies
$\mathsf AT(t)\bm{z}=T(t)\mathsf A\bm{z}$ and
$T(t)\bm{z}-\bm{z}=\int_0^tT(r)\mathsf A\bm{z}\,\mathrm{d}r$ for $\bm{z}\in D(\mathsf A)$
\citep[Chapter II]{engel2000semigroups}.
Strong continuity and uniform boundedness give
$M_\tau:=\sup_{0\le t\le2\tau}\|T(t)\|<\infty$, so
\[
 \int_0^{2\tau}\mathbb E(\|\bm{Z}_t\|^2+\|\mathsf A\bm{Z}_t\|^2)\,\mathrm{d}t
 \le 2\tau M_\tau^2\mathbb E(\|\bm{Z}_0\|^2+\|\mathsf A\bm{Z}_0\|^2)<\infty.
\]
The same bound and dominated convergence show that
$t\mapsto \bm{Z}_t$ is continuous in $L^2(\Omega;\mathcal Z)$.
For $s\le t$, coupling $P_D\bm{Z}_s$ and $P_D\bm{Z}_t$ and using
$\bm{Z}_t-\bm{Z}_s=\int_s^t\mathsf A\bm{Z}_r\,\mathrm{d}r$ yields
\[
 W_2(\mu_s^D,\mu_t^D)
 \le\|P_D(\bm{Z}_t-\bm{Z}_s)\|_{L^2}
 \le\int_s^t\|\mathsf A\bm{Z}_r\|_{L^2}\,\mathrm{d}r.
\]
Since the last integrand belongs to $L^2(0,2\tau)$,
$\mu_\cdot^D\in AC^2([0,2\tau];\mathcal P_2(\mathcal Z_D))$.
The weak-continuity framework of the preceding subsections therefore
applies, with weak population objective
\begin{equation}\label{eq:JD-v9}
 \mathcal J_D(\bm{B})=\frac12\int_0^{2\tau}
 \|\partial_t\mu_t^D+\nabla\cdot(\bm{B}\bm{y}\,\mu_t^D)\|_{-1,\mu_t^D}^2\,\mathrm{d}t.
\end{equation}
Theorem~\ref{thm:dual-geometry-v6} ensures that a minimizer
$\bm{A}_D\in\operatorname*{argmin}_{\bm{B}\in\mathcal L(\mathcal Z_D)}\mathcal J_D(\bm{B})$
exists; uniqueness is unnecessary for the statements below. 

\begin{lemma}[Comparison with the projected path velocity]
\label{lem:velocity-comparison-v9}
Under the assumptions above, every $\bm{B}\in\mathcal L(\mathcal Z_D)$ satisfies
\[
 \sqrt{2\mathcal J_D(\bm{B})}\le\|P_D\mathsf A\bm{Z}-\bm{B}P_D\bm{Z}\|_{\mathbb L}.
\]
\end{lemma}
\begin{proof}
Set $\bm{Y}_t=P_D\bm{Z}_t$ and $\bm{V}_t=P_D\mathsf A\bm{Z}_t$. Conditioning $\bm{V}$ on $(t,\bm{Y}_t)$
under the probability measure $(2\tau)^{-1}\mathrm{d}t\,\mathrm{d}\mathbb P$ gives a jointly
measurable field $\bm{w}_t^D(\bm{y})=\mathbb E[\bm{V}_t\mid \bm{Y}_t=\bm{y}]$ for almost every $t$.
Conditional Jensen's inequality gives
$\|\bm{w}^D(\bm{Y})\|_{\mathbb L}\le\|\bm{V}\|_{\mathbb L}<\infty$.
For $\psi\in C_c^\infty(\mathcal Z_D)$, the chain rule along $\bm{Y}_t$ gives
\[
 \mathbb E\psi(\bm{Y}_t)-\mathbb E\psi(\bm{Y}_s)
 =\int_s^t\int\langle\nabla\psi(\bm{y}),\bm{w}_r^D(\bm{y})\rangle\,\mathrm{d}\mu_r^D(\bm{y})\,\mathrm{d}r.
\]
Here Fubini's theorem applies because $\nabla\psi$ is bounded and
$\bm{V}\in\mathbb L$. Thus $\bm{w}^D$ satisfies the projected continuity equation
$\partial_t\mu_t^D+\nabla_{\bm{y}}\cdot(\bm{w}_t^D\mu_t^D)=0$
in the sense of distributions on $(0,2\tau)\times\mathcal Z_D$.
Writing $P_{\mu_t^D}$ for the orthogonal projection onto its Wasserstein
tangent space, the canonical velocity is $P_{\mu_t^D}\bm{w}_t^D$.
Theorem~\ref{thm:dual-geometry-v6} and contraction of this projection imply
\[
 2\mathcal J_D(\bm{B})
 =\int_0^{2\tau}\|P_{\mu_t^D}(\bm{w}_t^D-\bm{B}\bm{y})\|_{L^2(\mu_t^D)}^2\,\mathrm{d}t
 \le\|\bm{w}^D(\bm{Y})-\bm{B}\bm{Y}\|_{\mathbb L}^2.
\]
Since $\bm{w}_t^D(\bm{Y}_t)-\bm{B}\bm{Y}_t=\mathbb E[\bm{V}_t-\bm{B}\bm{Y}_t\mid \bm{Y}_t]$, another application
of conditional Jensen gives
$\|\bm{w}^D(\bm{Y})-\bm{B}\bm{Y}\|_{\mathbb L}\le\|\bm{V}-\bm{B}\bm{Y}\|_{\mathbb L}$, proving the claim.
\end{proof}

To approximate the projected velocity, we use the bounded difference
quotient $C_h=(T(h)-I)/h$ before projecting onto $\mathcal Z_D$.
This avoids applying the unbounded generator to $P_D\bm{Z}_t$.

\begin{theorem}[Convergence of the weak population residual]
\label{thm:consistency-v9}
Under the assumptions above, every choice of weak population minimizers $\bm{A}_D$
satisfies $\mathcal J_D(\bm{A}_D)\to0$ as $D\to\infty$. More precisely, define
$a(h)=\|(\mathsf A-C_h)\bm{Z}\|_{\mathbb L}$ and
$b_D=\|(I-P_D)\bm{Z}\|_{\mathbb L}$. Then $a(h)\to0$ as $h\downarrow0$,
$b_D\to0$ as $D\to\infty$, and
\[
 \sqrt{2\mathcal J_D(\bm{A}_D)}\le a(h)+\|C_h\|b_D\qquad(h>0).
\]
\end{theorem}
\begin{proof}
For fixed $h>0$, $C_h$ is bounded on $\mathcal Z$, so
$\bm{B}_{D,h}=P_DC_h|_{\mathcal Z_D}$ is an admissible comparison matrix.
The velocity error decomposes as
\[
 P_D\mathsf A\bm{Z}-P_DC_hP_D\bm{Z}
 =P_D(\mathsf A-C_h)\bm{Z}+P_DC_h(I-P_D)\bm{Z}.
\]
The preceding lemma, the triangle inequality in $\mathbb L$, and
$\|P_D\|\le1$ therefore give
\[
 \sqrt{2\mathcal J_D(\bm{A}_D)}
 \le\sqrt{2\mathcal J_D(\bm{B}_{D,h})}
 \le a(h)+\|C_h\|b_D.
\]
Strong convergence of $P_D$ gives $(I-P_D)\bm{Z}_t\to0$ pointwise, while
$\|(I-P_D)\bm{Z}_t\|^2\le\|\bm{Z}_t\|^2$ is integrable over time and probability.
Dominated convergence yields $b_D\to0$.

For $\bm{z}\in D(\mathsf A)$,
$C_h\bm{z}=h^{-1}\int_0^hT(r)\mathsf A\bm{z}\,\mathrm{d}r\to\mathsf A\bm{z}$ by strong
continuity. Fix $h_0>0$ and let
$M_0=\sup_{0\le r\le h_0}\|T(r)\|<\infty$.
For $0<h\le h_0$, $\|C_h\bm{z}\|\le M_0\|\mathsf A\bm{z}\|$, hence
$\|(\mathsf A-C_h)\bm{Z}_t\|^2\le(1+M_0)^2\|\mathsf A\bm{Z}_t\|^2$.
The finite orbit energy established above makes this bound integrable,
so dominated convergence gives $a(h)\to0$.

The bound on the minimum now proves the result by first fixing $h$ and
then increasing $D$. Specifically, for any $\varepsilon>0$, choose $h$
with $a(h)<\varepsilon/2$, and then choose $D$ sufficiently large that
$b_D<\varepsilon/[2(1+\|C_h\|)]$.
It follows that $\sqrt{2\mathcal J_D(\bm{A}_D)}<\varepsilon$.
\end{proof}

\subsection{From weak residual consistency to strong semigroup convergence}
\label{app:koopman-consequences-v6}\label{sec:D-v9}

Section~\ref{sec:weak-residual-convergence} establishes
$\mathcal J_D(\bm{A}_D)\to0$ for the weak population minimizers $\bm{A}_D$. In this section,
under a uniform stability bound on $e^{t\bm{A}_D}$, we first show that this
vanishing weak residual implies uniform Wasserstein convergence of the
predicted distributions on the training interval.
The subsequent subsections establish the stronger conclusion of
semigroup convergence at every initial feature vector: compactness
provides strong operator subsequences, and finite-time linear determinacy
(FT) identifies their limits with $T(t)$.

\begin{assumption}[Finite-time stability of weak population minimizers]
\label{ass:stability-v9}
For the selected minimizers
$\bm{A}_D\in\operatorname*{argmin}_{\bm{B}\in\mathcal L(\mathcal Z_D)}\mathcal J_D(\bm{B})$,
there exists $M\ge1$, independent of $D$, such that
\[
 \sup_D\sup_{0\le t\le\tau}
 \|e^{t\bm{A}_D}\|_{\mathcal L(\mathcal Z_D)}\le M.
\]
\end{assumption}
This assumption bounds the finite-time evolution generated by the
selected minimizers. It is an additional hypothesis beyond convergence
of the weak residual and does not require a uniform bound on $\|\bm{A}_D\|$.

\subsubsection{The weak residual controls probability prediction}
\label{sec:weak-residual-prediction}

The weak residual measures the minimum correction to a candidate linear
velocity needed to transport a prescribed distribution curve.
Propagating this correction gives a prediction bound on any subinterval.

\begin{lemma}[The weak residual controls finite-time prediction]
\label{lem:rollout-v9}
Let $\alpha_\cdot\in AC^2([s,t];\mathcal P_2(\mathbb R^d))$ and
$\bm{B}\in\mathbb R^{d\times d}$. For $r\in[s,t]$, write
$r_{\bm{B}}(r)=\partial_r\alpha_r+\nabla_{\bm{y}}\cdot(\bm{B}\bm{y}\,\alpha_r)$. Then
\[
 W_2((e^{(t-s)\bm{B}})_\#\alpha_s,\alpha_t)
 \le\int_s^t\|e^{(t-r)\bm{B}}\|\,\|r_{\bm{B}}(r)\|_{-1,\alpha_r}\,\mathrm{d}r.
\]
\end{lemma}
\begin{proof}
Let $\bm{v}_r$ be the canonical velocity of $\alpha_r$ and set
$\bm{u}_r=\bm{v}_r-P_{\alpha_r}(\bm{B}\bm{y})$, where $P_{\alpha_r}$ is the orthogonal
projection onto the Wasserstein tangent space.
Lemma~\ref{lem:min-flux-v6} and Theorem~\ref{thm:dual-geometry-v6} give
$-\nabla_{\bm{y}}\cdot(\bm{u}_r\alpha_r)=r_{\bm{B}}(r)$ and
$\|\bm{u}_r\|_{L^2(\alpha_r)}=\|r_{\bm{B}}(r)\|_{-1,\alpha_r}$ for almost every $r$.
Hence $\partial_r\alpha_r+\nabla_{\bm{y}}\cdot((\bm{B}\bm{y}+\bm{u}_r)\alpha_r)=0$
in the sense of distributions.
Choose a jointly Borel representative of $\bm{u}$.
The corrected velocity $\bm{B}\bm{y}+\bm{u}_r$ has finite squared energy because
$\bm{v}\in L^2(\mathrm{d}r\,\mathrm{d}\alpha_r)$, tangent projection is contractive, and
$\alpha_\cdot$ has bounded second moments on $[s,t]$.

The superposition principle therefore gives a probability measure on
absolutely continuous paths $\bm{\gamma}$ with marginals $\alpha_r$ and
$\dot{\bm{\gamma}}_r=\bm{B}\bm{\gamma}_r+\bm{u}_r(\bm{\gamma}_r)$ almost everywhere
\citep[Theorem 8.2.1]{ambrosioGradientFlowsMetric2008}.
Variation of constants yields
\[
 \bm{\gamma}_t-e^{(t-s)\bm{B}}\bm{\gamma}_s
 =\int_s^t e^{(t-r)\bm{B}}\bm{u}_r(\bm{\gamma}_r)\,\mathrm{d}r.
\]
The pair $(e^{(t-s)\bm{B}}\bm{\gamma}_s,\bm{\gamma}_t)$ couples the two distributions
in the claim. Minkowski's integral inequality and the marginal identity
$\operatorname{Law}(\bm{\gamma}_r)=\alpha_r$ give
\[
 \left(\mathbb E\|\bm{\gamma}_t-e^{(t-s)\bm{B}}\bm{\gamma}_s\|^2\right)^{1/2}
 \le\int_s^t\|e^{(t-r)\bm{B}}\|\,\|\bm{u}_r\|_{L^2(\alpha_r)}\,\mathrm{d}r.
\]
Substituting the weak residual norm proves the bound.
\end{proof}

To compare the learned linear propagator in the common space $\mathcal Z$, define
\[
 K_D(t)=\iota_De^{t\bm{A}_D}P_D,\qquad S_D(t)=K_D(t)+(I-P_D),
\]
where $\iota_D:\mathcal Z_D\hookrightarrow\mathcal Z$ is inclusion.
$K_D(t)$ evolves the projected features, while $S_D(t)$
also retains the omitted unresolved feature space.
On $\mathcal Z_D\oplus\mathcal Z_D^\perp$, $S_D(t)$ acts as
$e^{t\bm{A}_D}\oplus I$, so it is a $C_0$-semigroup with $S_D(0)=I$ and
$\|S_D(t)\|\le\max\{\|e^{t\bm{A}_D}\|,1\}\le M$ for $0\le t\le\tau$.
We use $S_D$ only for the subsequent operator-convergence argument;
$K_D$ remains the propagator over the finite feature subspace, with $K_D(0)=P_D$.

The following result combines the weak residual bound with the error
from projecting $\bm{Z}_t$. It is uniform in both the starting time and the
prediction lag, as needed later to compare operators on all source distribution laws.

\begin{theorem}
\label{thm:flow-v9}
Under Theorem~\ref{thm:consistency-v9} and
Assumption~\ref{ass:stability-v9}, let
$q_D=\sup_{0\le r\le2\tau}\|(I-P_D)\bm{Z}_r\|_{L^2}$, with the $L^2$ norm
defined in Section~\ref{sec:weak-residual-convergence}.
Then $q_D\to0$ and
\begin{align*}
 \sup_{s,t\in[0,\tau]}W_2^{\mathcal Z}(K_D(t)_\#\mu_s,\mu_{s+t})
 &\le q_D+M\sqrt{2\tau\mathcal J_D(\bm{A}_D)}\longrightarrow0,\\
 \sup_{s,t\in[0,\tau]}W_2^{\mathcal Z}(S_D(t)_\#\mu_s,\mu_{s+t})
 &\le2q_D+M\sqrt{2\tau\mathcal J_D(\bm{A}_D)}\longrightarrow0.
\end{align*}
Prediction from the initial law over the full training interval satisfies
\[
 \sup_{0\le t\le2\tau}W_2^{\mathcal Z}(K_D(t)_\#\mu_0,\mu_t)
 \le q_D+M^2\sqrt{4\tau\mathcal J_D(\bm{A}_D)}\longrightarrow0.
\]
\end{theorem}
\begin{proof}
Strong convergence of $P_D$ and dominated convergence imply
$\|(I-P_D)\bm{V}\|_{L^2}\to0$ for every $\bm{V}\in L^2(\Omega;\mathcal Z)$.
Since $r\mapsto \bm{Z}_r$ is continuous in this space, its image on
$[0,2\tau]$ is compact. For any $\varepsilon>0$, choose a finite
$\varepsilon$-net $\{\bm{Z}_{r_j}\}_{j=1}^m$ of this image.
Contraction of $I-P_D$ gives
$q_D\le\varepsilon+\max_j\|(I-P_D)\bm{Z}_{r_j}\|_{L^2}$;
letting $D\to\infty$ and then $\varepsilon\downarrow0$ proves $q_D\to0$.

For $s,t\in[0,\tau]$, apply Lemma~\ref{lem:rollout-v9} to
$\mu_\cdot^D$ on $[s,s+t]\subset[0,2\tau]$.
With $r_{\bm{A}_D}(r)=\partial_r\mu_r^D+\nabla_{\bm{y}}\cdot(\bm{A}_D\bm{y}\,\mu_r^D)$,
stability and Cauchy--Schwarz yield
\[
 W_2((e^{t\bm{A}_D})_\#\mu_s^D,\mu_{s+t}^D)
 \le M\int_s^{s+t}\|r_{\bm{A}_D}(r)\|_{-1,\mu_r^D}\,\mathrm{d}r
 \le M\sqrt{2t\mathcal J_D(\bm{A}_D)}.
\]
The last inequality uses
$\int_0^{2\tau}\|r_{\bm{A}_D}(r)\|_{-1,\mu_r^D}^2\,\mathrm{d}r=2\mathcal J_D(\bm{A}_D)$.
Isometric inclusion into $\mathcal Z$ preserves this Wasserstein distance,
and coupling $P_D\bm{Z}_{s+t}$ with $\bm{Z}_{s+t}$ adds at most $q_D$.
This proves the bound for $K_D$ since $t\le\tau$.
Likewise, $S_D(t)\bm{Z}_s-K_D(t)\bm{Z}_s=(I-P_D)\bm{Z}_s$, so the coupling induced by
$\bm{Z}_s$ adds at most another $q_D$, proving the bound for $S_D$.

For $0\le r\le2\tau$, the semigroup law gives
$\|e^{r\bm{A}_D}\|=\|(e^{(r/2)\bm{A}_D})^2\|\le M^2$.
Applying the same argument on $[0,t]$ therefore bounds the full-interval
prediction error by $q_D+M^2\sqrt{2t\mathcal J_D(\bm{A}_D)}$, which is at
most the stated bound for $t\le2\tau$.
All three bounds vanish because $q_D\to0$ and
$\mathcal J_D(\bm{A}_D)\to0$ by Theorem~\ref{thm:consistency-v9}.
\end{proof}

\subsubsection{Distributional identifiability of the linear propagator}

Theorem~\ref{thm:flow-v9} shows that a vanishing weak residual yields
convergent distribution predictions. To establish convergence of the
linear propagators themselves, we also need the distribution dynamics
to identify the underlying map. This is a fundamental issue when
inferring dynamics from unpaired snapshots: distinct trajectories can
produce the same distribution at every time.
For linear SDEs with additive noise in finite dimensions,
\citet{guan2024snapshots} characterize the obstruction to identifiability
through generalized rotational symmetries of the initial distribution.
For example, a rotational flow and a stationary flow preserve the same
isotropic Gaussian distribution, despite moving individual states
differently.

In our setting, the subsequent operator-convergence argument therefore
requires the family $\{\mu_s:0\le s\le\tau\}$ to determine $T(t)$
uniquely for each lag $t\in[0,\tau]$. We impose this requirement directly
on bounded linear maps on $\mathcal Z$ through the following assumption.

\begin{assumption}[Finite-time linear determinacy (FT)]
\label{ass:FT-v9}
For every lag $t\in[0,\tau]$ and every bounded linear map
$B:\mathcal Z\to\mathcal Z$,
\begin{equation}\label{eq:FT-v9}
 B_\#\mu_s=\mu_{s+t}\quad\text{for all }s\in[0,\tau]
 \qquad\Longrightarrow\qquad B=T(t).
\end{equation}
\end{assumption}

For a fixed lag, FT identifies the propagator from its action on the
entire family of source distributions. At $t=0$, it rules out nonidentity
linear symmetries shared by all these distributions. Allowing every
bounded linear candidate $B$, including noninvertible ones, ensures that
the condition applies to the operator limits considered below.
FT supplies uniqueness; stability and compactness will provide the
remaining ingredients for convergence. The next proposition relates
FT to the linear span of the distribution supports and gives a sufficient
condition in terms of their means.

\begin{proposition}[Total linear support and a sufficient criterion for FT]
\label{prop:FT-meaning-v9}
Let $\bar\mu=\tau^{-1}\int_0^\tau\mu_s\,\mathrm{d}s$. Under FT,
\begin{equation}\label{eq:total-support-v9}
 \overline{\operatorname{span}(\operatorname{supp}\bar\mu)}=\mathcal Z.
\end{equation}
A sufficient condition for FT is
\begin{equation}\label{eq:total-mean-v9}
 \overline{\operatorname{span}\{\bm{m}_s:0\le s\le\tau\}}=\mathcal Z,
 \qquad \bm{m}_s=\int \bm{z}\,\mathrm{d}\mu_s(\bm{z}).
\end{equation}
\end{proposition}
\begin{proof}
If \Eqref{eq:total-support-v9} failed, there would be a nonzero $\bm{h}$ with
$\langle \bm{h},\bm{z}\rangle=0$ for $\bar\mu$-almost every $\bm{z}$. Consequently
\[
 \int_0^\tau\mathbb E|\langle \bm{h},\bm{Z}_s\rangle|^2\mathrm{d}s=0.
\]
The integrand is nonnegative and continuous in $s$, because $\bm{Z}_s$ is
$L^2$-continuous. It is therefore zero for every $s\in[0,\tau]$.
The bounded linear map $B\bm{z}=\bm{z}+\langle \bm{h},\bm{z}\rangle \bm{h}$ fixes $\bm{Z}_s$ almost
surely for each $s$, and hence $B_\#\mu_s=\mu_s$ for all $s$. But $B\ne I$,
contradicting FT at $t=0$. This proves the first claim.

For the sufficient condition, take $B$ satisfying the premise of FT for a
fixed lag $t$. Finite second moments allow Bochner expectations, and
bounded linear maps commute with them. Thus
\[
 B\bm{m}_s=\bm{m}_{s+t}=T(t)\bm{m}_s\qquad(0\le s\le\tau).
\]
The bounded operator $B-T(t)$ vanishes on the dense linear span in
\Eqref{eq:total-mean-v9}, so it vanishes everywhere. This proves FT.
\end{proof}

\subsubsection{From distributional compactness to strong operator compactness}
\label{sec:distributional-compactness}

Theorem~\ref{thm:flow-v9} establishes $W_2$ convergence of the pushforward
measures $S_D(t)_\#\mu_s$ to $\mu_{s+t}$ as the weak residual vanishes.
To obtain convergence of the linear propagators, we next show that
relative compactness of $\{(B_n)_\#\nu\}$ in $W_2$ yields strongly
convergent operator subsequences when $(B_n)$ is uniformly bounded and
$\operatorname{supp}\nu$ has dense linear span in $\mathcal Z$.
In our setting, stability supplies
the bound, and Proposition~\ref{prop:FT-meaning-v9} establishes the support
condition for the mixture $\bar\mu=\tau^{-1}\int_0^\tau\mu_s\,\mathrm{d}s$.
The following lemma provides the required compactness; the next
subsection uses FT to identify each operator limit with the true
propagator $T(t)$.

\begin{lemma}[From $W_2$ compactness to strong operator compactness]
\label{lem:compactness-v9}
Let $\mathcal Z$ be a separable real Hilbert space and let
$\nu\in\mathcal P_2(\mathcal Z)$ satisfy
$\overline{\operatorname{span}(\operatorname{supp}\nu)}=\mathcal Z$.
Suppose $B_n\in\mathcal L(\mathcal Z)$,
$\sup_n\|B_n\|\le R<\infty$, and $\{(B_n)_\#\nu\}$ is relatively compact
in $W_2$. Then every subsequence of $(B_n)$ has a further subsequence and
a bounded linear map $B$, $\|B\|\le R$, such that
\begin{equation}\label{eq:compactness-conclusion-v9}
 \|B_n\bm{z}-B\bm{z}\|\longrightarrow0
 \qquad\text{for every }\bm{z}\in\mathcal Z
\end{equation}
along that further subsequence.
\end{lemma}
\begin{proof}
Start with any subsequence and retain the index $n$.
Choose a countable dense set $\{\bm{q}_j\}$ in $\mathcal Z$.
The uniform operator bound allows a diagonal extraction for which
$\langle B_n\bm{q}_i,\bm{q}_j\rangle$ converges for every $i,j$.
By density and the same bound, these limits extend to a bilinear form
$b(\bm{x},\bm{y})$ with $|b(\bm{x},\bm{y})|\le R\|\bm{x}\|\|\bm{y}\|$.
Riesz representation gives $B\in\mathcal L(\mathcal Z)$ with
$b(\bm{x},\bm{y})=\langle B\bm{x},\bm{y}\rangle$ and $\|B\|\le R$, so $B_n$ converges to
$B$ in the weak operator topology. Extracting once more using the
$W_2$ relative compactness gives
\begin{equation}\label{eq:two-limits-v9}
 B_n\xrightarrow{\mathrm{WOT}}B,\qquad
 (B_n)_\#\nu\xrightarrow{W_2}\lambda
 \quad\text{for some }\lambda\in\mathcal P_2(\mathcal Z).
\end{equation}

For every $\bm{h}\in\mathcal Z$, weak operator convergence and dominated
convergence imply
\[
 \int e^{i\langle \bm{h},\bm{y}\rangle}\,\mathrm{d}\lambda(\bm{y})
 =\lim_n\int e^{i\langle \bm{h},B_n\bm{z}\rangle}\,\mathrm{d}\nu(\bm{z})
 =\int e^{i\langle \bm{h},B\bm{z}\rangle}\,\mathrm{d}\nu(\bm{z}).
\]
The first equality follows from $W_2$ convergence, which implies weak
convergence of the measures $(B_n)_\#\nu$ to $\lambda$.
Thus $\lambda$ and $B_\#\nu$ have the
same characteristic functional. Restricting $\bm{h}$ to each finite span of
an orthonormal basis gives equality of all finite-coordinate
 distributions. These coordinates generate the Borel sigma field of
$\mathcal Z$, since it is separable, and hence
$\lambda=B_\#\nu$.

Convergence in $W_2$ also gives convergence of second moments, so
$\int\|B_n\bm{z}\|^2\,\mathrm{d}\nu(\bm{z})\to\int\|B\bm{z}\|^2\,\mathrm{d}\nu(\bm{z})$.
Indeed, the difference between the square roots of these moments is
bounded by $W_2((B_n)_\#\nu,B_\#\nu)$, by the reverse triangle inequality
in any transport coupling.
For each fixed $\bm{z}$, weak operator convergence gives
$\langle B_n\bm{z},B\bm{z}\rangle\to\|B\bm{z}\|^2$, and
$|\langle B_n\bm{z},B\bm{z}\rangle|\le R^2\|\bm{z}\|^2$ is $\nu$-integrable.
Dominated convergence therefore gives convergence of the corresponding
integrals. Expanding the square yields
\begin{equation}\label{eq:same-coupling-v9}
 \int\|(B_n-B)\bm{z}\|^2\,\mathrm{d}\nu(\bm{z})
 =\int\bigl(\|B_n\bm{z}\|^2+\|B\bm{z}\|^2-2\langle B_n\bm{z},B\bm{z}\rangle\bigr)\,\mathrm{d}\nu(\bm{z})
 \longrightarrow0.
\end{equation}

To deduce pointwise convergence on $\operatorname{supp}\nu$ from
convergence in $L^2(\nu;\mathcal Z)$, fix
$\bm{z}\in\operatorname{supp}\nu$ and $\delta>0$. The open ball
$\mathbb B_\delta(\bm{z})$ has positive $\nu$-measure.
For $\bm{y}$ in this ball,
$\|(B_n-B)\bm{z}\|\le2R\delta+\|(B_n-B)\bm{y}\|$.
Averaging over the ball and applying Cauchy--Schwarz gives
\[
 \|(B_n-B)\bm{z}\|
 \le2R\delta+
 \frac{\left(\int\|(B_n-B)\bm{y}\|^2\,\mathrm{d}\nu(\bm{y})\right)^{1/2}}
 {\nu(\mathbb B_\delta(\bm{z}))^{1/2}}.
\]
Letting $n\to\infty$ and then $\delta\downarrow0$ proves convergence
at every point of $\operatorname{supp}\nu$ along the same subsequence.
Linearity extends it to the span of this support, which is dense by
assumption. Finally, for any $\bm{z}\in\mathcal Z$ and any $\bm{x}$ in this span,
$\|(B_n-B)\bm{z}\|\le2R\|\bm{z}-\bm{x}\|+\|(B_n-B)\bm{x}\|$.
Approximating $\bm{z}$ by such $\bm{x}$ proves
\Eqref{eq:compactness-conclusion-v9}. The starting subsequence was
arbitrary, so the asserted strong operator relative compactness follows.
\end{proof}

In the next subsection, we apply the lemma with $\nu=\bar\mu$ and
$B_n=S_{D_n}(t_n)$. Theorem~\ref{thm:flow-v9} and $W_2$ continuity of
$t\mapsto\mu_t$ establish the required compactness of
$\{(S_{D_n}(t_n))_\#\bar\mu\}$, even when the time lags $t_n$ vary.
Along subsequences with $t_n\to t$, FT identifies the strong operator
limit $B$ through the identities $B_\#\mu_s=\mu_{s+t}$ for all
$s\in[0,\tau]$.

\subsubsection{Locally uniform strong convergence of the learned propagators}
\label{sec:strong-semigroup-convergence}

Theorem~\ref{thm:flow-v9} establishes convergence of the pushforward
measures $S_D(t)_\#\mu_s$, and Lemma~\ref{lem:compactness-v9} converts
$W_2$ compactness of these measures into strong operator subsequences.
We now combine these results with the finite-time linear determinacy
condition in Assumption~\ref{ass:FT-v9} to identify every subsequential
limit with $T(t)$. Allowing the time lags to vary along each subsequence
will establish convergence uniformly on $[0,\tau]$ for every fixed
$\bm{z}\in\mathcal Z$. The semigroup property then extends the result to any
finite time interval.

\begin{theorem}[Locally uniform strong convergence under stability and FT]
\label{thm:semigroup-v9}
In the setting of Sections~\ref{sec:rkhs-preliminaries}
and~\ref{sec:weak-residual-convergence}, assume
$\bm{Z}_0\in D(\mathsf A)$ almost surely and
$\mathbb E(\|\bm{Z}_0\|^2+\|\mathsf A\bm{Z}_0\|^2)<\infty$.
Let $P_D$ be the increasing finite-rank orthogonal projections onto
$\mathcal Z_D$, with $P_D\bm{z}\to \bm{z}$ for every $\bm{z}\in\mathcal Z$, and choose
$\bm{A}_D\in\operatorname*{argmin}_{\bm{B}\in\mathcal L(\mathcal Z_D)}\mathcal J_D(\bm{B})$
for the weak population objective \Eqref{eq:JD-v9}.
Suppose the selected minimizers satisfy the finite-time stability
condition in Assumption~\ref{ass:stability-v9}, and the distribution
family $\{\mu_s\}_{0\le s\le2\tau}$ satisfies FT
(Assumption~\ref{ass:FT-v9}). Then, for every $\bm{z}\in\mathcal Z$ and
$H>0$,
\begin{equation}\label{eq:SOT-v9}
 \sup_{0\le t\le H}\|K_D(t)\bm{z}-T(t)\bm{z}\|\longrightarrow0
 \qquad(D\to\infty).
\end{equation}
The same conclusion holds for $S_D(t)=K_D(t)+(I-P_D)$.
\end{theorem}
\begin{proof}
\emph{Step 1: Wasserstein convergence of the averaged measures.}
Theorem~\ref{thm:flow-v9} gives
\begin{equation}\label{eq:eD-v9}
 e_D:=\sup_{s,t\in[0,\tau]}
 W_2(S_D(t)_\#\mu_s,\mu_{s+t})\longrightarrow0.
\end{equation}
For $t\in[0,\tau]$, define
\[
 \bar\mu=\frac1\tau\int_0^\tau\mu_s\,\mathrm{d}s,\qquad
 \bar\mu_t=\frac1\tau\int_0^\tau\mu_{s+t}\,\mathrm{d}s.
\]
Both measures belong to $\mathcal P_2(\mathcal Z)$ because
$r\mapsto \bm{Z}_r$ is continuous in $L^2(\Omega;\mathcal Z)$ on
$[0,2\tau]$. Proposition~\ref{prop:FT-meaning-v9} gives
$\overline{\operatorname{span}(\operatorname{supp}\bar\mu)}=\mathcal Z$.
For $r,t\in[0,\tau]$, using the same $\bm{Z}_0$ and an independent uniform
time in $[0,\tau]$ to couple the averaged measures yields
\[
 W_2(\bar\mu_t,\bar\mu_r)^2
 \le\frac1\tau\int_0^\tau\|\bm{Z}_{s+t}-\bm{Z}_{s+r}\|_{L^2}^2\,\mathrm{d}s
 \longrightarrow0\qquad(t\to r).
\]
Here the convergence follows from uniform $L^2$ continuity of $\bm{Z}_r$
on $[0,2\tau]$. Averaging the bounds in \Eqref{eq:eD-v9} gives
\begin{equation}\label{eq:mixture-v9}
 W_2(S_D(t)_\#\bar\mu,\bar\mu_t)^2
 \le\frac1\tau\int_0^\tau
 W_2(S_D(t)_\#\mu_s,\mu_{s+t})^2\,\mathrm{d}s
 \le e_D^2.
\end{equation}
To justify this inequality, first approximate each time integral by
a Riemann sum on a uniform partition. Averaging optimal couplings for the
finitely many pairs of measures proves the inequality for these sums.
For fixed $D,t$, the discrete averages converge in $W_2$ to
$S_D(t)_\#\bar\mu$ and $\bar\mu_t$: for $s$ in a partition interval
with left endpoint $s_j$, use the pairs
$(S_D(t)\bm{Z}_s,S_D(t)\bm{Z}_{s_j})$ and $(\bm{Z}_{s+t},\bm{Z}_{s_j+t})$, respectively.
Uniform $L^2$ continuity and boundedness of $S_D(t)$
make the coupling costs vanish. The integrand on the right is continuous
in $s$, since both curves of measures are $W_2$-continuous.
Passing to the limit proves \Eqref{eq:mixture-v9}.

\emph{Step 2: Strong operator subsequences for convergent time lags.}
Take arbitrary sequences $D_n\to\infty$ and $t_n\in[0,\tau]$.
After passing to a subsequence, compactness of $[0,\tau]$ gives
$t_n\to t_*$. By \Eqref{eq:mixture-v9} and continuity of
$t\mapsto\bar\mu_t$,
\[
 W_2((S_{D_n}(t_n))_\#\bar\mu,\bar\mu_{t_*})
 \le e_{D_n}+W_2(\bar\mu_{t_n},\bar\mu_{t_*})\longrightarrow0.
\]
Assumption~\ref{ass:stability-v9} gives
$\|S_{D_n}(t_n)\|\le M$. Since $\operatorname{supp}\bar\mu$ has dense
linear span, Lemma~\ref{lem:compactness-v9} applies and provides a
further subsequence, with the same notation, such that
\begin{equation}\label{eq:moving-SOT-v9}
 S_{D_n}(t_n)\bm{z}\longrightarrow B\bm{z}\quad\text{for every }\bm{z}\in\mathcal Z,
 \qquad B\in\mathcal L(\mathcal Z),\quad\|B\|\le M.
\end{equation}

\emph{Step 3: Identification of subsequential limits under FT.}
Fix $s\in[0,\tau]$. The pair
$(S_{D_n}(t_n)\bm{Z}_s,B\bm{Z}_s)$ gives
\[
 W_2((S_{D_n}(t_n))_\#\mu_s,B_\#\mu_s)^2
 \le\mathbb E\|(S_{D_n}(t_n)-B)\bm{Z}_s\|^2\longrightarrow0.
\]
Indeed, \Eqref{eq:moving-SOT-v9} gives pointwise convergence, and the
integrand is bounded by $4M^2\|\bm{Z}_s\|^2$, which is integrable.
Combining this limit with \Eqref{eq:eD-v9} and $W_2$ continuity of
$r\mapsto\mu_r$ gives
\begin{align*}
 W_2(B_\#\mu_s,\mu_{s+t_*})
 \le W_2(B_\#\mu_s,(S_{D_n}(t_n))_\#\mu_s)+e_{D_n}+W_2(\mu_{s+t_n},\mu_{s+t_*})\longrightarrow0.
\end{align*}
The subsequence in \Eqref{eq:moving-SOT-v9} is independent of $s$
and converges at every $\bm{z}\in\mathcal Z$.
Thus the same argument applies to every $s\in[0,\tau]$ without further
extraction, establishing $B_\#\mu_s=\mu_{s+t_*}$ for all such $s$.
Assumption~\ref{ass:FT-v9} therefore implies $B=T(t_*)$, including
$B=I$ when $t_*=0$.

\emph{Step 4: Uniform strong convergence on $[0,\tau]$.}
If convergence failed to be uniform for some fixed $\bm{z}\in\mathcal Z$,
there would exist $\varepsilon>0$, $D_n\to\infty$, and
$t_n\in[0,\tau]$ with
$\|S_{D_n}(t_n)\bm{z}-T(t_n)\bm{z}\|\ge\varepsilon$.
Steps 2--3 give a further subsequence for which $t_n\to t_*$ and
$S_{D_n}(t_n)\bm{z}\to T(t_*)\bm{z}$.
Strong continuity gives $T(t_n)\bm{z}\to T(t_*)\bm{z}$, a contradiction.
Consequently,
\begin{equation}\label{eq:local-SOT-v9}
 \sup_{0\le t\le\tau}\|S_D(t)\bm{z}-T(t)\bm{z}\|\longrightarrow0
 \qquad\text{for every }\bm{z}\in\mathcal Z.
\end{equation}

\emph{Step 5: Extension to arbitrary finite time intervals.}
For each fixed integer $k\ge0$,
$S_D(\tau)^k\bm{z}\to T(\tau)^k\bm{z}$.
This follows inductively from \Eqref{eq:local-SOT-v9} and
$\|S_D(\tau)\|\le M$, using
\begin{align*}
 S_D(\tau)^{k+1}\bm{z}-T(\tau)^{k+1}\bm{z}
 ={}S_D(\tau)\bigl(S_D(\tau)^k\bm{z}-T(\tau)^k\bm{z}\bigr)+\bigl(S_D(\tau)-T(\tau)\bigr)T(\tau)^k\bm{z}.
\end{align*}
For $t=k\tau+r$ with $0\le r<\tau$, the semigroup property gives
\[
 \|S_D(t)\bm{z}-T(t)\bm{z}\|
 \le M\|S_D(\tau)^k\bm{z}-T(\tau)^k\bm{z}\|
    +\|(S_D(r)-T(r))T(\tau)^k\bm{z}\|.
\]
For $t\in[0,H]$, only finitely many $k$ occur. The first term tends to
zero uniformly in $r$, and \Eqref{eq:local-SOT-v9}, applied to the
finitely many fixed vectors $T(\tau)^k\bm{z}$, makes the second term vanish
uniformly in $r$. Hence
$\sup_{0\le t\le H}\|S_D(t)\bm{z}-T(t)\bm{z}\|\to0$.
Finally, $S_D(t)-K_D(t)=I-P_D$ gives
\[
 \sup_{0\le t\le H}\|K_D(t)\bm{z}-T(t)\bm{z}\|
 \le\sup_{0\le t\le H}\|S_D(t)\bm{z}-T(t)\bm{z}\|+\|(I-P_D)\bm{z}\|
 \longrightarrow0,
\]
which proves \Eqref{eq:SOT-v9}.
\end{proof}

The theorem establishes strong operator convergence locally uniformly in
time, including on finite intervals beyond $[0,2\tau]$.
The extension relies on the global semigroup property after stability
and FT have established convergence on $[0,\tau]$.
\citet{korda2018convergence} analyze EDMD, which fits the Koopman action
on a finite observable space by least squares from paired state
observations. For a fixed observable space, the infinite-sample limit
projects each evolved observable back onto that space in the $L^2$
inner product induced by the sampling measure. Under their assumptions,
strong operator convergence then follows as the observable space is
enlarged.
Here, convergence follows from the weak population objective,
compactness of pushforward measures, and FT.
\paragraph{Limitations.}
Our convergence analysis as $D\to\infty$ assumes an underlying autonomous
vector field whose induced distributions form an absolutely continuous
Wasserstein flow. It concerns minimizers of the exact weak population
residual over a compatible family of observables
obtained from nested projections of a fixed Koopman-invariant RKHS.
Under the regularity conditions in
Sections~\ref{sec:rkhs-preliminaries}--\ref{sec:weak-residual-convergence},
the stability condition in Assumption~\ref{ass:stability-v9}, and
finite-time linear determinacy (FT) in Assumption~\ref{ass:FT-v9},
the feature propagators $K_D(t)$ and $S_D(t)$ converge strongly to
$T(t)=U(t)^*$, locally uniformly in time.
In practice, only finite samples at sparse observation times are available,
so we discretize the time integrals using trapezoidal quadrature.
We further use ridge regularization and exponential moving average (EMA)
updates to stabilize the estimated generator during alternating optimization
of neural representations and dynamics (Appendix~\ref{app:algorithmic-details}).
The analysis does not control the resulting sampling, quadrature,
regularization, or optimization errors, nor establish convergence of this training procedure
or of independently learned neural representations as $D$ increases.
Strong convergence of the adjoints to $U(t)$, operator-norm convergence,
and spectral convergence also remain outside these guarantees.

\section{Algorithmic Details}
\label{app:algorithmic-details}

This appendix develops the training objectives and algorithms for
KoopCell and its memory-augmented extension, KoopCell-M, introduced in
Sections~\ref{sec:linear-comparison} and~\ref{subsec:memory}.
We first describe how KoopCell alternates between VAE representation
learning and closed-form estimation of the latent dynamics.
We then introduce the memory embedding and developmental-path
conditioning in KoopCell-M, detailing the resulting inference
and training procedures.
Throughout, $\rho_{t_k}=n_k^{-1}\sum_{i=1}^{n_k}\delta_{\bm{x}_{k,i}}$ denotes
an observed snapshot, with $\bm{x}_{k,i}\in\mathbb R^{d_x}$ and
$\mathcal T=\{0=t_0<\cdots<t_N=T\}$ the training times.
We retain $\bm{\theta},\bm{\phi}$ for the encoder and decoder parameters and $D$
for the latent dimension. Held-out snapshots are used only for evaluation.

\subsection{KoopCell: Koopman dynamics and alternating training}
\label{app:linear-method}

KoopCell alternates between learning a VAE representation for a given
latent dynamics generator and estimating the linear dynamics generator in the current latent coordinates.
We describe the VAE objective first, then derive the empirical generator
update from the finite-test formulation proposed in Theorem~\ref{thm:main-finite-test}.

\subsubsection{VAE representation learning}
\label{app:linear-elbo}

For a cell with expression $\bm{x}$ observed at time $t$, we use a posterior encoder
$q_{\bm{\theta}}(\cdot\mid \bm{x},t)=\mathcal N(\bm{m}_{\bm{\theta}}(\bm{x},t),\bm{S}_{\bm{\theta}}(\bm{x},t))$ with diagonal
covariance $\bm{S}_{\bm{\theta}}$. A shared MLP takes $(\bm{x},t)$ as input and predicts the mean
and log-variance. The observation model is
$p_{\bm{\phi}}(\bm{x}\mid \bm{z})=\mathcal N(\bm{x};\operatorname{Dec}_{\bm{\phi}}(\bm{z}),\sigma_x^2\bm{I}_{d_x})$,
with fixed variance $\sigma_x^2$.
Reconstruction and population matching use reparameterized samples
$\bm{z}=\bm{m}_{\bm{\theta}}+\bm{S}_{\bm{\theta}}^{1/2}\bm{\epsilon}$, $\bm{\epsilon}\sim\mathcal N(\bm{0},\bm{I}_D)$,
and predicted expression is given by $\operatorname{Dec}_{\bm{\phi}}(\bm{z})$.
The resulting encoded population is
$\mu_t=\int q_{\bm{\theta}}(\cdot\mid \bm{x},t)\mathrm{d}\rho_t(\bm{x})$.

Given the linear latent flow $\mathcal F_t(\bm{z})=\bm{F}_t\bm{z}+\bm{c}_t$
(Appendix~\ref{app:linear-weak-regression}), we regularize the
representation with the transported prior
$p_t=(\mathcal F_t)_\#\mathcal N(\bm{0},\bm{I}_D)=\mathcal N(\bm{c}_t,\bm{\Gamma}_t)$,
where $\bm{\Gamma}_t=\bm{F}_t\bm{F}_t^\top$.
Here $\mu_t$ denotes the encoder-induced population and $p_t$ the generative prior.
We write $q_{\bm{\theta}}(\cdot\mid \bm{x},t)$ for the encoder distribution and $q_{\bm{\theta}}(\bm{z}\mid \bm{x},t)$ for its density at $\bm{z}$.
Within fixed-time expectations, we suppress time subscripts on random states.
The resulting generative model
$p_{\bm{\phi},t}(\bm{x},\bm{z})=p_{\bm{\phi}}(\bm{x}\mid \bm{z})p_t(\bm{z})$ yields
$p_{\bm{\phi},t}(\bm{x})=\mathbb E_{\bm{Z}\sim q_{\bm{\theta}}(\cdot\mid \bm{x},t)}
[p_{\bm{\phi}}(\bm{x}\mid \bm{Z})p_t(\bm{Z})/q_{\bm{\theta}}(\bm{Z}\mid \bm{x},t)]$.
Jensen's inequality therefore gives the evidence lower bound (ELBO)
\begin{equation}
 \log p_{\bm{\phi},t}(\bm{x})\ge
 \mathcal E_{\mathrm{KC}}(\bm{x},t)
 :=\mathbb E_{q_{\bm{\theta}}}\log p_{\bm{\phi}}(\bm{x}\mid \bm{Z})
   -\operatorname{KL}\bigl(q_{\bm{\theta}}(\cdot\mid \bm{x},t)\,\|\,p_t\bigr).
 \label{eq:linear-elbo}
\end{equation}
The Gaussian observation model gives a squared reconstruction error
up to an additive constant. Following \Eqref{eq:linear-comparison-vae},
we normalize it by the expression dimension:
$\mathcal L_{\mathrm{rec}}=-d_x^{-1}\mathbb E_{t,\bm{x}}
\mathbb E_{q_{\bm{\theta}}}\log p_{\bm{\phi}}(\bm{x}\mid \bm{Z})$,
where $\mathbb E_{t,\bm{x}}$ averages equally over training times and uniformly
over cells within each time.
Writing $\bm{m}=\bm{m}_{\bm{\theta}}(\bm{x},t)$ and $\bm{S}=\bm{S}_{\bm{\theta}}(\bm{x},t)$, we compute
the Gaussian KL divergence analytically:
\begin{equation}
 \operatorname{KL}\bigl(\mathcal N(\bm{m},\bm{S})\,\|\,p_t\bigr)
 =\tfrac12\left[\operatorname{tr}(\bm{\Gamma}_t^{-1}\bm{S})
 +(\bm{m}-\bm{c}_t)^\top\bm{\Gamma}_t^{-1}(\bm{m}-\bm{c}_t)-D
 +\log\frac{\det\bm{\Gamma}_t}{\det \bm{S}}\right].
 \label{eq:linear-gaussian-kl}
\end{equation}
The training objective
$\mathcal L_{\mathrm{VAE}}=\mathcal L_{\mathrm{rec}}
+\beta\mathbb E_{t,\bm{x}}\operatorname{KL}(q_{\bm{\theta}}(\cdot\mid \bm{x},t)\,\|\,p_t)$
thus equals the negative ELBO divided by $d_x$ when $\beta=1/d_x$;
otherwise, $\beta$ controls the strength of the variational regularization.

To learn representations whose evolution under the latent flow matches
the observed populations, we augment $\mathcal L_{\mathrm{VAE}}$ with
prediction losses in latent and expression space. For every forward pair
$(s,t)\in\mathcal P=\{(s,t)\in\mathcal T^2:s<t\}$, we form
$\widehat\mu_{s\to t}=(\mathcal F_{t-s})_\#\mu_s$ and
$\widehat\rho_{s\to t}=(\operatorname{Dec}_{\bm{\phi}})_\#\widehat\mu_{s\to t}$,
and average the prediction losses over these pairs:
\begin{equation}
 \mathcal L_{\mathrm{pop}}
 =\frac1{|\mathcal P|}\sum_{(s,t)\in\mathcal P}\bigl[
 \lambda_z\mathcal S_\varepsilon(\mu_t,\widehat\mu_{s\to t})
 +\lambda_x\mathcal S_\varepsilon(\rho_t,\widehat\rho_{s\to t})\bigr].
 \label{eq:algorithm-population-loss}
\end{equation}
Here $\mathcal S_\varepsilon$ is the debiased Sinkhorn divergence with
quadratic cost.
With the dynamics fixed, we update $\bm{\theta},\bm{\phi}$ by minimizing
$\mathcal L_{\mathrm{KC}}=\mathcal L_{\mathrm{VAE}}
+\mathcal L_{\mathrm{pop}}+\lambda_{\mathrm{weak}}\mathcal L_{\mathrm{weak}}$,
where the weak-residual loss is defined below.
Reconstruction and prediction losses update both the encoder and decoder;
the weak-residual term updates only the encoder.

To initialize the representation before alternating training, we first
pretrain the VAE with a standard-normal prior at every training time,
minimizing
\begin{equation}
 \mathcal L_{\mathrm{pre}}=\mathcal L_{\mathrm{rec}}
 +\beta_{\mathrm{pre}}\mathbb E_{t,\bm{x}}\operatorname{KL}
 \bigl(q_{\bm{\theta}}(\cdot\mid \bm{x},t)\,\|\,\mathcal N(\bm{0},\bm{I}_D)\bigr),
 \label{eq:linear-pretraining-loss}
\end{equation}
with a gradually increased $\beta_{\mathrm{pre}}$.
The pretrained encoder then provides the initial latent coordinates for
the linear dynamics estimation described next.

\subsubsection{Empirical weak residual and closed-form dynamics estimation}
\label{app:linear-weak-regression}
\label{app:finite-test-regression-v6}

With the representation fixed, we now derive a closed-form estimator
for the linear generator that best fits the evolution of the latent distributions.
For notational simplicity, Section~\ref{sec:linear-comparison} presents
the method using $\dot{\bm{Z}}_t=\bm{A}\bm{Z}_t$. In the implementation, we include
the constant observable $1$, giving $\dot{\bm{Z}}_t=\bm{A}\bm{Z}_t+\bm{b}_z$ and coefficients
$\bm{\Theta}=[\bm{A},\bm{b}_z]\in\mathbb R^{D\times(D+1)}$.
Writing $\widetilde{\bm{z}}=(\bm{z}^\top,1)^\top$ and
$\widetilde{\bm{A}}=\left(\begin{smallmatrix}\bm{A}&\bm{b}_z\\\bm{0}^\top&0\end{smallmatrix}\right)$,
we evaluate the flow through
\begin{equation}
 \exp(\Delta\widetilde{\bm{A}})
 =\begin{pmatrix}\bm{F}_\Delta&\bm{c}_\Delta\\\bm{0}^\top&1\end{pmatrix},\qquad
 \mathcal F_\Delta(\bm{z})=\bm{F}_\Delta \bm{z}+\bm{c}_\Delta.
 \label{eq:linear-affine-transition}
\end{equation}
The constant coordinate stays equal to one, and only the upper $D$ rows
are estimated.

Theorem~\ref{thm:main-finite-test} shows that restricting the dual
supremum to a finite test space gives a least-squares regression
objective for estimating the generator;
Lemma~\ref{lem:finite-dual-v6} in Appendix~\ref{app:finite-tests-v6}
provides the corresponding derivation.
For the linear drift $\bm{\Theta}\widetilde{\bm{z}}$, the moment definitions in
\Eqref{eq:linear-comparison-finite-statistics} require only replacing
$\bm{z}^\top$ by $\widetilde{\bm{z}}^\top$ in $\bm{H}_j$; $\bm{y}$ and $\bm{M}$ are unchanged.
Thus, with $\bm{a}=\operatorname{vec}(\bm{\Theta})$, the residual vector remains
$\bm{y}-\bm{G}\bm{a}$ and the objective is $\frac12(\bm{y}-\bm{G}\bm{a})^\top \bm{M}^{-1}(\bm{y}-\bm{G}\bm{a})$ when
$\bm{M}\succ0$. We next construct its empirical counterpart from the
observed snapshots by forming these moments on adjacent time intervals.

Specifically, on each interval $I_k=[t_k,t_{k+1}]$, we use a smooth spatial test
$\psi_j$ with bounded values and gradients.
The residual identity \Eqref{eq:residual-pair-app-v6} in
Appendix~\ref{app:wc-geometry-v6}, applied to
$\varphi_{kj}(t,\bm{z})=\mathbf1_{I_k}(t)\psi_j(\bm{z})$, gives\footnote{The
interval test is understood by smooth approximation in $\mathbb V_D$,
using the continuous residual extension in
\Eqref{eq:residual-velocity-pair-v6}.}
\begin{equation}\label{eq:interval-exact-v6}
 \mathcal R_D(\bm{\Theta};\varphi_{kj})
 =\mu_{t_{k+1}}[\psi_j]-\mu_{t_k}[\psi_j]
 -\left\langle\bm{\Theta},\int_{I_k}
      \mu_t[\nabla\psi_j(\bm{z})\widetilde{\bm{z}}^\top]\mathrm{d}t\right\rangle_F.
\end{equation}
Here the endpoint difference supplies $y_{kj}$, and the time integral
supplies $\bm{H}_{kj}$ in the finite-test construction.
The Gram matrix defined in \Eqref{eq:M-finite-v6} is block diagonal,
$\bm{M}=\operatorname{diag}(\bm{M}_0,\ldots,\bm{M}_{N-1})$; each block
is $(\bm{M}_k)_{ij}=\int_{I_k}\mu_t[\nabla\psi_i^\top\nabla\psi_j]\mathrm{d}t$.

In the implementation, we use random Fourier tests
$\cos(\bm{\xi}_j^\top \bm{z})$ and $\sin(\bm{\xi}_j^\top \bm{z})$ to construct this finite
test space. Frequencies are sampled once from
$\mathcal N(\bm{0},s^2\bm{I}_D/D)$ in equal-sized groups for
$s\in\{0.5,1,2,4\}$, then held fixed during training.
The standard bank contains 1,024 frequencies and hence 2,048 tests.
Their gradients, $-\bm{\xi}_j\sin(\bm{\xi}_j^\top \bm{z})$ and
$\bm{\xi}_j\cos(\bm{\xi}_j^\top \bm{z})$, allow all required moments to be evaluated
directly from encoded cells.

For the empirical regression, we choose the encoder means as the
deterministic observables to which the theorem is applied.
At each training time $t_k$, we encode the observed cells as
$\bm{z}_{k,i}=\bm{m}_{\bm{\theta}}(\bm{x}_{k,i},t_k)$, yielding the empirical latent distribution
$\mu_{t_k}^{\mathrm{mean}}=(\bm{m}_{\bm{\theta}}(\cdot,t_k))_\#\rho_{t_k}$.
With $\widetilde{\bm{z}}_{k,i}=(\bm{z}_{k,i}^\top,1)^\top$, the endpoint statistics are
\begin{equation}
 u_{kj}=\frac1{n_k}\sum_i\psi_j(\bm{z}_{k,i}),\qquad
 \bm{D}_{kj}=\frac1{n_k}\sum_i
       \nabla\psi_j(\bm{z}_{k,i})\widetilde{\bm{z}}_{k,i}^\top.
 \label{eq:linear-endpoint-statistics}
\end{equation}
In \Eqref{eq:interval-exact-v6}, $u_{kj}$ replaces the endpoint expectation
$\mu_{t_k}[\psi_j]$, and $\bm{D}_{kj}$ replaces the integrand
$\mu_{t_k}[\nabla\psi_j(\bm{z})\widetilde{\bm{z}}^\top]$ at $t_k$.
Approximating the time integral by the trapezoidal rule, with
$\Delta t_k=t_{k+1}-t_k$, therefore gives
\begin{equation}
 y_{kj}=u_{k+1,j}-u_{kj},\qquad
 \bm{H}_{kj}=\frac{\Delta t_k}{2}(\bm{D}_{kj}+\bm{D}_{k+1,j}),\qquad
 \bm{G}_{(k,j),:}=\operatorname{vec}(\bm{H}_{kj})^\top.
 \label{eq:linear-empirical-design}
\end{equation}
Following the regression interpretation in the
\hyperref[rem:weak-residual-regression]{remark} after
Lemma~\ref{lem:finite-dual-v6}, the empirical residual over interval $I_k=[t_k,t_{k+1}]$ is
$\widehat{\mathcal R}_{kj}(\bm{\Theta})
=y_{kj}-\langle\bm{\Theta},\bm{H}_{kj}\rangle_F
=y_{kj}-(\bm{G}\bm{a})_{(k,j)}$.
This is the empirical, time-discretized approximation of
$\mathcal R_D(\bm{\Theta};\varphi_{kj})$ in \Eqref{eq:interval-exact-v6}.
The construction accommodates irregular observation times; further, if the
moment integrand has two bounded time derivatives, the interval
quadrature error is $\mathcal O(\Delta t_k^3)$.
Applying the same quadrature to the Gram matrix gives
\begin{equation}
 (\bm{M}_k)_{ij}=\frac{\Delta t_k}{2}
 \left\{\mu_{t_k}^{\mathrm{mean}}[
          \nabla\psi_i^\top\nabla\psi_j]
       +\mu_{t_{k+1}}^{\mathrm{mean}}[
          \nabla\psi_i^\top\nabla\psi_j]\right\}.
 \label{eq:linear-empirical-gram}
\end{equation}
Finally, we stack these empirical residuals over all interval--test pairs $(k, j)$
to obtain $\bm{y}-\bm{G}\bm{a}$, the residual vector in the finite-test objective of
\Eqref{eq:finite-dual-v6}.

With these empirical quantities and $\bm{W}=(\bm{M}+\epsilon_M\bm{I})^{-1}$, we
 define
$\mathcal L_{\mathrm{weak}}=(\bm{G}\bm{a}-\bm{y})^\top \bm{W}(\bm{G}\bm{a}-\bm{y})$.
To stabilize the estimation when $\bm{G}$ is poorly conditioned,
we additionally introduce a ridge penalty:
\begin{equation}
 \widehat{\bm{a}}
 =\operatorname*{arg\,min}_{\bm{a}}
   \left\{\tfrac12(\bm{G}\bm{a}-\bm{y})^\top \bm{W}(\bm{G}\bm{a}-\bm{y})
             +\tfrac\lambda2\|\bm{a}\|_2^2\right\}
 =(\bm{G}^\top \bm{W}\bm{G}+\lambda \bm{I})^{-1}\bm{G}^\top \bm{W}\bm{y},
 \qquad \lambda>0.
 \label{eq:linear-ridge-solve}
\end{equation}
We form the regression statistics from all training cells and solve the
normal equations using preconditioned conjugate gradients. The fitted 
coefficients $\widehat{\bm{\Theta}}=\operatorname{reshape}(\widehat{\bm{a}})$
are then used to update the dynamics before refining the representation.
However, the pretrained latent distributions need not follow a common
linear flow, so directly using $\widehat{\bm{\Theta}}$ in the representation
objective can induce abrupt updates and distort the pretrained latent
manifold. We therefore introduce the closed-form dynamics generator gradually through an
exponential moving average,
$\bm{\Theta}\leftarrow\alpha\bm{\Theta}+(1-\alpha)\widehat{\bm{\Theta}}$, with $\alpha=0.99$ in the real
single-cell dynamics learning procedure.

With the updated $\bm{\Theta}$ held fixed, we optimize the VAE with the objective in
Appendix~\ref{app:linear-elbo}, recomputing the weak residual on minibatches.
Algorithm~\ref{alg:linear-training} summarizes the alternating procedure.
In both algorithms, $\operatorname{sg}$ stops gradients,
$\|\bm{v}\|_{\bm{W}}^2=\bm{v}^\top \bm{W}\bm{v}$, and
$\operatorname{AdamW}(\bm{w},\nabla_{\bm{w}}\mathcal L)$ denotes one optimizer
update of parameters $\bm{w}$ using the indicated gradient.
In Algorithm~\ref{alg:linear-training},
$\operatorname{WeakStats}(\{\bm{z}_{k,i}\};\mathcal T,\{\psi_j\})$ returns
$(\bm{G},\bm{y},\bm{M})$ using \Eqref{eq:linear-endpoint-statistics},
\Eqref{eq:linear-empirical-design}, and \Eqref{eq:linear-empirical-gram};
the same operation is used for full snapshots and minibatches.
$\operatorname{Transition}(\bm{\Theta},\Delta)$ returns $(\bm{F}_\Delta,\bm{c}_\Delta)$
via \Eqref{eq:linear-affine-transition}, and $\operatorname{PCG}(\bm{Q},\bm{r})$
solves $\bm{Q}\bm{a}=\bm{r}$ by preconditioned conjugate gradients.
The loss routines evaluate the referenced objectives on the indicated
minibatches $\mathcal B=\{\mathcal B_k\}_{k=0}^N$.
$\operatorname{Sample}$ draws $n_{\mathrm b}$ cells independently from
each snapshot.
The pairwise predictions depend only on the source samples and fixed
dynamics, so they can be computed in parallel.

\begin{algorithm}[!t]
 \caption{KoopCell: alternating closed-form generator estimation and representation learning}
 \label{alg:linear-training}\label{alg:training}
 \small
 \algrenewcommand\algorithmicrequire{\textbf{Input:}}
 \algrenewcommand\algorithmicensure{\textbf{Output:}}
 \algrenewcommand\algorithmiccomment[1]{\hfill{\footnotesize\color{algcomment}$\triangleright$\,#1}}
 \algrenewcommand\alglinenumber[1]{{\scriptsize\color{algcomment}#1:}}
 \setlength{\fboxsep}{2.5pt}
 \begin{algorithmic}[1]
  \Require Training snapshots $\{(t_k,\rho_{t_k})\}_{k=0}^N$; fixed Fourier tests $\{\psi_j\}_{j=1}^m$;\newline
    batch size $n_{\mathrm b}$; $\lambda,\epsilon_M>0$; $\alpha\in[0,1)$; loss weights
  \Ensure Encoder $q_{\bm{\theta}}$, decoder $\operatorname{Dec}_{\bm{\phi}}$, and linear dynamics parameters $\bm{\Theta}=[\bm{A},\bm{b}_z]$
  \Statex \colorbox{algpretrain!8}{\parbox{\dimexpr\linewidth-2\fboxsep\relax}{%
    \strut\textcolor{algpretrain}{\textbf{Stage 1: VAE pretraining}}}}
  \For{each pretraining iteration}
   \State $\mathcal B_k\gets\operatorname{Sample}(\rho_{t_k},n_{\mathrm b}),\quad k=0,\ldots,N$
     \Comment{Independent sampling at each time}
   \State $\mathcal L_{\mathrm{pre}}\gets\operatorname{PretrainLoss}(\mathcal B;\bm{\theta},\bm{\phi},\beta_{\mathrm{pre}})$
     \Comment{Standard-normal VAE; \Eqref{eq:linear-pretraining-loss}}
   \State $(\bm{\theta},\bm{\phi})\gets\operatorname{AdamW}((\bm{\theta},\bm{\phi}),\nabla_{\bm{\theta},\bm{\phi}}\mathcal L_{\mathrm{pre}})$
     \Comment{Pretrain encoder and decoder}
  \EndFor
  \Statex \colorbox{algalternating!8}{\parbox{\dimexpr\linewidth-2\fboxsep\relax}{%
    \strut\textcolor{algalternating}{\textbf{Stage 2: Alternating dynamics estimation and representation learning}}}}
  \State $\bm{\Theta}\gets\bm{0}$
  \For{each training iteration}
   \Statex \hspace{\algorithmicindent}\textcolor{algalternating}{\textbf{(a) Closed-form dynamics estimation}}
   \State $\bar{\bm{z}}_{k,i}\gets\operatorname{sg}(\bm{m}_{\bm{\theta}}(\bm{x}_{k,i},t_k)),\quad k=0,\ldots,N,\ i=1,\ldots,n_k$
   \State $(\bm{G},\bm{y},\bm{M})\gets\operatorname{WeakStats}(\{\bar{\bm{z}}_{k,i}\};\mathcal T,\{\psi_j\})$;
     $\bm{W}\gets(\bm{M}+\epsilon_M \bm{I})^{-1}$
   \State $\widehat{\bm{a}}\gets\operatorname{PCG}(\bm{G}^\top \bm{W}\bm{G}+\lambda \bm{I},\bm{G}^\top \bm{W}\bm{y})$;
     $\widehat{\bm{\Theta}}\gets\operatorname{reshape}(\widehat{\bm{a}})$
     \Comment{closed-form estimation; \Eqref{eq:linear-ridge-solve}}
   \State $\bm{\Theta}\gets\operatorname{sg}(\alpha\bm{\Theta}+(1-\alpha)\widehat{\bm{\Theta}})$
     \Comment{EMA update}
   \Statex \hspace{\algorithmicindent}\textcolor{algalternating}{\textbf{(b) Representation update}}
   \State $\mathcal B_k\gets\operatorname{Sample}(\rho_{t_k},n_{\mathrm b}),\quad k=0,\ldots,N$
     \Comment{Fresh minibatches for VAE updates}
   \State $\bar{\bm{z}}^{\mathcal B}_{k,i}\gets \bm{m}_{\bm{\theta}}(\bm{x}_{k,i},t_k),\quad \bm{x}_{k,i}\in\mathcal B_k$;
     $\bm{\epsilon}_{k,i}\sim\mathcal N(\bm{0},\bm{I}_D)$
   \State $\bm{z}_{k,i}\gets\bar{\bm{z}}^{\mathcal B}_{k,i}+\bm{S}_{\bm{\theta}}(\bm{x}_{k,i},t_k)^{1/2}\bm{\epsilon}_{k,i}$
     \Comment{Reparameterized posterior samples}
   \State $(\bm{F}_\Delta,\bm{c}_\Delta)\gets\operatorname{Transition}(\bm{\Theta},\Delta),\quad
     \Delta\in\mathcal T\cup\{t-s:(s,t)\in\mathcal P\}$
     \Comment{Calculate the transition matrix}
   \State $p_{t_k}\gets\mathcal N(\bm{c}_{t_k},\bm{F}_{t_k}\bm{F}_{t_k}^\top)$
     \Comment{Transported Gaussian prior}
   \State $\mathcal L_{\mathrm{VAE}}\gets\operatorname{VAELoss}(\mathcal B;\bm{\theta},\bm{\phi},\{p_{t_k}\})$
     \Comment{Reconstruction + KL; \Eqref{eq:linear-comparison-vae}}
   \For{each $(s,t)\in\mathcal P$ \textbf{in parallel}}
    \State $\widehat{\bm{z}}_{s\to t,i}\gets \bm{F}_{t-s}\bm{z}_{s,i}+\bm{c}_{t-s}$;
      $\widehat{\bm{x}}_{s\to t,i}\gets\operatorname{Dec}_{\bm{\phi}}(\widehat{\bm{z}}_{s\to t,i})$
   \EndFor
   \State $\mathcal L_{\mathrm{pop}}\gets\operatorname{PopulationLoss}
     (\mathcal B,\{\bm{z}_{k,i}\},\{\widehat{\bm{z}}_{s\to t,i},\widehat{\bm{x}}_{s\to t,i}\})$
     \Comment{Population matching; \Eqref{eq:algorithm-population-loss}}
   \State $(\bm{G}_{\mathcal B},\bm{y}_{\mathcal B},\bm{M}_{\mathcal B})\gets
     \operatorname{WeakStats}(\{\bar{\bm{z}}^{\mathcal B}_{k,i}\};\mathcal T,\{\psi_j\})$
   \State $\bm{W}_{\mathcal B}\gets(\bm{M}_{\mathcal B}+\epsilon_M \bm{I})^{-1}$;
     $\mathcal L_{\mathrm{weak}}\gets
     \|\bm{G}_{\mathcal B}\operatorname{vec}(\bm{\Theta})-\bm{y}_{\mathcal B}\|_{\bm{W}_{\mathcal B}}^2$
     \Comment{Weak population residual loss}
   \State $\mathcal L_{\mathrm{KC}}\gets\mathcal L_{\mathrm{VAE}}+\mathcal L_{\mathrm{pop}}
     +\lambda_{\mathrm{weak}}\mathcal L_{\mathrm{weak}}$
     \Comment{Combined training objective}
   \State $(\bm{\theta},\bm{\phi})\gets\operatorname{AdamW}((\bm{\theta},\bm{\phi}),\nabla_{\bm{\theta},\bm{\phi}}\mathcal L_{\mathrm{KC}})$
     \Comment{Update encoder and decoder}
  \EndFor
  \State \Return $q_{\bm{\theta}},\operatorname{Dec}_{\bm{\phi}},\bm{A},\bm{b}_z$
 \end{algorithmic}
\end{algorithm}

For evaluation, we sample $\bm{Z}_0$ from the encoder distribution of the initial
observed population snapshot and return $\operatorname{Dec}_{\bm{\phi}}(\mathcal F_t(\bm{Z}_0))$
at each requested time. 

\subsection{KoopCell-M: memory-augmented branching dynamics and alternating training}
\label{app:memory-method}

This section develops the memory extension introduced in Section~\ref{subsec:memory}.
We first introduce the augmented latent dynamics for modeling branching, then
derive the probabilistic model and training objectives.
The resulting algorithm retains the two alternating steps of KoopCell:
closed-form dynamics estimation followed by representation learning.

\subsubsection{Memory-augmented dynamics and developmental paths}
\label{app:memory-dynamics}
\label{app:memory-branching}

\citet[Proposition~3.3]{zhao2026delay} establishes that an ODE with a Lipschitz vector field
cannot exactly transport between distributions supported on different
numbers of disjoint compact, path-connected components.
In particular, a linear flow cannot split a connected initial population
into disconnected fate populations; continuous decoding also preserves
connectedness. When branching is understood as the emergence of separated
density modes, a linear latent flow has a further limitation: it preserves
Gaussianity, so a single Gaussian population remains unimodal.
Although nonlinear decoding can introduce additional density modes, a decoder
with Lipschitz constant $L_{\mathrm{dec}}$ limits the amplification of latent
separation:
\begin{equation}
 \|\operatorname{Dec}_{\bm{\phi}}(\mathcal F_t(\bm{z}))
   -\operatorname{Dec}_{\bm{\phi}}(\mathcal F_t(\bm{z}'))\|
 \le L_{\mathrm{dec}}\|e^{\bm{A}t}\|\,\|\bm{z}-\bm{z}'\|.
 \label{eq:memory-decoder-separation}
\end{equation}
To model branching in the latent dynamics, we draw on the
Mori--Zwanzig formalism: unresolved observables influence
the resolved dynamics through memory (Section~\ref{sec:preliminaries}).
Specifically, we augment the \emph{visible latent state}
$\bm{Z}_t\in\mathbb R^D$, which represents gene expression, with
\emph{memory variables} $\bm{H}_t\in\mathbb R^{d_h}$ that influence the
dynamics of $\bm{Z}_t$. The decoder reconstructs gene expression from $\bm{Z}_t$
alone. The memory variables can represent fate-relevant
influences not captured by the measured expression state
\citep{weinreb2020lineage}.

Further, a linear augmented process with a single joint Gaussian initialization
would still have Gaussian marginals. To generate distinct developmental
branches, we therefore introduce a discrete variable $R\in\mathscr R$
indexing heterogeneous hidden initial conditions:
\begin{equation}
 R\sim\operatorname{Cat}(\bm{\pi}),\qquad
 \bm{Z}_0\mid R=r\sim\mathcal N(\bm{0},\bm{I}_D),\qquad
 \bm{H}_0\mid \bm{Z}_0=\bm{z},R=r\sim\mathcal N(\bm{F}_r\bm{z}+\bm{a}_r,\bm{S}_r).
 \label{eq:memory-initial}
\end{equation}
Here $\pi_r$ are normalized mixture weights and $\bm{S}_r\succ0$.
Conditional on $R=r$, we evolve the augmented state $(\bm{Z}_t, \bm{H}_t)$ by
\begin{equation}
 \mathrm{d}\bm{Z}_t=(\bm{A}\bm{Z}_t+\bm{B}\bm{H}_t+\bm{b}_z)\,\mathrm{d}t,\qquad
 \mathrm{d}\bm{H}_t=(\bm{C}\bm{Z}_t+\bm{\Lambda} \bm{H}_t+\bm{b}_h(r))\,\mathrm{d}t+\bm{\Sigma}_h\,\mathrm{d}\bm{W}_t,
 \label{eq:memory-augmented-dynamics}
\end{equation}
where $\bm{W}_t$ is a standard Brownian motion. Motivated by the use of
decaying memory in related models~\citep{lei2016parameterization,iLED},
we parameterize
$\bm{\Lambda}=(\bm{L}-\bm{L}^\top)/2-\operatorname{diag}(\delta+\operatorname{softplus}(\bm{d}))$,
with $\delta>0$, making the homogeneous $\bm{H}$ block dissipative.
Through the coupling $\bm{B}\bm{H}_t$, cells with similar visible states but
different hidden initial conditions can follow distinct latent evolution.
Each component remains Gaussian, but their mixture can
develop multiple modes when their separation exceeds their spread.
With nondegenerate component covariances, the mixture retains connected
support; branching here refers to the separation of latent density modes.

To connect this augmented model with the Mori--Zwanzig formulation in
Section~\ref{sec:preliminaries}, we eliminate $\bm{H}_t$ from
\Eqref{eq:memory-augmented-dynamics}, obtaining a non-Markovian
generalized Langevin equation for $\bm{Z}_t$:
\begin{equation}
 \begin{aligned}
 \dot{\bm{Z}}_t&=\bm{A}\bm{Z}_t+\bm{b}_z+\int_0^t \bm{B}e^{\bm{\Lambda}(t-s)}\bm{C}\bm{Z}_s\,\mathrm{d}s+\bm{\eta}_r(t),\\
 \bm{\eta}_r(t)&=\bm{B}e^{\bm{\Lambda} t}\bm{H}_0
 +\int_0^t \bm{B}e^{\bm{\Lambda}(t-s)}\bm{b}_h(r)\,\mathrm{d}s
 +\bm{B}\int_0^t e^{\bm{\Lambda}(t-s)}\bm{\Sigma}_h\,\mathrm{d}\bm{W}_s.
 \end{aligned}
 \label{eq:memory-elimination}
\end{equation}
The kernel $\bm{K}(u)=\bm{B}e^{\bm{\Lambda} u}\bm{C}$ describes dependence on the past
history, while $\bm{\eta}_r(t)$ collects the effects of the hidden initial
state, bias, and noise. This decomposition into memory and forcing is
motivated by Mori--Zwanzig reduction~\citep{lin2021datamori}.

\begin{figure}[tb]
 \centering
 \includegraphics[width=0.94\linewidth]{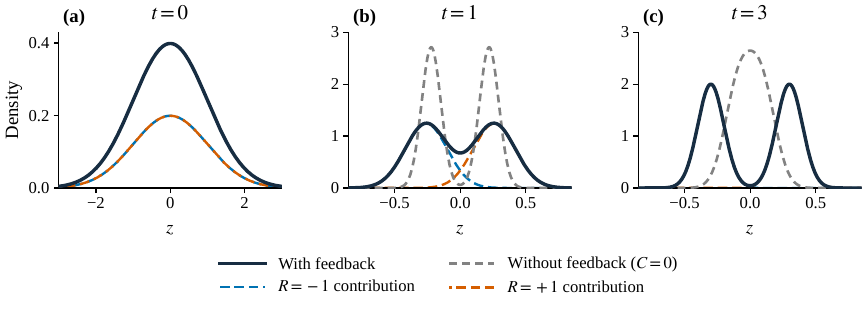}
 \caption{\textbf{Memory feedback sustains visible branching.}
 Analytic densities in Example~\ref{ex:memory-branching}, with $a=0.6$ and
 $\sigma_H=0.2$. Both initial marginals are unimodal.
 The solid curve is the visible mixture with $K(u)=e^{-u}$; colored dashed
 curves illustrate its weighted components. The gray dashed curve removes only
 the feedback ($C=0$), keeping the same initialization. Both models develop
 two modes at $t=1$, but the comparison returns to a single mode by $t=3$.
 Axis scales vary across panels.}
 \label{fig:memory-branching-example}
\end{figure}

\begin{example}[Emergence of branching in the memory-augmented model]
\label{ex:memory-branching}
Let $R\in\{-1,1\}$ have equal probabilities,
$Z_0\mid R=r\sim\mathcal N(0,1)$, and
$H_0\mid Z_0=z,R=r\sim\mathcal N(-z+ra,\sigma_H^2)$, with
$0<\sigma_H<a<\sqrt{1+\sigma_H^2}$.
Both initial marginals are unimodal: $Z_0$ is standard Gaussian, and
$H_0$ is an equal mixture of $\mathcal N(\pm a,1+\sigma_H^2)$.
Consider the augmented dynamics
$\mathrm{d}Z_t=(-Z_t+H_t)\,\mathrm{d}t$ and $\mathrm{d}H_t=(Z_t-H_t)\,\mathrm{d}t$ in
\Eqref{eq:memory-augmented-dynamics}.
Here $A=\Lambda=-1$ and $B=C=1$, so eliminating $H_t$ gives the
memory kernel $K(u)=e^{-u}$.

Writing $\kappa_t=(1-e^{-2t})/2$, the solution is
$Z_t=e^{-2t}Z_0+\kappa_t(Z_0+H_0)$.
Conditional on $R=r$, the initialization makes $Z_0+H_0$ independent of
$Z_0$, with law $\mathcal N(ra,\sigma_H^2)$. Thus the visible density is
\begin{equation}
 p_t(z)=\tfrac12\mathcal N(z;-a\kappa_t,v_t)
       +\tfrac12\mathcal N(z;a\kappa_t,v_t),\qquad
 v_t=e^{-4t}+\sigma_H^2\kappa_t^2.
 \label{eq:memory-branching-example}
\end{equation}
The components coincide at $t=0$. When $a^2\kappa_t^2>v_t$,
$p_t''(0)=(a^2\kappa_t^2-v_t)v_t^{-2}\mathcal N(0;a\kappa_t,v_t)>0$;
symmetry and decay at infinity then imply a maximum on each side of zero.
This condition holds for
$t>\tfrac12\log(1+2/\sqrt{a^2-\sigma_H^2})$.
As $t\to\infty$, the component means approach $\pm a/2$ and their
variance approaches $\sigma_H^2/4$, so the two modes persist.

To isolate the role of memory feedback, set $C=0$ while keeping the other
coefficients and initialization fixed. The component means then become
$\pm at e^{-t}$, with common variance
$e^{-2t}[(1-t)^2+\sigma_H^2t^2]$.
Since $a^2<1+\sigma_H^2$, this comparison becomes unimodal again for
sufficiently large $t$, as illustrated in
Figure~\ref{fig:memory-branching-example}.
The joint initialization is non-Gaussian: $R$ encodes latent fate
information, and the memory feedback sustains the resulting visible modes.
\end{example}

{In this construction, $R$ parameterizes hidden heterogeneity.
For cellular development, we use $R$ to represent \textit{candidate developmental
pathways}, whose progression is supervised using biological state annotations.}

\paragraph{{Biological state annotation.}}
{We use coarse biological states to model developmental
paths, encompassing broad cell identities and developmental-stage
categories. Initial training labels are obtained by matching cell IDs to
\href{https://shendure-web.gs.washington.edu/content/members/DEAP_website/public/}{published annotations for DR}~\citep{doi:10.1126/science.abn5800}
and \href{https://drive.google.com/file/d/1E494DhIx5RLy0qv_6eWa9426Bfmq28po/view}{WOT cell sets for SC}~\citep{schiebinger2019optimal}.
For ZB, we annotate Louvain
clusters at the latest training time using marker expression, following
the terminal-cluster annotation approach of
\citet{doi:10.1126/science.aar3131}. We also define early developmental
states using sampling time. For example, the first two training snapshots
are labeled Blastula progenitor in ZB and Blastoderm in DR; the initial SC
population is labeled MEF. For ZB and SC, we fit a logistic regression
classifier to the initially labeled training cells and predict states for
the remaining training cells. Logistic regression is also used for
supervised cell annotation in CellTypist~\citep{dominguezconde2022crosstissue}.
We calibrate the predicted probabilities and use only assignments that
pass confidence, support, and consistency filters as additional supervision.
For DR, state supervision uses only labels obtained from the published
annotations or assigned using sampling time. Classifier fitting and label
supervision use only training cells.}

\paragraph{{Candidate developmental paths.}}
{We use these annotated biological states as nodes to define a
candidate directed acyclic graph (DAG). We specify its core edges using
published developmental relationships~\citep{doi:10.1126/science.aar3131,doi:10.1126/science.abn5800,schiebinger2019optimal}
and retain them as biological priors.
Additional edges are selected using optimal transport between adjacent
training snapshots, subject to biological compatibility and support
criteria. An edge represents an admissible transition between coarse
states. Enumerating root-to-terminal paths defines $\mathscr R$, with
$r=(c_1,\ldots,c_{\ell_r})$ specifying an ordered developmental pathway.
Each path parameterizes a hidden initial law in \Eqref{eq:memory-initial}.
The path identity $R$ remains fixed along a trajectory, while the biological
state progresses along the pathway. We further introduce a path
consistency loss to encourage generated trajectories to follow the
developmental order specified by $R$
(\Eqref{eq:memory-path-loss} in Appendix~\ref{app:memory-objectives}).
Figure~\ref{fig:sc-candidate-dag} shows the graph for SC, whose four
root-to-terminal paths define the candidate values of $R$.}

\begin{figure}[tb]
 \centering
 \includegraphics[width=0.91\linewidth]{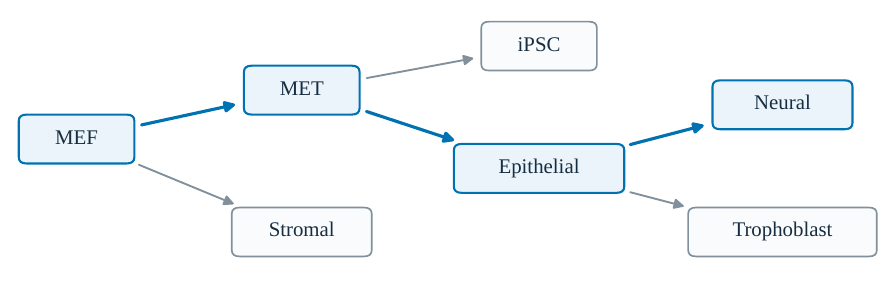}
 \caption{\textbf{Candidate developmental paths for SC.}
 {Nodes represent seven coarse biological states, and edges
 represent the developmental transitions described by
 \citet{schiebinger2019optimal}. The graph defines four root-to-terminal
 paths.} The highlighted sequence
 MEF $\to$ MET $\to$ Epithelial $\to$ Neural illustrates one value of $R$.
 {Each path parameterizes a hidden initial distribution and
 specifies the developmental order used in the path consistency loss
 (\Eqref{eq:memory-path-loss}).}}
 \label{fig:sc-candidate-dag}
\end{figure}

\subsubsection{Variational inference and representation learning}
\label{app:memory-objectives}
\label{app:memory-gaussian-inference}
\label{app:memory-transition}

The augmented latent dynamics define a time-dependent prior for learning the
VAE representation. At a known sampling time $t$, the observed variable
is a single-cell expression vector $\bm{X}_t$; the latent variables are the developmental
path $R$, the visible state $\bm{Z}_t$, and the memory state $\bm{H}_t$.

\textbf{Generative model.}
We draw $R\sim\operatorname{Cat}(\bm{\pi})$ and evolve
$\bm{Y}_t=(\bm{Z}_t^\top,\bm{H}_t^\top)^\top$ from the path-dependent initialization
in \Eqref{eq:memory-initial} under
\Eqref{eq:memory-augmented-dynamics}. This gives the Gaussian prior
$p_{\bm{\vartheta},t}(\bm{z},\bm{h}\mid r)$. Gene expression is generated from $\bm{Z}_t$
alone through the same observation model as KoopCell:
\begin{equation}
 \begin{aligned}
 p_{\bm{\phi},\bm{\vartheta},t}(\bm{x},\bm{z},\bm{h},r)
 &=\pi_r\,p_{\bm{\vartheta},t}(\bm{z},\bm{h}\mid r)\,p_{\bm{\phi}}(\bm{x}\mid \bm{z}),\\
 p_{\bm{\phi}}(\bm{x}\mid \bm{z})
 &=\mathcal N\bigl(\bm{x};\operatorname{Dec}_{\bm{\phi}}(\bm{z}),\sigma_x^2\bm{I}_{d_x}\bigr).
 \end{aligned}
 \label{eq:memory-generative-model}
\end{equation}
Here $\sigma_x^2$ is fixed, as in Appendix~\ref{app:linear-elbo}, and
$\bm{\vartheta}$ collects memory and initial-distribution parameters;
dependence of the prior on $\bm{\Theta}=[\bm{A},\bm{b}_z]$ is left implicit.

\textbf{Inference model.}
Given expression $\bm{x}$ at time $t$, we approximate the joint posterior
$p_{\bm{\phi},\bm{\vartheta},t}(\bm{z},\bm{h},r\mid \bm{x})$ by
\begin{equation}
 q_t(\bm{z},\bm{h},r\mid \bm{x})
 =q_{\bm{\theta}}(\bm{z}\mid \bm{x},t)\,q_t(r\mid \bm{z},\bm{x})\,q_t^H(\bm{h}\mid \bm{z},r,\bm{x}).
 \label{eq:memory-variational-family}
\end{equation}
The Gaussian encoder
$q_{\bm{\theta}}(\cdot\mid \bm{x},t)=\mathcal N(\bm{m}_{\bm{\theta}}(\bm{x},t),\bm{S}_{\bm{\theta}}(\bm{x},t))$
infers the visible latent state; $q_t(r\mid \bm{z},\bm{x})$ assigns path probabilities,
and $q_t^H(\bm{h}\mid \bm{z},r,\bm{x})$ infers the memory state conditional on $\bm{z}$ and $r$.
Using the factorization in \Eqref{eq:memory-generative-model}, Jensen's
inequality gives
\begin{equation}
 \begin{aligned}
 \log p_{\bm{\phi},\bm{\vartheta},t}(\bm{x})
 &\ge \mathbb E_{q_t}\left[
   \log\frac{p_{\bm{\phi},\bm{\vartheta},t}(\bm{x},\bm{Z},\bm{H},R)}{q_t(\bm{Z},\bm{H},R\mid \bm{x})}\right]\\
 &=\mathbb E_{q_{\bm{\theta}}}\log p_{\bm{\phi}}(\bm{x}\mid \bm{Z})
   -\operatorname{KL}\bigl(q_t(\cdot\mid \bm{x})\,\|\,
                            p_{\bm{\vartheta},t}\bigr),
 \end{aligned}
 \label{eq:memory-full-elbo}
\end{equation}
where $p_{\bm{\vartheta},t}(\bm{z},\bm{h},r)=\pi_r p_{\bm{\vartheta},t}(\bm{z},\bm{h}\mid r)$.
We next specify the memory and path posteriors in
\Eqref{eq:memory-variational-family} to evaluate this bound.

Conditional on $R=r$, the Gaussian initialization and linear augmented
dynamics give a jointly Gaussian prior for $(\bm{Z}_t,\bm{H}_t)$.
The conditional distribution $p_{\bm{\vartheta}}(\bm{h}\mid \bm{z},r,t)$ is therefore
Gaussian and can be computed in closed form.
Because the observation model depends only on $\bm{Z}_t$,
\Eqref{eq:memory-generative-model} implies
$p(\bm{H}_t\mid \bm{Z}_t,R,\bm{X}_t)=p(\bm{H}_t\mid \bm{Z}_t,R)$.
We therefore use this Gaussian conditional as the memory posterior,
setting $q_t^H(\bm{h}\mid \bm{z},r,\bm{x})=p_{\bm{\vartheta}}(\bm{h}\mid \bm{z},r,t)$.
To derive its mean and covariance, write
$\bm{\mathsf{K}}=\left(\begin{smallmatrix}\bm{A}&\bm{B}\\\bm{C}&\bm{\Lambda}\end{smallmatrix}\right)$,
$\bm{b}_r=(\bm{b}_z^\top,\bm{b}_h(r)^\top)^\top$, and
$\bm{\Sigma}=(\bm{0},\bm{\Sigma}_h^\top)^\top$.
The transition is $\bm{Y}_{t+\Delta}\mid \bm{Y}_t,R=r\sim
\mathcal N(\bm{\Phi}_\Delta \bm{Y}_t+\bm{J}_\Delta \bm{b}_r,\bm{Q}_\Delta)$, with
\begin{equation}
 \bm{\Phi}_\Delta=e^{\bm{\mathsf{K}}\Delta},\qquad
 \bm{J}_\Delta=\int_0^\Delta e^{\bm{\mathsf{K}}s}\,\mathrm{d}s,\qquad
 \bm{Q}_\Delta=\int_0^\Delta e^{\bm{\mathsf{K}}s}\bm{\Sigma}\bm{\Sigma}^\top e^{\bm{\mathsf{K}}^\top s}\,\mathrm{d}s.
 \label{eq:memory-transition}
\end{equation}
The initialization in \Eqref{eq:memory-initial} gives
\begin{equation}
 \bm{m}_{0,r}=\begin{pmatrix}\bm{0}\\\bm{a}_r\end{pmatrix},\qquad
 \bm{V}_{0,r}=\begin{pmatrix}\bm{I}_D&\bm{F}_r^\top\\\bm{F}_r&\bm{F}_r\bm{F}_r^\top+\bm{S}_r\end{pmatrix}.
 \label{eq:memory-joint-initial}
\end{equation}
Hence $\bm{Y}_t\mid R=r\sim\mathcal N(\bm{m}_{t,r},\bm{V}_{t,r})$, where
$\bm{m}_{t,r}=\bm{\Phi}_t \bm{m}_{0,r}+\bm{J}_t \bm{b}_r$ and
$\bm{V}_{t,r}=\bm{\Phi}_t \bm{V}_{0,r}\bm{\Phi}_t^\top+\bm{Q}_t$.
Gaussian conditioning gives
$q_t^H=\mathcal N(\bar{\bm{h}}_{t,r}(\bm{z}),\bm{V}^{h\mid z}_{t,r})$, with
\begin{equation}
 \bar{\bm{h}}_{t,r}(\bm{z})=\bm{m}^h_{t,r}
   +\bm{V}^{hz}_{t,r}(\bm{V}^{zz}_{t,r})^{-1}(\bm{z}-\bm{m}^z_{t,r}),\qquad
 \bm{V}^{h\mid z}_{t,r}=\bm{V}^{hh}_{t,r}
   -\bm{V}^{hz}_{t,r}(\bm{V}^{zz}_{t,r})^{-1}\bm{V}^{zh}_{t,r},
 \label{eq:memory-conditioning}
\end{equation}
Having specified the memory conditional $q_t^H$, we complete the
inference model by defining the path probabilities $q_t(r\mid \bm{z},\bm{x})$.
The same joint Gaussian prior gives the visible marginal
$p_{\bm{\vartheta},t}(\bm{z}\mid r)=\mathcal N(\bm{z};\bm{m}^z_{t,r},\bm{V}^{zz}_{t,r})$.
We combine this density with the path prior $\pi_r$ and an
{annotation weight for biological states}, $\chi_r(\bm{x})$, then
normalize over candidate paths:
\begin{equation}
 q_t(r\mid \bm{z},\bm{x})
 =\frac{\pi_r p_{\bm{\vartheta},t}(\bm{z}\mid r)\chi_r(\bm{x})}
 {\sum_{r'\in\mathscr R}\pi_{r'}p_{\bm{\vartheta},t}(\bm{z}\mid r')\chi_{r'}(\bm{x})}.
 \label{eq:memory-path-probabilities}
\end{equation}
{Here $\chi_r(\bm{x})$ sums the annotation probabilities over
biological states on path $r$: an accepted state label restricts support
to paths containing that state, soft annotations give soft weights, and
an unlabeled cell uses $\chi_r(\bm{x})=1$.}
At least one compatible path is required for an annotated cell.

\textbf{ELBO and training objective.}
With both posteriors specified, we return to the bound in
\Eqref{eq:memory-full-elbo}.
Write $q_{\bm{\theta},t}(\bm{z},r\mid \bm{x})=q_{\bm{\theta}}(\bm{z}\mid \bm{x},t)q_t(r\mid \bm{z},\bm{x})$ and
$p_{\bm{\vartheta},t}(\bm{z},r)=\pi_r p_{\bm{\vartheta},t}(\bm{z}\mid r)$.
Since $q_t^H(\bm{h}\mid \bm{z},r,\bm{x})=p_{\bm{\vartheta}}(\bm{h}\mid \bm{z},r,t)$, the joint-density
ratio in the bound becomes
\begin{equation}
 \begin{aligned}
 \frac{p_{\bm{\phi},\bm{\vartheta},t}(\bm{x},\bm{z},\bm{h},r)}{q_t(\bm{z},\bm{h},r\mid \bm{x})}
 &=\frac{p_{\bm{\phi}}(\bm{x}\mid \bm{z})\,p_{\bm{\vartheta},t}(\bm{z},r)\,
          p_{\bm{\vartheta}}(\bm{h}\mid \bm{z},r,t)}
        {q_{\bm{\theta},t}(\bm{z},r\mid \bm{x})\,p_{\bm{\vartheta}}(\bm{h}\mid \bm{z},r,t)}\\
 &=\frac{p_{\bm{\phi}}(\bm{x}\mid \bm{z})\,p_{\bm{\vartheta},t}(\bm{z},r)}
        {q_{\bm{\theta},t}(\bm{z},r\mid \bm{x})}.
 \end{aligned}
 \label{eq:memory-elbo-cancellation}
\end{equation}
The common memory conditional cancels, so the ratio no longer depends
on $\bm{h}$. Taking its logarithm and expectation therefore integrates out
$\bm{H}$ and gives
\begin{equation}
 \begin{aligned}
 \log p_{\bm{\phi},\bm{\vartheta},t}(\bm{x})\ge\mathcal E_{\mathrm M}(\bm{x},t)
 &:=\mathbb E_{q_{\bm{\theta}}}\log p_{\bm{\phi}}(\bm{x}\mid \bm{Z})
   +\mathbb E_{q_{\bm{\theta},t}}\log
     \frac{p_{\bm{\vartheta},t}(\bm{Z},R)}{q_{\bm{\theta},t}(\bm{Z},R\mid \bm{x})}\\
 &=\mathbb E_{q_{\bm{\theta}}}\log p_{\bm{\phi}}(\bm{x}\mid \bm{Z})
   -\operatorname{KL}\bigl(q_{\bm{\theta},t}(\cdot\mid \bm{x})\,\|\,
                            p_{\bm{\vartheta},t}\bigr).
 \end{aligned}
 \label{eq:memory-elbo}
\end{equation}
Expanding the remaining KL over the discrete paths yields
\begin{equation}
 \operatorname{KL}(q_{\bm{\theta},t}\,\|\,p_{\bm{\vartheta},t})
 =\mathbb E_{\bm{Z}\sim q_{\bm{\theta}}}\sum_{r\in\mathscr R}q_t(r\mid \bm{Z},\bm{x})
 \log\frac{q_{\bm{\theta}}(\bm{Z}\mid \bm{x},t)q_t(r\mid \bm{Z},\bm{x})}
 {\pi_r\mathcal N(\bm{Z};\bm{m}^z_{t,r},\bm{V}^{zz}_{t,r})}.
 \label{eq:memory-joint-kl}
\end{equation}
We enumerate the paths and estimate the expectation with reparameterized
encoder samples. Thus the conditional hidden posterior is analytic, while the
mixture KL is evaluated by sampling over $\bm{Z}$.

{To encourage generated trajectories to follow the candidate
developmental pathway represented by $R$, we introduce a path consistency
loss based on the probabilities over biological states predicted by a
latent state classifier $c_{\bm{\omega}}(c\mid \bm{z})$.}
We train the classifier using the annotations above by minimizing the
cross-entropy loss
${\mathcal L_{\mathrm{state}}}=-\mathbb E_{t,\bm{x},c}
\mathbb E_{\bm{Z}\sim q_{\bm{\theta}}(\cdot\mid \bm{x},t)}\log c_{\bm{\omega}}(c\mid \bm{Z})$,
where the outer expectation averages over training cells with supervised
{biological state labels} $c$, and the inner expectation is over their encoder samples.
We combine ${\mathcal L_{\mathrm{state}}}$ with the normalized reconstruction
loss and KL regularization from Appendix~\ref{app:linear-elbo} to form
the snapshot objective
\begin{equation}
 \mathcal L_{\mathrm{snap}}
 =\mathcal L_{\mathrm{rec}}
 +\beta\mathbb E_{t,\bm{x}}\operatorname{KL}
 \bigl(q_{\bm{\theta},t}(\cdot\mid \bm{x})\,\|\,p_{\bm{\vartheta},t}\bigr)
 +\lambda_c{\mathcal L_{\mathrm{state}}}.
 \label{eq:memory-snapshot-loss}
\end{equation}

As in KoopCell, we also match predicted and observed populations over all
forward training pairs $s<t$. In KoopCell-M this requires inferring
memory at the source time. The inferred augmented source population is
\begin{equation}
 \widetilde\mu_s(\mathrm{d}\bm{z},\mathrm{d}\bm{h},r)
 =\int q_{\bm{\theta}}(\mathrm{d}\bm{z}\mid \bm{x},s)\,q_s(r\mid \bm{z},\bm{x})\,
 p_{\bm{\vartheta}}(\mathrm{d}\bm{h}\mid \bm{Z}_s=\bm{z},R=r)\,\mathrm{d}\rho_s(\bm{x}).
 \label{eq:memory-source-population}
\end{equation}
We propagate $\widetilde\mu_s$ to time $t$ using
\Eqref{eq:memory-transition}, keeping $r$ fixed along each trajectory.
We denote the predicted $\bm{Z}_t$-marginal by $\widehat\mu_{s\to t}$ and
map this latent population to gene-expression space through the decoder:
$\widehat\rho_{s\to t}=(\operatorname{Dec}_{\bm{\phi}})_\#\widehat\mu_{s\to t}$.
We compare these predictions with the corresponding populations at time
$t$ using $\mathcal L_{\mathrm{pop}}$ defined in
\Eqref{eq:algorithm-population-loss}.

\paragraph{Path consistency loss.}
{For a candidate developmental pathway
$r=(c_1,\ldots,c_{\ell_r})$, we sum over admissible timings of transitions
between successive biological states.}
On a grid $0=u_1<\cdots<u_M=T$, let $\mathcal I_r$ contain index sequences
with $i_1=1$, $i_m-i_{m-1}\in\{0,1\}$, and $1\le i_m\le\ell_r$.
The classifier assigns the path-consistency probability
\begin{equation}
 \mathcal P_r(\bm{Z}_{u_1:u_M})
 =\sum_{(i_1,\ldots,i_M)\in\mathcal I_r}
   \prod_{m=1}^M c_{\bm{\omega}}(c_{i_m}\mid \bm{Z}_{u_m}),
 \label{eq:memory-path-loss}
\end{equation}
and we minimize
$\mathcal L_{\mathrm{path}}=-\sum_r\bar q_T(r)
\mathbb E\log\mathcal P_r(\bm{Z}_{u_1:u_M})$.
Here trajectories start from uniformly sampled initial cells conditional
on each path, and $\bar q_T(r)$ is the detached mean path probability at the
latest training time. We evaluate \Eqref{eq:memory-path-loss} by forward
dynamic programming, adapting the alignment marginalization in
connectionist temporal classification (CTC; \citealp{graves2006connectionist})
to the constraints in $\mathcal I_r$.
The recursion uses no blank symbol and sums over all final path positions,
allowing trajectories to end before the {terminal state}.
The path loss updates both the classifier and the parameters governing
the rollout.
In Algorithm~\ref{alg:memory-training}, $\bm{Z}^{(r)}_{u_1:u_M}$
denotes a batch of simulated visible trajectories conditional on $R=r$.

\begin{algorithm}[!t]
 \caption{KoopCell-M: memory augmented latent dynamics for modeling branching}
 \label{alg:memory-training}
 \small
 \algrenewcommand\algorithmicrequire{\textbf{Input:}}
 \algrenewcommand\algorithmicensure{\textbf{Output:}}
 \algrenewcommand\algorithmiccomment[1]{\hfill{\footnotesize\color{algcomment}$\triangleright$\,#1}}
 \algrenewcommand\alglinenumber[1]{{\scriptsize\color{algcomment}#1:}}
 \setlength{\fboxsep}{2.5pt}
 \begin{algorithmic}[1]
  \Require Training snapshots $\mathcal D=\{(t_k,\rho_{t_k})\}_{k=0}^N$
    {with biological state annotations};\newline
    candidate paths $\mathscr R$;
    fixed Fourier tests $\{\psi_j\}_{j=1}^m$; batch size $n_{\mathrm b}$;
    $\lambda,\epsilon_M>0$; $\alpha\in[0,1)$; loss weights
  \Ensure Encoder $\bm{q}_{\bm{\theta}}$, decoder $\operatorname{Dec}_{\bm{\phi}}$, {state classifier} $c_{\bm{\omega}}$,
    and dynamics parameters $\bm{\Theta}=[\bm{A},\bm{b}_z],\bm{\vartheta}$
  \Statex \colorbox{algpretrain!8}{\parbox{\dimexpr\linewidth-2\fboxsep\relax}{%
    \strut\textcolor{algpretrain}{\textbf{Stage 1: VAE and state classifier pretraining}}}}
  \For{each pretraining iteration}
   \State $\mathcal B_k\gets\operatorname{Sample}(\rho_{t_k},n_{\mathrm b}),\quad k=0,\ldots,N$
     \Comment{Independent sampling at each time}
   \State $\mathcal L\gets\operatorname{PretrainLoss}(\mathcal B;\bm{\theta},\bm{\phi})
     +\lambda_c{\operatorname{StateLoss}}(\mathcal B;\bm{\theta},\bm{\omega})$
     \Comment{VAE loss: \Eqref{eq:linear-pretraining-loss}}
   \State $(\bm{\theta},\bm{\phi},\bm{\omega})\gets\operatorname{AdamW}((\bm{\theta},\bm{\phi},\bm{\omega}),\nabla_{\bm{\theta},\bm{\phi},\bm{\omega}}\mathcal L)$
     \Comment{{Pretrain the VAE and state classifier}}
  \EndFor
  \Statex \colorbox{algalternating!8}{\parbox{\dimexpr\linewidth-2\fboxsep\relax}{%
    \strut\textcolor{algalternating}{\textbf{Stage 2: Alternating dynamics estimation and representation learning}}}}
  \State $\bm{\Theta}\gets\bm{0}$; initialize $\bm{B},\bm{C},\bm{\Lambda},\bm{\Sigma}_h$
    and $\{\pi_r,\bm{F}_r,\bm{a}_r,\bm{S}_r,\bm{b}_h(r)\}_{r\in\mathscr R}$
  \State $\bm{\eta}\gets(\bm{\theta},\bm{\phi},\bm{\vartheta},\bm{\omega})$
    \Comment{Load pretrained encoder, decoder, and classifier from Stage 1}
  \For{each training iteration}
   \Statex \hspace{\algorithmicindent}\textcolor{algalternating}{\textbf{(a) Closed-form dynamics estimation ($\bm{\eta}$ fixed)}}
   \State $\bm{z}^{\mathrm{all}}\gets\operatorname{sg}(\operatorname{Encode}(\mathcal D;\bm{\theta}))$
     \Comment{Reparameterized samples from all training cells}
   \State $(\bm{q}^{\mathrm{all}},\mathcal C^{\mathrm{all}})\gets
     \operatorname{Infer}(\mathcal D,\bm{z}^{\mathrm{all}};\operatorname{sg}(\bm{\Theta},\bm{\vartheta}))$
   \Statex \Comment{$\bm{q}$: path probabilities (\Eqref{eq:memory-path-probabilities});
     $\mathcal C$: Gaussian moments and conditional coefficients (\Eqref{eq:memory-conditioning})}
   \State $(\bm{G},\bm{y}^{\mathrm M},\bm{M})\gets
     \operatorname{MemoryStats}(\bm{z}^{\mathrm{all}},\bm{q}^{\mathrm{all}},\mathcal C^{\mathrm{all}},\operatorname{sg}(\bm{B}))$
   \Statex \Comment{Compute $\bm{G},\bm{y}^{\mathrm M},\bm{M}$ by sample averaging and trapezoidal quadrature;
     \Eqref{eq:memory-weak-design}, \Eqref{eq:linear-empirical-gram}}
   \State $\bm{W}\gets(\bm{M}+\epsilon_M \bm{I})^{-1}$;
     $\widehat{\bm{a}}\gets\operatorname{PCG}(\bm{G}^\top \bm{W}\bm{G}+\lambda \bm{I},\bm{G}^\top \bm{W}\bm{y}^{\mathrm M})$
   \State $\widehat{\bm{\Theta}}\gets\operatorname{reshape}(\widehat{\bm{a}})$;
     $\bm{\Theta}\gets\operatorname{sg}(\alpha\bm{\Theta}+(1-\alpha)\widehat{\bm{\Theta}})$
     \Comment{EMA update}
   \Statex \hspace{\algorithmicindent}\textcolor{algalternating}{\textbf{(b) Representation and memory update ($\bm{\Theta}$ fixed)}}
   \State $\mathcal B_k\gets\operatorname{Sample}(\rho_{t_k},n_{\mathrm b}),\quad k=0,\ldots,N$;
     $\bm{\epsilon}_{k,i}\sim\mathcal N(\bm{0},\bm{I}_D)$
   \State $\bm{z}_{k,i}\gets \bm{m}_{\bm{\theta}}(\bm{x}_{k,i},t_k)+\bm{S}_{\bm{\theta}}(\bm{x}_{k,i},t_k)^{1/2}\bm{\epsilon}_{k,i}$
     \Comment{Reparameterized posterior samples}
   \State $(\bm{\Phi}_\Delta,\bm{J}_\Delta,\bm{Q}_\Delta)\gets\operatorname{OUTransition}(\bm{\Theta},\bm{\vartheta},\Delta),\quad
     \Delta\in\mathcal T\cup\{t-s:(s,t)\in\mathcal P\}$
   \State $(\bm{q},\mathcal C)\gets\operatorname{Infer}(\mathcal B,\bm{z};\bm{\Theta},\bm{\vartheta})$
     \Comment{Path probabilities and Gaussian conditionals}
   \State $\mathcal L_{\mathrm{snap}}\gets\operatorname{SnapshotLoss}(\mathcal B,\bm{z},\bm{q},\mathcal C;\bm{\eta})$
     \Comment{{Reconstruction, KL, state supervision}; \Eqref{eq:memory-snapshot-loss}}
   \For{each $(s,t)\in\mathcal P$ \textbf{in parallel}}
    \State $\bm{h}_{s,i,r}\sim\mathcal N(\bar{\bm{h}}_{s,r}(\bm{z}_{s,i}),\bm{V}^{h\mid z}_{s,r})$ for all $i,r$
      \Comment{Infer memory at the source time $s$}
    \State $\widehat{\bm{y}}_{s\to t,i,r}\gets
      \bm{\Phi}_{t-s}(\bm{z}_{s,i}^\top,\bm{h}_{s,i,r}^\top)^\top+\bm{J}_{t-s}\bm{b}_r+\bm{\epsilon}_{i,r}$,
      $\bm{\epsilon}_{i,r}\sim\mathcal N(\bm{0},\bm{Q}_{t-s})$
    \State $\widehat\mu_{s\to t}\gets
      |\mathcal B_s|^{-1}\sum_{i,r}q_{s,i,r}
      \delta_{[\widehat{\bm{y}}_{s\to t,i,r}]_{\bm{z}}}$;
      $\widehat\rho_{s\to t}\gets(\operatorname{Dec}_{\bm{\phi}})_\#\widehat\mu_{s\to t}$
   \EndFor
   \State $\mathcal L_{\mathrm{pop}}\gets\operatorname{PopulationLoss}(\mathcal B,\bm{z},
     \{\widehat\mu_{s\to t},\widehat\rho_{s\to t}\})$
     \Comment{Population matching; \Eqref{eq:algorithm-population-loss}}
   \State $\bm{Z}^{(r)}_{u_1:u_M}\gets\operatorname{PathRollout}(\mathcal B_0,r;\bm{\theta},\bm{\Theta},\bm{\vartheta}),\quad r\in\mathscr R$
   \State $\bar{\bm{q}}_T\gets\operatorname{sg}(|\mathcal B_N|^{-1}\sum_i \bm{q}_{N,i,\cdot})$;
     $\mathcal L_{\mathrm{path}}\gets\operatorname{PathLoss}(\{\bm{Z}^{(r)}_{u_1:u_M}\},\bar{\bm{q}}_T;\bm{\omega})$
   \State $(\bm{G}_{\mathcal B},\bm{y}^{\mathrm M}_{\mathcal B},\bm{M}_{\mathcal B})\gets
     \operatorname{MemoryStats}(\bm{z},\operatorname{sg}(\bm{q}),\operatorname{sg}(\mathcal C),\operatorname{sg}(\bm{B}))$
     \Comment{\Eqref{eq:memory-weak-design}}
   \State $\bm{W}_{\mathcal B}\gets(\bm{M}_{\mathcal B}+\epsilon_M \bm{I})^{-1}$
   \State $\mathcal L_{\mathrm{weak}}\gets
      \|\bm{y}^{\mathrm M}_{\mathcal B}-\bm{G}_{\mathcal B}\operatorname{vec}(\bm{\Theta})\|_{\bm{W}_{\mathcal B}}^2$
     \Comment{Retain encoder gradients through $\bm{z}$}
   \State $\mathcal L_{\mathrm{M}}\gets\mathcal L_{\mathrm{snap}}+\mathcal L_{\mathrm{pop}}
      +\lambda_R\mathcal L_{\mathrm{path}}+\lambda_{\mathrm{weak}}\mathcal L_{\mathrm{weak}}$
     \Comment{Combined objective; \Eqref{eq:memory-training-objective}}
   \State $\bm{\eta}\gets\operatorname{AdamW}(\bm{\eta},\nabla_{\bm{\eta}}\mathcal L_{\mathrm{M}})$
     \Comment{Update VAE, memory dynamics, and classifier}
  \EndFor
  \State \Return $\bm{q}_{\bm{\theta}},\operatorname{Dec}_{\bm{\phi}},c_{\bm{\omega}},\bm{\Theta},\bm{\vartheta}$
 \end{algorithmic}
\end{algorithm}

\subsubsection{Closed-form dynamics estimation and alternating training}
\label{app:memory-weak-regression}

We now derive the closed-form update of $\bm{\Theta}=[\bm{A},\bm{b}_z]$ with the
representation and memory statistics fixed. Let $\widetilde\mu_t$ be the
inferred augmented population in \Eqref{eq:memory-source-population},
whose $\bm{Z}$-marginal is $\mu_t$, and write
$\widetilde{\bm{z}}=(\bm{z}^\top,1)^\top$ and $\widetilde{\bm{Z}}_t=(\bm{Z}_t^\top,1)^\top$.
For a smooth test $\varphi(t,\bm{z})$, substituting the visible drift
$\bm{\Theta}\widetilde{\bm{z}}+\bm{B}\bm{h}$ into the definition of the weak population
residual in \Eqref{eq:residual-pair-app-v6}
(Appendix~\ref{app:wc-geometry-v6}) gives
\begin{equation}
 \begin{aligned}
 \mathcal R_D^{\mathrm M}(\bm{\Theta};\varphi)
 &=\mu_T[\varphi(T)]-\mu_0[\varphi(0)]\\
 &\quad-\int_0^T\sum_{r\in\mathscr R}\int
   \bigl[\partial_t\varphi(t,\bm{z})
   +\nabla_{\bm{z}}\varphi(t,\bm{z})^\top(\bm{\Theta}\widetilde{\bm{z}}+\bm{B}\bm{h})\bigr]
   \,\widetilde\mu_t(\mathrm{d}\bm{z},\mathrm{d}\bm{h},r)\,\mathrm{d}t\\
 &=\mathcal R_D(\bm{\Theta};\varphi)
   -\int_0^T\sum_{r\in\mathscr R}\int
   \nabla_{\bm{z}}\varphi(t,\bm{z})^\top \bm{B}\bm{h}\,\widetilde\mu_t(\mathrm{d}\bm{z},\mathrm{d}\bm{h},r)\,\mathrm{d}t,
 \end{aligned}
 \label{eq:memory-residual-functional}
\end{equation}
where $\mathcal R_D(\bm{\Theta};\varphi)$ is the population residual used in
Appendix~\ref{app:linear-weak-regression}.
Using \Eqref{eq:memory-source-population} and the conditional mean
$\int \bm{h}\,p_{\bm{\vartheta}}(\mathrm{d}\bm{h}\mid \bm{z},r,t)=\bar{\bm{h}}_{t,r}(\bm{z})$ from
\Eqref{eq:memory-conditioning} gives
\begin{equation}
 \begin{aligned}
 &\sum_{r\in\mathscr R}\int
   \nabla_{\bm{z}}\varphi(t,\bm{z})^\top \bm{B}\bm{h}\,\widetilde\mu_t(\mathrm{d}\bm{z},\mathrm{d}\bm{h},r)\\
 &\quad=\int\!\!\int \nabla_{\bm{z}}\varphi(t,\bm{z})^\top \bm{B}
   \left[\sum_{r\in\mathscr R}q_t(r\mid \bm{z},\bm{x})\bar{\bm{h}}_{t,r}(\bm{z})\right]
   q_{\bm{\theta}}(\mathrm{d}\bm{z}\mid \bm{x},t)\,\rho_t(\mathrm{d}\bm{x})\\
 &\quad=\mathbb E\bigl[
   \nabla_{\bm{z}}\varphi(t,\bm{Z}_t)^\top \bm{B}\bar{\bm{h}}_t(\bm{Z}_t,\bm{X}_t)\bigr],
 \end{aligned}
 \label{eq:memory-residual-conditioning}
\end{equation}
where $\bar{\bm{h}}_t(\bm{z},\bm{x})=\sum_r q_t(r\mid \bm{z},\bm{x})\bar{\bm{h}}_{t,r}(\bm{z})$ and the
expectation is over $\bm{X}_t\sim\rho_t$,
$\bm{Z}_t\sim q_{\bm{\theta}}(\cdot\mid \bm{X}_t,t)$.

Applying the spatial test $\psi_j$ on $I_k=[t_k,t_{k+1}]$, as in
\Eqref{eq:interval-exact-v6}, gives
$\mathcal R_D^{\mathrm M}(\bm{\Theta};\varphi_{kj})
=y_{kj}-\langle\bm{\Theta},\bm{H}_{kj}\rangle_F-c_{kj}$, where
\begin{equation}
 \begin{aligned}
 y_{kj}&=\mu_{t_{k+1}}[\psi_j]-\mu_{t_k}[\psi_j],\qquad
 \bm{H}_{kj}=\int_{I_k}\mathbb E[
       \nabla\psi_j(\bm{Z}_t)\widetilde{\bm{Z}}_t^\top]\mathrm{d}t,\\
 c_{kj}&=\int_{I_k}\mathbb E[
       \nabla\psi_j(\bm{Z}_t)^\top \bm{B}\bar{\bm{h}}_t(\bm{Z}_t,\bm{X}_t)]\mathrm{d}t,
 \qquad y^{\mathrm M}_{kj}=y_{kj}-c_{kj}.
 \end{aligned}
 \label{eq:memory-weak-design}
\end{equation}
We estimate the expectations from encoded cells and approximate the time
integrals using the trapezoidal rule, as in KoopCell
(\Eqref{eq:linear-empirical-design}).
Stacking the interval--test pairs gives the residual vector
$\bm{y}^{\mathrm M}-\bm{G}\bm{a}$, with $\bm{a}=\operatorname{vec}(\bm{\Theta})$ and
$\bm{G}_{(k,j),:}=\operatorname{vec}(\bm{H}_{kj})^\top$.
Similarly, we solve the regularized regression used in KoopCell,
replacing $\bm{y}$ with $\bm{y}^{\mathrm M}=\bm{y}-\bm{c}$:
\begin{equation}
 \widehat{\bm{a}}=\operatorname*{arg\,min}_{\bm{a}}
 \left\{\tfrac12\|\bm{y}^{\mathrm M}-\bm{G}\bm{a}\|_{\bm{W}}^2
             +\tfrac\lambda2\|\bm{a}\|_2^2\right\}
 =(\bm{G}^\top \bm{W}\bm{G}+\lambda \bm{I})^{-1}\bm{G}^\top \bm{W}\bm{y}^{\mathrm M}.
 \label{eq:memory-ridge}
\end{equation}
We evaluate the path probabilities $q_t(r\mid \bm{z},\bm{x})$ and conditional
memory means $\bar{\bm{h}}_{t,r}(\bm{z})$ under the current model and hold them
fixed together with the memory coupling matrix $\bm{B}$ when updating
$\bm{\Theta}$.

Finally, we combine this closed-form update of $\bm{\Theta}$ with gradient-based
learning of the representation and memory dynamics, following the
alternating training scheme adopted in KoopCell (Algorithm~\ref{alg:linear-training}).
We first jointly pretrain the VAE and {state classifier} by minimizing
$\mathcal L_{\mathrm{pre}}+\lambda_c{\mathcal L_{\mathrm{state}}}$, with
$\mathcal L_{\mathrm{pre}}$ defined in
\Eqref{eq:linear-pretraining-loss}. In Stage~2, we alternate between the
following two updates.
First, we form the weak statistics, solve
\Eqref{eq:memory-ridge}, and update
$\bm{\Theta}\leftarrow\alpha\bm{\Theta}+(1-\alpha)\widehat{\bm{\Theta}}$.
Second, with $\bm{\Theta}$ fixed, we update
$\bm{\eta}=(\bm{\theta},\bm{\phi},\bm{\vartheta},\bm{\omega})$ by minimizing
\begin{equation}
 \mathcal L_{\mathrm M}
 =\mathcal L_{\mathrm{snap}}+\mathcal L_{\mathrm{pop}}
 +\lambda_R\mathcal L_{\mathrm{path}}
 +\lambda_{\mathrm{weak}}\mathcal L_{\mathrm{weak}},\qquad
 \mathcal L_{\mathrm{weak}}=\|\bm{y}^{\mathrm M}-\bm{G}\bm{a}\|_{\bm{W}}^2.
 \label{eq:memory-training-objective}
\end{equation}

We summarize the training procedure in Algorithm~\ref{alg:memory-training}.
{The routine $\operatorname{StateLoss}$ evaluates the supervised
classification loss $\mathcal L_{\mathrm{state}}$ in
\Eqref{eq:memory-snapshot-loss}.}
In $\operatorname{Infer}$, we compute path probabilities $\bm{q}$ using
\Eqref{eq:memory-path-probabilities}, and Gaussian moments and conditional
coefficients $\mathcal C$ using \Eqref{eq:memory-transition} and
\Eqref{eq:memory-conditioning}.
In $\operatorname{MemoryStats}$, we form $(\bm{G},\bm{y}^{\mathrm M},\bm{M})$ from
\Eqref{eq:memory-weak-design} and \Eqref{eq:linear-empirical-gram}
using empirical averages and the trapezoidal rule.
For the minibatch weak loss, we hold $(\bm{q},\mathcal C,\bm{B})$ fixed while
retaining gradients through $\bm{z}$.
Rollouts and the path consistency loss follow \Eqref{eq:memory-transition} and
\Eqref{eq:memory-path-loss}, respectively.
The notation $\operatorname{sg}$ and $\operatorname{PCG}$ follows
Algorithm~\ref{alg:linear-training}.

\section{Spectral Recovery Experiments}
\label{app:toy-models}

This section gives the data-generation and evaluation protocols for the
two deterministic toy models in Section~\ref{subsec:toy-spectral-recovery}.
Let $\Phi_t$ denote the flow of the
corresponding dynamical system and let $\rho_0$ be its initial population
distribution.  For each toy model, the dataset consists of empirical
population snapshots $\{\mathcal X_{t_k}\}_{k=0}^{K}$, where each
$\mathcal X_{t_k}=\{\bm{x}_{t_k}^{(i)}\}_{i=1}^{N_k}$ contains independent samples
from the time marginal $\rho_{t_k}=(\Phi_{t_k})_\#\rho_0$.  To generate the
empirical snapshot at time $t_k$, we independently draw a population of initial states
from $\rho_0$, solve the
governing ODE from time $0$ to $t_k$, and retain only the terminal states. Repeating this construction independently at every observation time
removes cross-time cell correspondence and yields the unpaired population
snapshots used for training and evaluation.

\subsection{Toy 1: finite-dimensional nonlinear flow}
\label{app:toy1}

\paragraph{Dynamics and snapshot data generation.}
We consider the two-dimensional nonlinear flow
\begin{equation}
\dot x_1=\mu x_1,\qquad
\dot x_2=\lambda(x_2-x_1^2),
\label{eq:toy1-flow}
\end{equation}
with $(\mu,\lambda)=(-0.2,-1)$.  The origin $(0,0)$ is globally attracting, and the
invariant parabola $x_2=\frac{\lambda}{\lambda-2\mu}x_1^2
=\frac53x_1^2$ is an attracting slow manifold.  Indeed, the observables
\begin{equation}
\psi_1(\bm{x})=x_1,\qquad
\psi_2(\bm{x})=x_2-\frac53x_1^2,\qquad
\psi_3(\bm{x})=x_1^2
\label{eq:toy1-eigenfunctions}
\end{equation}
form a finite-dimensional Koopman-invariant subspace and satisfy
$\dot\psi_j=\lambda_j\psi_j$ with
$(\lambda_1,\lambda_2,\lambda_3)=(-0.2,-1,-0.4)$.  The transverse coordinate
$\psi_2$ therefore decays faster than the coordinates that parameterize the
slow manifold.

\begin{figure}[H]
  \centering
  \includegraphics[width=\textwidth]{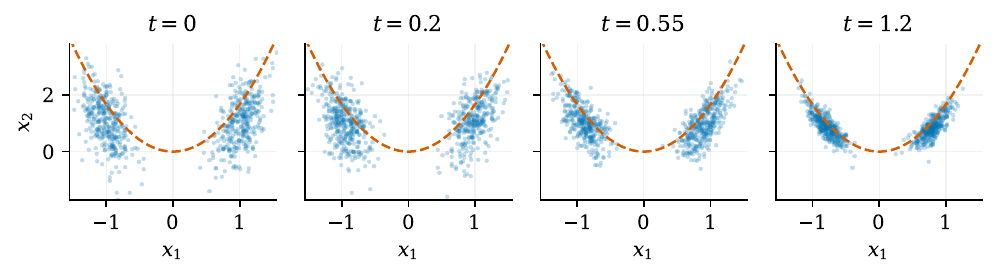}
  \caption{Toy 1: generated population snapshots. Each panel is an
  independently generated sample from the corresponding time marginal. The
  dashed curve denotes the attracting slow manifold.}
  \label{fig:toy1-snapshots}
\end{figure}

We define $\rho_0$ by
$x_1(0)\sim\frac12\mathcal N(-1,0.2^2)+
\frac12\mathcal N(1,0.2^2)$ and
$x_2(0)=x_1(0)^2+\eta$, where $\eta\sim\mathcal N(0,0.8^2)$.  For each selected
time, 3,000 new initial states are sampled from this law and advanced
with the analytic solution of \Eqref{eq:toy1-flow}.  The nine
training snapshots are observed at
$t\in\{0,0.1,0.2,0.3,0.4,0.55,0.7,0.9,1.2\}$; independent target snapshots
are generated every $0.2$ time units from $t=1.4$ to $t=4.0$ for
extrapolation.  Figure~\ref{fig:toy1-snapshots} illustrates the generated snapshots, where we observe a rapid
contraction toward the slow manifold.

\paragraph{Experimental settings.}
We consider three settings.  The learned-observable and high-dimensional
settings are briefly described in
Section~\ref{subsec:toy-spectral-recovery}, whereas the known-observable setting
is presented only in this appendix due to space constraints.  The first is the
\emph{known-observable} setting.  Here we use the analytically specified
observable map $\bm{z}=(x_1,x_2,x_1^2)^\top$, whose components span a
finite-dimensional Koopman-invariant subspace in which the dynamics are exactly
linear.  We therefore fix $\bm{z}$ as the encoder and estimate only the linear
generator, thereby isolating spectral recovery from representation learning.
Second, in the \emph{learned-observable} setting, the model receives
only the two physical coordinates $(x_1,x_2)$ and jointly learns a three-dimensional
observable and its corresponding linear propagator.  
Third, in the \emph{high-dimensional} setting, we
generate 64-dimensional observations by applying a fixed injective map:
\begin{equation}
\bm{Y}=\operatorname{softplus}(\bm{W}\bm{x}+\bm{c})+\bm{\epsilon}_Y\in\mathbb R^{64},
\qquad \bm{\epsilon}_Y\sim\mathcal N(\bm{0},0.03^2 \bm{I}).
\label{eq:toy1-highdim-observation}
\end{equation}
Here $\bm{x}=(x_1,x_2)^\top$ is the physical state and
$\bm{W}\in\mathbb R^{64\times2}$ has full column rank.
Only $\bm{Y}$ is provided to either model; $\bm{x}$, $\bm{W}$, and $\bm{c}$ remain unknown. This
setting evaluates representation learning and population prediction in the
observed 64D space, and we do not report an inferred spectrum for it.

The learned-observable models use 2,048 multiscale random Fourier test
functions and a three-dimensional latent space.  KoopCell alternates the weak
generator solve with representation updates for 500 epochs.  scNODE receives
the same snapshots and latent dimension, replacing the linear latent generator
with an autonomous neural ODE.  Table~\ref{tab:toy1-spectrum-detail} reports
the recovered spectrum only for the two settings in which the physical
observable structure is available for an unambiguous comparison.  The linear observable
alignment used to compare learned and analytic eigenfunctions is applied only
after training and does not enter either the objective or the rollout.

\begin{table}[H]
\centering
\caption{Toy 1 spectral recovery with known and learned observables.
Parentheses contain absolute eigenvalue errors.}
\label{tab:toy1-spectrum-detail}
\small
\setlength{\tabcolsep}{5pt}
\begin{tabular}{ccc}
\toprule
$\lambda_\star$ & Known observable & Learned observable \\
\midrule
$-0.2$ & $-0.1983\;(0.0017)$ & $-0.1952\;(0.0048)$ \\
$-1.0$ & $-0.9949\;(0.0051)$ & $-0.9774\;(0.0226)$ \\
$-0.4$ & $-0.3893\;(0.0107)$ & $-0.4227\;(0.0227)$ \\
\bottomrule
\end{tabular}
\end{table}

\begin{figure}[p]
  \centering
  {\fontsize{12}{14}\selectfont Predicted snapshots with KoopCell\par}
  \vspace{0.1em}
  \includegraphics[width=\textwidth,trim=0 0 0 22bp,clip]{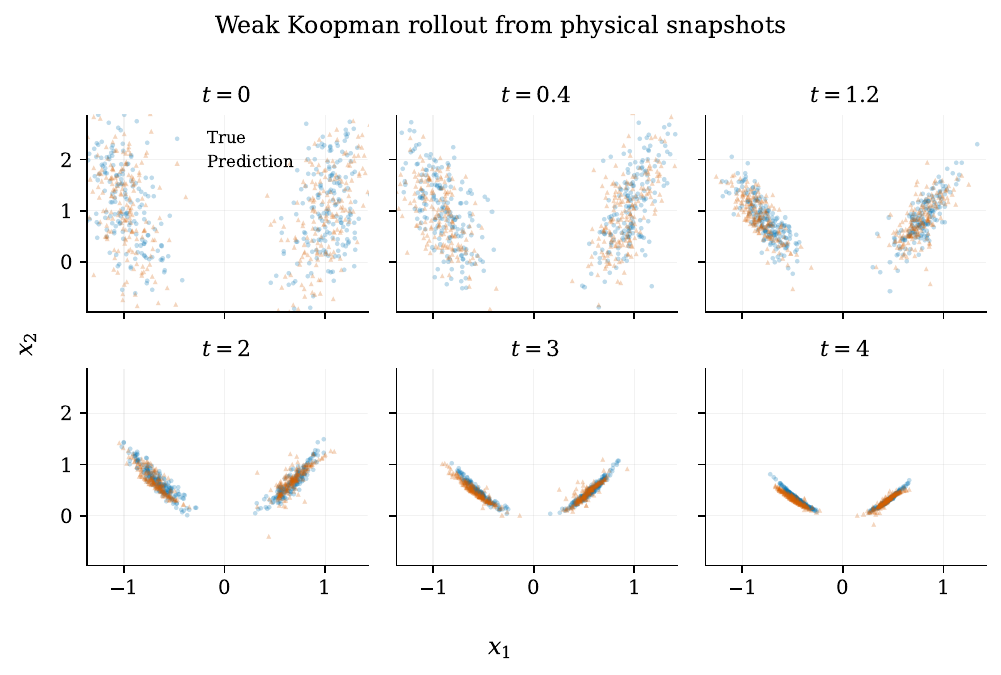}
  \vspace{-0.2em}
  {\fontsize{12}{14}\selectfont Predicted snapshots with scNODE\par}
  \vspace{0.1em}
  \includegraphics[width=\textwidth,trim=0 0 0 22bp,clip]{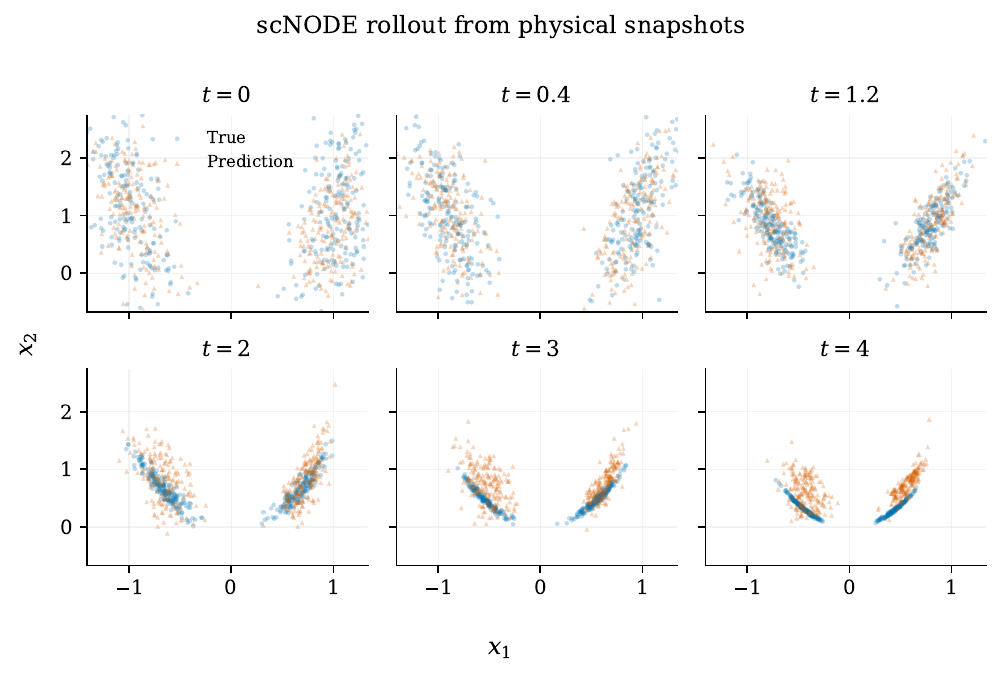}
  \caption{Predicted snapshots with KoopCell/scNODE. Both rollouts start from
  the observed Toy 1 population at $t=0$. Blue circles denote independently
  generated target snapshots, and orange triangles denote predictions. Times
  after $t=1.2$ are extrapolations.}
  \label{fig:toy1-rollouts}
\end{figure}

\paragraph{Rollout and evaluation.}
Each rollout starts from the observed population at $t=0$.  KoopCell encodes
this population, advances every latent state with the learned latent linear dynamics, and
decodes at the requested time; scNODE instead integrates its learned latent
neural ODE.  At each horizon, we compute the sliced Wasserstein distance (SWD)
between the predicted population and the independently generated target
snapshot.  We use physical-space SWD for the learned-observable setting and
native observation-space SWD for the 64D setting, and average the metric over
all extrapolation times. Figure~\ref{fig:toy1-rollouts} compares the population
rollouts of KoopCell and scNODE. Predicted distributions from KoopCell agree
more closely with the ground-truth distributions.

\clearpage
\subsection{Toy 2: damped Duffing flow}
\label{app:toy2}

\paragraph{Dynamics and snapshot data generation.}
We consider the damped, unforced Duffing system studied by
\citet{otto2019linearly},
\begin{equation}
\dot q=p,\qquad \dot p=-0.5p+q-q^3.
\label{eq:toy2-duffing}
\end{equation}
Its energy
$H(q,p)=\frac12p^2-\frac12q^2+\frac14q^4$ satisfies
$\frac{\mathrm{d}}{\mathrm{d}t}H(q(t),p(t))=-0.5p(t)^2\leq0$, so the flow is dissipative.  The
origin is a saddle point, while $(q,p)=(\pm1,0)$ are stable spiral equilibria.  The
linearization at either stable equilibrium has eigenvalues
$(-1\pm\sqrt{31}i)/4$.  The corresponding global Koopman eigenfunctions do not
have closed-form expressions.  As described by \citet{otto2019linearly}, the
magnitude and phase of the complex eigenfunction provide action--angle-like
coordinates over each basin of attraction: the magnitude vanishes at the
stable equilibrium, while the phase tracks the damped rotation.  A further
eigenfunction with eigenvalue zero is constant along trajectories and takes
distinct values in the two basins of attraction.

For the initial population, we assign equal mass to two connected annular sectors, one in
each basin.  For a sign $s\in\{-1,+1\}$, we sample
$\theta\sim\mathcal U[-2.20,1.10]$ and
$r\sim\mathcal U[0.16,0.48]$, set
$(q,p)=(s(1+r\cos\theta),-1.35sr\sin\theta)$, and retain samples with
$H(q,p)<-0.025$.  This construction covers a range of intrabasin amplitudes
and phases while keeping the population away from the energy barrier.  Each
time marginal contains 3,000 independently resampled initial states propagated
with a high-accuracy numerical solver.  Training uses the ten marginals at
$t=0,0.5,\ldots,4.5$; evaluation uses independently generated extrapolation
targets at $t\in\{5,6,8,10,12\}$.

\begin{figure}[H]
  \centering
  \includegraphics[width=\textwidth]{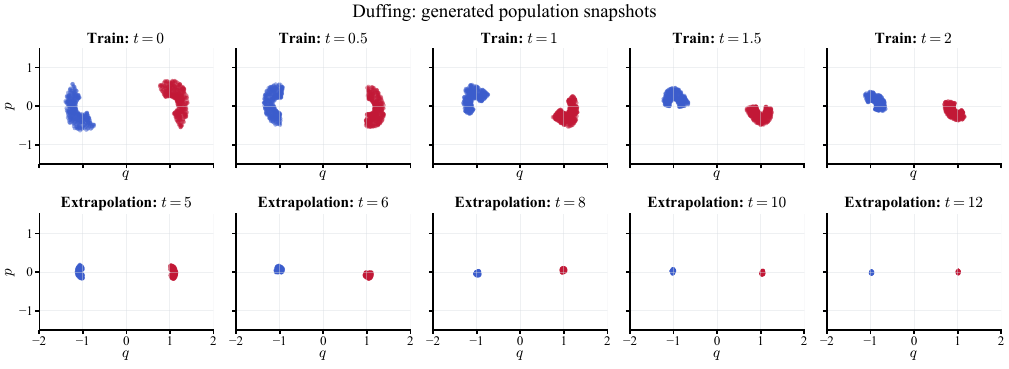}
  \caption{Toy 2 Duffing System: generated population snapshots. The first row shows
  training distribution marginals and the second row shows extrapolation ground truth. Each
  panel is generated independently from the same initial population law.}
  \label{fig:toy2-snapshots}
\end{figure}

\paragraph{Experimental settings.}
We use two settings.  In the \emph{learned-observable} setting, KoopCell receives
the physical snapshots $(q,p)$ and jointly learns a three-dimensional
observable, its linear generator, and a nonlinear decoder.  Both the encoder
and decoder are two-layer MLPs with hidden widths $64$--$64$, and training uses
2,048 multiscale random Fourier test functions for 500 epochs.  Because the
relevant eigenfunctions are not available analytically, the objective of this
experiment is to determine whether snapshot data recover their characteristic
behavior: a complex mode encoding decaying intrabasin amplitude and phase, and
a near-zero mode that remains constant within each basin and distinguishes the
two long-term outcomes.  Figure~\ref{fig:toy2-eigenmodes} visualizes these
learned modes, and Table~\ref{tab:toy2-spectrum-detail} compares their
eigenvalues with the values implied by the Duffing dynamics.

In the \emph{high-dimensional} setting, the same physical marginals are
observed through a fixed injective map
\begin{equation}
\bm{Y}=\operatorname{softplus}(\bm{W}[q,p]^\top+\bm{c})+\bm{\epsilon}_Y\in\mathbb R^{64},
\qquad \bm{\epsilon}_Y\sim\mathcal N(\bm{0},0.02^2 \bm{I}).
\label{eq:toy2-highdim-observation}
\end{equation}
Neither method receives $(q,p)$ or the observation map during training.  Both
KoopCell and scNODE use a three-dimensional latent state and are evaluated by
their population predictions in the native 64D observation space. 

\begin{table}[H]
\centering
\caption{Toy 2 spectral recovery in the learned-observable setting.}
\label{tab:toy2-spectrum-detail}
\small
\setlength{\tabcolsep}{4pt}
\begin{tabular}{lrrr}
\toprule
Mode & Reference $\lambda$ & Estimated $\widehat\lambda$ & Absolute error \\
\midrule
Oscillatory pair & $-0.2500\pm1.3919i$ & $-0.2979\pm1.4811i$ & $0.1012$ \\
Basin mode & $0$ & $0.0105$ & $0.0105$ \\
\bottomrule
\end{tabular}
\end{table}

\begin{figure}[H]
  \centering
  \includegraphics[width=\textwidth]{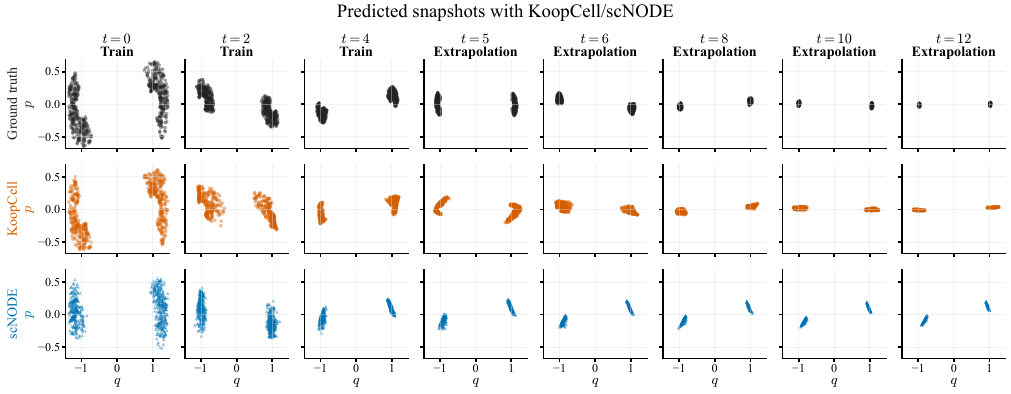}
  \caption{Predicted Duffing snapshots with KoopCell/scNODE. Rows show
  independently generated ground truth, KoopCell predictions, and scNODE
  predictions; columns progress from training times to extrapolation through
  $t=12$.}
  \label{fig:toy2-rollout}
\end{figure}

\paragraph{Rollout and evaluation.}
We follow the same protocol as in the Toy 1 experiment.
Figure~\ref{fig:toy2-rollout} compares the population rollouts produced by
KoopCell and scNODE.  KoopCell more faithfully reproduces the target population
over the full prediction horizon through $t=12$.

\subsection{Rollout-error accumulation}
\label{app:toy-rollout-error}

\begin{figure}[H]
  \centering
  \includegraphics[width=\textwidth]{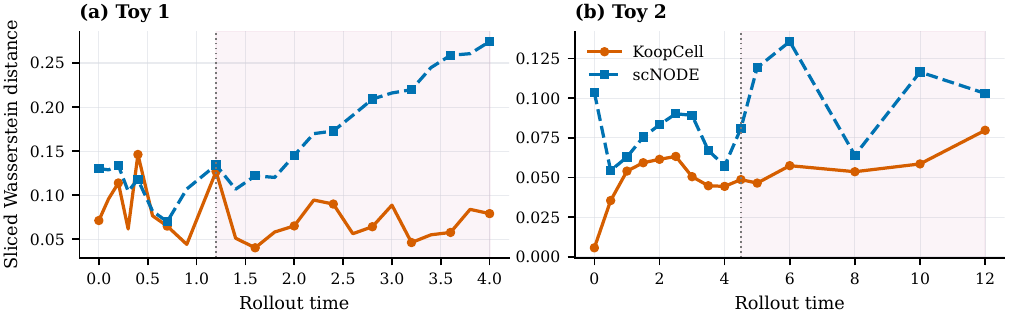}
  \caption{Accumulation of population rollout error. Left: Toy 1. Right: Toy
  2. Each curve reports the per-time SWD for a rollout initialized from the
  observed population at $t=0$.}
  \label{fig:toy-rollout-error}
\end{figure}

Figure~\ref{fig:toy-rollout-error} compares the rollout errors of KoopCell and
scNODE over time for both toy models. The vertical dotted line marks the final
training snapshot, and the shaded region denotes extrapolation. On Toy~1,
scNODE's error increases steadily beyond the training window, whereas KoopCell's error
remains low throughout extrapolation. On Toy~2, KoopCell also maintains lower
rollout errors than scNODE throughout the evaluated horizon.

\section{Branching Dynamics Experiments}
\label{app:toy-branching}

This section gives the data-generation, training, and evaluation protocols
for the branching experiment in Section~\ref{subsec:toy-branching}.

\paragraph{Dynamics and snapshot data generation.}
We generate a noisy pitchfork process with progression coordinate $s$ and
branch coordinate $b$:
\begin{equation}
  \mathrm{d}s=0.21\,\mathrm{d}t,\qquad \mathrm{d}b=1.5(2.4sb-b^3)\,\mathrm{d}t+0.15\,\mathrm{d}W_t,
  \label{eq:branching-sde}
\end{equation}
where $s_0\sim\mathcal N(-1.05,0.07^2)$ and
$b_0\sim\mathcal N(0,0.04^2)$ are independent. As $s$ becomes positive,
the central equilibrium of the branch drift loses stability and two stable
branches emerge. We simulate 4,000 trajectories with an Euler--Maruyama
step of $0.0025$, updating $s$ before evaluating the branch drift.
To simulate high-dimensional data from this two-dimensional process, we use
the nonlinear observation map
$\bm{x}=2[\bm{Q}\bm{u}+0.2\{\tanh(\bm{W}\bm{u}+\bm{c})-\tanh(\bm{c})\}]+\bm{\epsilon}\in\mathbb R^{64}$,
where $\bm{u}=(s/1.2,b/1.8)^\top$, $\bm{Q}^\top \bm{Q}=\bm{I}_2$, $\|\bm{W}\|_2=1$,
$c_j\sim\mathcal U[-0.5,0.5]$, and
$\bm{\epsilon}\sim\mathcal N(\bm{0},0.09^2\bm{I}_{64})$ is independent observation noise.
Each snapshot is independently shuffled, and training uses only
unpaired populations, without trajectory correspondences or branch labels.
The nine training times are
$0,1.2,2.4,3.6,4.8,5.8,6.8,7.8,9.2$; interpolation is evaluated at
$4.2,8.3$ and extrapolation at $9.5,10$.

\paragraph{Models and training.}
At each training time, 3,000 cells are used for fitting, 500 for validation,
and 500 are reserved for testing. We use the means and standard deviations
estimated from fitting cells to standardize observations, and rescale time
as $\tau=t/9.2$. Both models use a Gaussian encoder with a 10-dimensional
latent state. The encoder and decoder each have two hidden layers of width
128, with SiLU and tanh activation functions, respectively. Decoder weights
are spectrally normalized with output scale 4 to control the decoder's
Lipschitz constant in both models and thus limit the amplification of latent
differences during decoding. KoopCell uses the linear latent
flow $\dot{\bm{Z}}_\tau=\bm{A}\bm{Z}_\tau+\bm{b}_z$ with a stable parameterization of $\bm{A}$.
KoopCell-M uses the augmented dynamics in
\Eqref{eq:memory-augmented-dynamics} (Appendix~\ref{app:memory-dynamics}),
with visible state $\bm{Z}_\tau\in\mathbb R^{10}$ and memory state
$\bm{H}_\tau\in\mathbb R^{24}$. In this experiment, $\bm{\Sigma}_h=\bm{0}$ and the
hidden bias $\bm{b}_h$ is shared across components.
Following the conditional-mixture construction in
\Eqref{eq:memory-initial}, we use two hidden components with
$R\sim\operatorname{Cat}(\pi_1,\pi_2)$ and
$\bm{H}_0\mid \bm{Z}_0=\bm{z},R=r\sim\mathcal N(\bm{m}_r(\bm{z}),\bm{S}(\bm{z}))$.
Here the mixture probabilities $\pi_r$ are learned, and neural networks
parameterize the conditional means $\bm{m}_r(\bm{z})$ and diagonal covariance $\bm{S}(\bm{z})$.

To isolate the effect of the dynamical model from that of the
generator-estimation procedure, we use a common gradient-based training
protocol for both models. For this controlled comparison, the closed-form
generator updates in Algorithms~\ref{alg:linear-training}
and~\ref{alg:memory-training} are replaced by joint optimization of the
representation and dynamics, using the same reconstruction and
population-matching objectives. We first pretrain a shared VAE for 2,400
Adam steps (learning rate $0.002$, 256 cells per time), then initialize
both models from this VAE and train each for 8,000 steps, with 128 cells
per time and a cosine-decayed learning rate starting at $0.001$.

\paragraph{Evaluation.}
After training, we encode the initial observed population, propagate its
latent states with the learned dynamics, and decode the predicted populations
at the evaluation times. We compare these predictions with the corresponding
ground-truth populations using the following metrics:
\begin{itemize}
  \item \textbf{Sliced Wasserstein distance (SWD).} The root mean square of
  one-dimensional $W_2$ distances along 512 shared random directions in
  standardized 64D observation space.
  \item \textbf{Energy distance.} The energy distance between predicted and
  observed populations in the same standardized space, normalized by
  $\sqrt{64}=8$.
  \item \textbf{Squared maximum mean discrepancy (MMD$^2$).} We use an RBF
  kernel with bandwidth set to the median pairwise distance among
  fitting cells in standardized observation space.
  \item \textbf{Branch-coordinate $W_1$.} After branching, the two populations
  occupy positive and negative values of the branch coordinate $b$.
  We fit a linear probe $\widehat b(\bm{x})=\bm{w}^\top \bm{x}+c$ to training observations
  and their known $b$ values, then apply it to the predicted 64D observations.
  The $W_1$ distance between the estimated and true $b$ distributions measures
  discrepancies in branch locations, widths, and relative population sizes.
  \item \textbf{Central mass error.} The interval $|b|<0.5$ defines a fixed
  central region around the initial, unbranched population. Using the same
  linear probe, we compute the absolute difference between predicted and
  true fractions of cells in this region. This measures whether the model
  reproduces the amount of population remaining between the two branches.
\end{itemize}
Table~\ref{tab:branching-detail} reports prediction errors averaged over the
corresponding training, interpolation, and extrapolation times.

\begin{table}[H]
  \centering
  \caption{Detailed branching prediction errors (mean $\pm$ standard
  deviation across three optimization seeds). Lower is better; the better
  mean is shown in bold.}
  \label{tab:branching-detail}
  \scriptsize
  \setlength{\tabcolsep}{3pt}
  \renewcommand{\arraystretch}{1.12}
  \input{figures/branching/table_detail.tex}
\end{table}

\section{Single-Cell Experimental Details}
\raggedbottom
\begingroup
\setlength{\intextsep}{4pt}
\setlength{\abovecaptionskip}{4pt}

\subsection{Single-cell datasets and preprocessing}
\label{app:single-cell-data}

We use the processed benchmark released with scNODE~\citep{zhang2024scnode}
on Figshare\footnote{\url{https://doi.org/10.6084/m9.figshare.25601610.v1}.},
which contains three high-dimensional scRNA-seq datasets.
ZB comprises 38,731 cells from 12 stages of zebrafish embryogenesis,
spanning 3.3--12 hours post fertilization~\citep{doi:10.1126/science.aar3131}.
DR contains 27,386 cells from 11 overlapping collection windows spanning
0--20 hours of Drosophila embryogenesis~\citep{doi:10.1126/science.abn5800}.
SC follows the reprogramming of mouse embryonic fibroblasts toward induced
pluripotent stem cells~\citep{schiebinger2019optimal}; we use the reduced
benchmark subset of 23,619 cells in 19 temporal groups, formed by pooling
samples collected within each integer day from day 0 to day 18.
Table~\ref{tab:single-cell-counts} shows the number of cells at each time
point in each dataset.

We consider three prediction tasks of increasing difficulty:
\emph{easy} withholds intermediate snapshots for interpolation;
\emph{medium} withholds the final snapshots for extrapolation; and
\emph{hard} combines intermediate and final held-out snapshots, leaving
sparser observations for training. Figure~\ref{fig:single-cell-task-splits}
shows the training and held-out snapshots for each dataset and task.

For each task, we use the 2,000 highly variable genes (HVGs) selected in
the benchmark. For ZB and DR, we normalize each cell's counts over these
genes to a total of $10^4$ and apply a log transformation:
$x_{ig}=\log(1+10^4c_{ig}/\sum_{h=1}^{2000}c_{ih})$, where $c_{ig}$ is
the count for gene $g$ in cell $i$. For SC, we use the normalized expression
values provided in the benchmark. We index the observed populations in
chronological order as $t=1,\ldots,T$.

Figure~\ref{fig:single-cell-datasets} visualizes all observed cells using the
hard-task gene sets. For each dataset, we reduce the expression profiles to
50 principal components and then construct a two-dimensional UMAP embedding,
with cells colored by snapshot index.

\begin{figure}[!htbp]
  \centering
  \includegraphics[width=\textwidth]{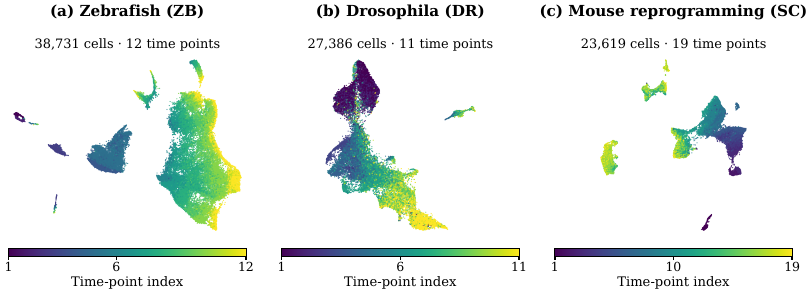}
  \caption{Observed populations in (a) ZB, (b) DR, and (c) SC.
  Each point represents a cell in a UMAP embedding of gene expression,
  colored by snapshot index.}
  \label{fig:single-cell-datasets}
\end{figure}

\begin{table}[htbp]
  \centering
  \caption{Number of observed cells at each time point. Indices start at
  $t=1$; dashes indicate indices beyond the observed horizon.}
  \label{tab:single-cell-counts}
  \scriptsize
  \setlength{\tabcolsep}{1.5pt}
  \begin{tabular*}{\textwidth}{@{\extracolsep{\fill}}l*{19}{r}@{}}
    \toprule
    Dataset & 1 & 2 & 3 & 4 & 5 & 6 & 7 & 8 & 9 & 10 & 11 & 12 & 13 & 14 & 15 & 16 & 17 & 18 & 19 \\
    \midrule
    ZB & 311 & 200 & 1158 & 1467 & 5716 & 1026 & 4101 & 6178 & 5442 & 7114 & 1614 & 4404 & -- & -- & -- & -- & -- & -- & -- \\
    DR & 3944 & 2801 & 947 & 3139 & 2074 & 2147 & 3311 & 1692 & 1856 & 1491 & 3984 & -- & -- & -- & -- & -- & -- & -- & -- \\
    SC & 800 & 560 & 1371 & 1413 & 1608 & 1377 & 1153 & 1156 & 2420 & 886 & 732 & 712 & 747 & 708 & 1444 & 2061 & 2106 & 1622 & 743 \\
    \bottomrule
  \end{tabular*}
\end{table}

\begin{figure}[H]
  \centering
  \includegraphics[width=\textwidth]{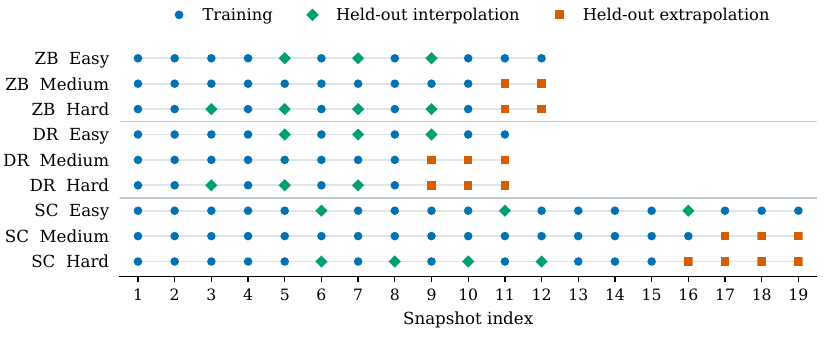}
  \caption{Training and held-out snapshots for each prediction task.}
  \label{fig:single-cell-task-splits}
\end{figure}

\subsubsection{Population prediction and evaluation}
\label{app:single-cell-protocol}

For a consistent evaluation across methods, KoopCell, KoopCell-M, and all
baselines each generate 2,000 cells at every held-out time point. We compare
these predicted populations with all observed cells at the corresponding
time point in the processed 2,000-gene space.

Every task retains the earliest observed population as the source for
prediction. For KoopCell and KoopCell-M, we uniformly sample 2,000 cells
from this population, with replacement when fewer than 2,000 cells are
available, and draw their initial latent states from the encoder posterior.
KoopCell propagates these states with its learned linear dynamics.
KoopCell-M additionally samples a path $R$ and memory state $\bm{H}_0$ for each
initial latent state, then propagates the augmented state while keeping
$R$ fixed. We decode the visible latent states at each evaluation time to
obtain the predicted gene expression distribution.

In the single-cell experiments, we evaluate predictions using
the quadratic optimal transport cost
\[
\frac12 W_2^2(\mu,\nu)
=
\inf_{\pi\in\Pi(\mu,\nu)}
\int \frac12\|\bm{x}-\bm{y}\|^2
\,\mathrm{d}\pi(\bm{x},\bm{y}).
\]
We report and summarize the results as
mean $\pm$ sample standard deviation over three training seeds.

\endgroup

\subsection{Implementation of Baseline Models} \label{appendixsubsec: implementation_baselines}

{We compare KoopCell and KoopCell-M with six representative models for modeling temporal single-cell population dynamics.} For \href{https://github.com/rsinghlab/scNODE}{scNODE} and \href{https://github.com/YigitBalik/LGP-OT}{LGP-OT}, the released repositories provide tuned configurations for the same datasets and prediction tasks, so we use those configurations directly. 
We select the hyperparameters of MIOFlow, PRESCIENT, PI-SDE and VGFM by training-only three-fold cross-validation, {following the protocol adopted in the scNODE benchmark.} Specifically, at every training time point, we randomly partition the cells into three approximately equal subsets, fit the model on two subsets, and validate on the remaining subset. We rotate the held-out subset over the three folds and select the configuration with the lowest mean validation optimal-transport distance across folds and training time points. The test time points defined by each prediction task remain inaccessible during model selection. After selecting a configuration, we retrain the model using all training cells and evaluate it on the held-out test time points.

\paragraph{\href{https://github.com/rsinghlab/scNODE}{scNODE}.} scNODE combines a variational autoencoder with a latent neural ordinary differential equation (ODE), and jointly optimizes the latent representation and continuous dynamics using population-level distribution matching and dynamic regularization~\citep{zhang2024scnode}. We use the task-specific settings released as \href{https://github.com/rsinghlab/scNODE/blob/main/benchmark/BenchmarkUtils.py}{\texttt{tunedOurPars}}: the latent dimension is fixed as 50, while the encoder, decoder, and ODE-network architectures depend on the dataset and prediction task. Because these settings were tuned on the same nine dataset--task combinations considered in our real world scRNA-seq experiments (section \ref{subsec: realworld_cellular_dynamics}), we do not perform an additional search.

\paragraph{\href{https://github.com/YigitBalik/LGP-OT}{LGP-OT}.} LGP-OT represents population dynamics with a heteroscedastic Gaussian process in a learned latent space, approximates the process with Hilbert-space basis functions, and trains a decoder by matching generated and observed cell populations with a Sinkhorn optimal-transport objective~\citep{balik2026modeling}. We adopt the released task-specific \href{https://github.com/YigitBalik/LGP-OT/blob/main/scripts/utils.py}{\texttt{modelParams}} configurations, which use a 32-dimensional latent space and select the decoder architecture and number of basis functions for each dataset and task. We therefore do not retune LGP-OT.

\paragraph{\href{https://github.com/KrishnaswamyLab/MIOFlow}{MIOFlow}.} MIOFlow first learns a geometry-preserving representation with a geodesic autoencoder and then fits neural-ODE dynamics using optimal-transport and density-regularization losses~\citep{huguet2022manifold}. We select the geodesic-autoencoder dimension from $\{10,50,100\}$, the partition-encoder architecture from $\{[50,100],[50,100,100]\}$, the ODE architecture from $\{[16],[16,16],[16,32,16]\}$, and the density-loss coefficient from $\{1,25,50,75,100\}$ using three-fold cross-validation.

\paragraph{\href{https://github.com/gifford-lab/prescient}{PRESCIENT}.} PRESCIENT learns a neural potential-energy landscape whose gradient defines the drift of a stochastic process, thereby generating cell-population trajectories in a fixed 50-dimensional PCA space~\citep{yeo2021generative}. We tune the Gaussian-noise standard deviation over $[0,1]$, the energy-regularization strength over $[0,0.1]$, the gradient-clipping threshold over $[0,1]$, and the number of hidden layers over $\{1,2,3\}$ by three-fold cross-validation.

\paragraph{\href{https://github.com/QiJiang-QJ/PI-SDE}{PI-SDE}.} PI-SDE learns a physics-informed stochastic differential equation in a fixed 50-dimensional PCA space by combining distribution matching with a potential-based action regularizer~\citep{10.1093/bioinformatics/btae400}. We fix the potential function to a two-layer fully connected network with 400 units per layer and Softplus activations. We tune only the diffusion parameterization among a fixed scalar (\texttt{const}), a learnable constant vector (\texttt{const\_param}), and a state-dependent neural network (\texttt{Mlp}); the scalar initialization is fixed to $0.1$. We also fix the learning rate to $0.005$, the action-regularization coefficient to $0.5$, and the sampled-cell fraction per training time point to $0.1$, and select the diffusion type by three-fold cross-validation.

\paragraph{\href{https://github.com/DongyiWang-66/VGFM}{VGFM}.} VGFM jointly models cellular velocity and growth through
flow matching and unbalanced optimal transport~\citep{wang2026joint}.
We fit PCA using training cells only and explore raw and standardized PCA
representations, dimensions in $\{50,100\}$, Tanh and LeakyReLU activation
functions, entropy regularization $\epsilon\in\{0.01,0.03\}$, and
source-marginal relaxation $\tau\in\{5,10\}$.
The reported results use standardized PCA with 50 components for the easy
and hard tasks and 100 components for the medium task. All reported
configurations use Tanh activations, $\epsilon=0.01$, and $\tau=10$.
The velocity and growth networks have five and three hidden layers,
respectively, with 512 units per hidden layer. Training comprises 2,000
pretraining epochs followed by 50 trajectory-training epochs, with learning
rates $10^{-3}$ and $10^{-4}$, respectively.

KoopCell(-M), scNODE, LGP-OT, MIOFlow, PRESCIENT, PI-SDE and VGFM are all implemented and trained on NVIDIA Tesla V100 GPUs with 32\,GB of memory.

\subsection{Model configurations}
\label{app:model-configurations}
\input{tables/model_configurations}

\subsection{Additional Single-Cell Results}
\label{app:easy-medium-results}
\raggedbottom

\begin{table}[H]
\centering
\caption{Easy-task Wasserstein-2 distance ($\downarrow$) at held-out time points
(mean $\pm$ sample standard deviation over three seeds). The best and second-best means are shaded green (bold) and blue, respectively.}
\label{tab:wasserstein-easy}
\small
\renewcommand{\arraystretch}{0.92}

\begin{tabular*}{\textwidth}{@{\extracolsep{\fill}}*{4}{c}}
\toprule
\textbf{Method} & \multicolumn{3}{c}{\textbf{ZB/easy task}} \\
\cmidrule(lr){2-4}
& \multicolumn{3}{c}{\textbf{Interpolation}} \\
\cmidrule(lr){2-4}
& $\boldsymbol{t=5}$ & $\boldsymbol{t=7}$ & $\boldsymbol{t=9}$ \\
\midrule
scNODE & $435.03\pm7.52$ & $393.17\pm3.84$ & $473.80\pm4.00$ \\
LGP-OT & $436.27\pm3.20$ & \secondresult{362.05\pm2.20} & \secondresult{432.68\pm1.19} \\
MIOFlow & $447.71\pm5.63$ & $415.37\pm5.93$ & $491.57\pm5.63$ \\
PRESCIENT & $475.52\pm8.25$ & $447.97\pm6.87$ & $507.70\pm2.65$ \\
PI-SDE & \bestresult{422.71\pm1.60} & $369.90\pm2.79$ & $435.65\pm3.68$ \\
VGFM & $443.05\pm5.89$ & $391.59\pm3.52$ & $490.49\pm4.32$ \\
KoopCell & $431.53\pm6.02$ & $366.01\pm1.99$ & $436.71\pm1.02$ \\
KoopCell-M & \secondresult{428.05\pm3.79} & \bestresult{355.84\pm2.31} & \bestresult{426.25\pm1.34} \\
\bottomrule
\end{tabular*}

\vspace{0.5em}

\begin{tabular*}{\textwidth}{@{\extracolsep{\fill}}*{4}{c}}
\toprule
\textbf{Method} & \multicolumn{3}{c}{\textbf{DR/easy task}} \\
\cmidrule(lr){2-4}
& \multicolumn{3}{c}{\textbf{Interpolation}} \\
\cmidrule(lr){2-4}
& $\boldsymbol{t=5}$ & $\boldsymbol{t=7}$ & $\boldsymbol{t=9}$ \\
\midrule
scNODE & $348.05\pm3.93$ & $410.98\pm3.28$ & $499.14\pm5.93$ \\
LGP-OT & $346.08\pm0.54$ & $400.04\pm0.63$ & $471.29\pm2.28$ \\
MIOFlow & $345.69\pm0.55$ & $404.59\pm1.80$ & $497.52\pm2.97$ \\
PRESCIENT & $365.36\pm0.65$ & $428.44\pm0.67$ & $500.78\pm0.58$ \\
PI-SDE & \bestresult{335.03\pm1.02} & \secondresult{392.45\pm0.72} & $474.66\pm1.54$ \\
VGFM & $343.37\pm3.20$ & $406.06\pm2.83$ & $500.79\pm1.07$ \\
KoopCell & \secondresult{339.74\pm2.81} & $396.39\pm6.49$ & \secondresult{464.00\pm2.58} \\
KoopCell-M & $340.94\pm0.34$ & \bestresult{389.83\pm1.27} & \bestresult{462.52\pm1.06} \\
\bottomrule
\end{tabular*}

\vspace{0.5em}

\begin{tabular*}{\textwidth}{@{\extracolsep{\fill}}*{4}{c}}
\toprule
\textbf{Method} & \multicolumn{3}{c}{\textbf{SC/easy task}} \\
\cmidrule(lr){2-4}
& \multicolumn{3}{c}{\textbf{Interpolation}} \\
\cmidrule(lr){2-4}
& $\boldsymbol{t=6}$ & $\boldsymbol{t=11}$ & $\boldsymbol{t=16}$ \\
\midrule
scNODE & $57.42\pm0.96$ & $134.96\pm5.05$ & $118.18\pm3.13$ \\
LGP-OT & $51.16\pm0.31$ & $124.83\pm0.71$ & $100.55\pm1.35$ \\
MIOFlow & $55.96\pm0.73$ & $137.79\pm0.78$ & $122.78\pm1.98$ \\
PRESCIENT & $68.12\pm0.67$ & $144.99\pm0.61$ & $122.43\pm1.48$ \\
PI-SDE & \secondresult{50.83\pm0.18} & $128.23\pm1.44$ & $101.86\pm1.09$ \\
VGFM & $51.94\pm2.60$ & $152.69\pm3.70$ & $160.06\pm1.12$ \\
KoopCell & $52.40\pm1.09$ & \bestresult{124.38\pm1.09} & \secondresult{96.33\pm0.51} \\
KoopCell-M & \bestresult{50.52\pm0.70} & \secondresult{124.65\pm0.93} & \bestresult{95.54\pm0.70} \\
\bottomrule
\end{tabular*}
\end{table}

\begin{table}[H]
\centering
\caption{Medium-task Wasserstein-2 distance ($\downarrow$) at held-out time
points (mean $\pm$ sample standard deviation over three seeds). The best and second-best means are shaded green (bold) and blue, respectively.}
\label{tab:wasserstein-medium}
\small
\renewcommand{\arraystretch}{0.92}

\begin{tabular*}{\textwidth}{@{\extracolsep{\fill}}*{3}{c}}
\toprule
\textbf{Method} & \multicolumn{2}{c}{\textbf{ZB/medium task}} \\
\cmidrule(lr){2-3}
& \multicolumn{2}{c}{\textbf{Extrapolation}} \\
\cmidrule(lr){2-3}
& $\boldsymbol{t=11}$ & $\boldsymbol{t=12}$ \\
\midrule
scNODE & $611.92\pm5.22$ & $692.19\pm5.75$ \\
LGP-OT & $570.90\pm5.15$ & $642.06\pm9.06$ \\
MIOFlow & $630.14\pm13.26$ & $755.77\pm18.34$ \\
PRESCIENT & $629.81\pm3.64$ & $677.10\pm3.37$ \\
PI-SDE & $572.99\pm1.83$ & $642.34\pm4.18$ \\
VGFM & $607.60\pm4.59$ & $665.91\pm12.69$ \\
KoopCell & \secondresult{558.09\pm2.51} & \secondresult{605.61\pm3.34} \\
KoopCell-M & \bestresult{548.66\pm2.12} & \bestresult{603.10\pm5.57} \\
\bottomrule
\end{tabular*}

\vspace{0.5em}

\begin{tabular*}{\textwidth}{@{\extracolsep{\fill}}*{4}{c}}
\toprule
\textbf{Method} & \multicolumn{3}{c}{\textbf{DR/medium task}} \\
\cmidrule(lr){2-4}
& \multicolumn{3}{c}{\textbf{Extrapolation}} \\
\cmidrule(lr){2-4}
& $\boldsymbol{t=9}$ & $\boldsymbol{t=10}$ & $\boldsymbol{t=11}$ \\
\midrule
scNODE & $561.97\pm16.65$ & $573.46\pm13.14$ & $715.56\pm24.23$ \\
LGP-OT & $514.14\pm1.59$ & $520.71\pm15.78$ & $661.35\pm4.50$ \\
MIOFlow & $558.93\pm5.41$ & $614.22\pm21.61$ & $803.22\pm62.71$ \\
PRESCIENT & $551.35\pm0.75$ & \secondresult{511.82\pm0.22} & $669.97\pm1.78$ \\
PI-SDE & $525.17\pm1.02$ & $543.32\pm2.89$ & $708.72\pm2.27$ \\
VGFM & $554.89\pm5.43$ & $670.18\pm41.27$ & $901.81\pm72.49$ \\
KoopCell & \secondresult{505.50\pm4.18} & $521.45\pm12.38$ & \secondresult{644.53\pm10.65} \\
KoopCell-M & \bestresult{495.08\pm1.65} & \bestresult{496.61\pm3.11} & \bestresult{643.43\pm5.02} \\
\bottomrule
\end{tabular*}

\vspace{0.5em}

\begin{tabular*}{\textwidth}{@{\extracolsep{\fill}}*{4}{c}}
\toprule
\textbf{Method} & \multicolumn{3}{c}{\textbf{SC/medium task}} \\
\cmidrule(lr){2-4}
& \multicolumn{3}{c}{\textbf{Extrapolation}} \\
\cmidrule(lr){2-4}
& $\boldsymbol{t=17}$ & $\boldsymbol{t=18}$ & $\boldsymbol{t=19}$ \\
\midrule
scNODE & $130.31\pm3.95$ & $124.75\pm2.33$ & $130.34\pm2.47$ \\
LGP-OT & $115.15\pm1.66$ & $122.09\pm3.98$ & $126.47\pm3.44$ \\
MIOFlow & $149.96\pm14.49$ & $154.17\pm23.86$ & $173.96\pm39.70$ \\
PRESCIENT & $145.16\pm2.48$ & $135.73\pm2.14$ & $136.07\pm2.06$ \\
PI-SDE & $122.47\pm5.89$ & $115.87\pm3.50$ & $126.16\pm4.15$ \\
VGFM & $164.07\pm1.59$ & $168.12\pm1.03$ & $174.05\pm1.34$ \\
KoopCell & \bestresult{110.59\pm0.65} & \secondresult{111.27\pm5.20} & \secondresult{115.10\pm5.88} \\
KoopCell-M & \secondresult{114.75\pm0.39} & \bestresult{109.09\pm2.17} & \bestresult{113.48\pm2.22} \\
\bottomrule
\end{tabular*}
\end{table}

\subsubsection{Predicted cell populations at held-out time points}
\label{app:single-cell-prediction-visualizations}

\begin{figure}[p]
  \centering
  \includegraphics[width=\textwidth]{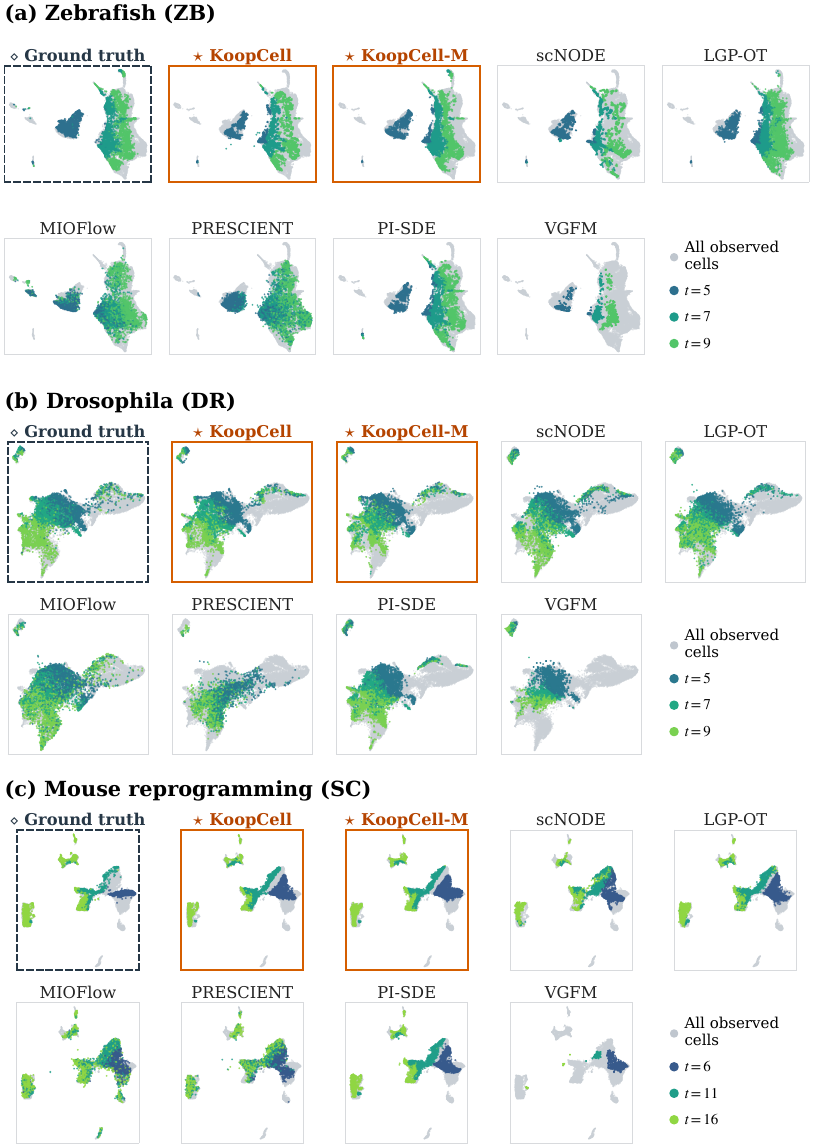}
  \caption{Easy-task predictions for ZB, DR, and SC. Colors indicate held-out
  time points; gray points show all observed cells. Diamonds and dark dashed
  frames identify ground truth; stars and orange frames mark KoopCell and KoopCell-M. All methods within each dataset share the
  same projection and axis ranges.}
  \label{fig:easy-populations}
\end{figure}

\begin{figure}[p]
  \centering
  \includegraphics[width=\textwidth]{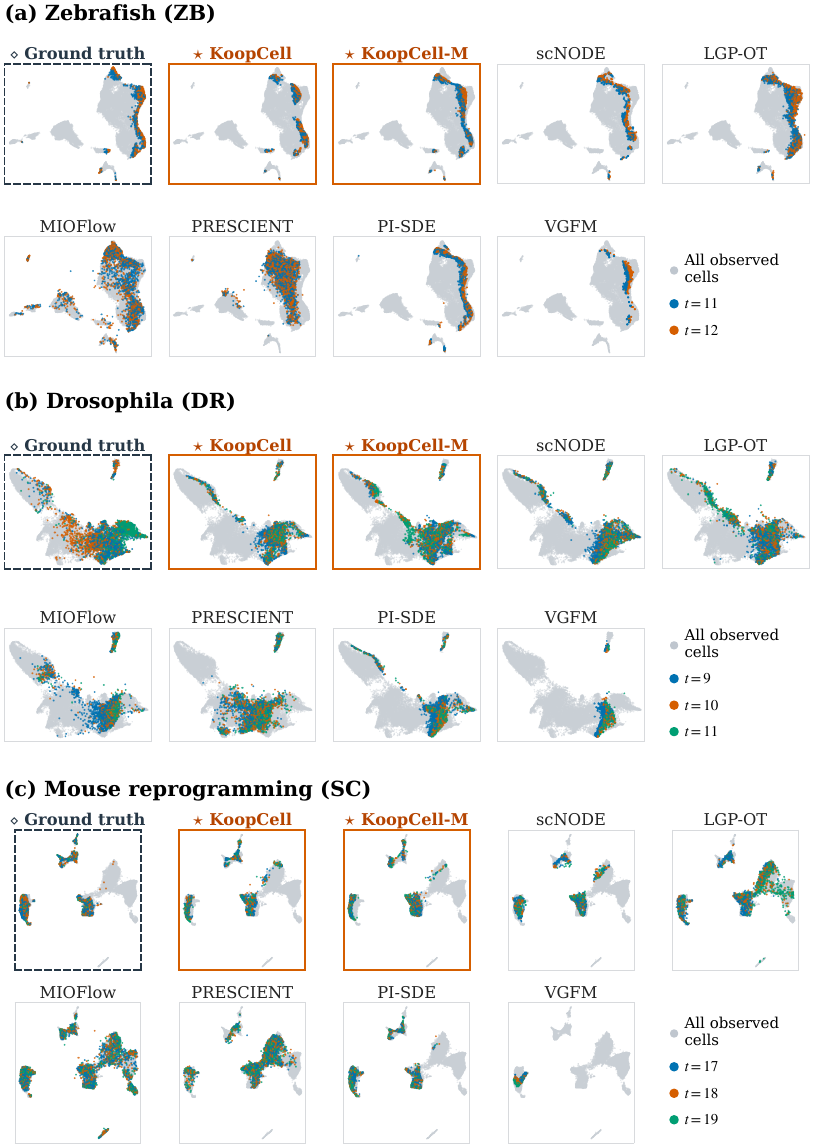}
  \caption{Medium-task predictions for ZB, DR, and SC. Colors indicate held-out
  time points; gray points show all observed cells. Diamonds and dark dashed
  frames identify ground truth; stars and orange frames mark KoopCell and KoopCell-M. All methods within each dataset share the
  same projection and axis ranges.}
  \label{fig:medium-populations}
\end{figure}

\begin{figure}[p]
  \centering
  \includegraphics[width=\textwidth]{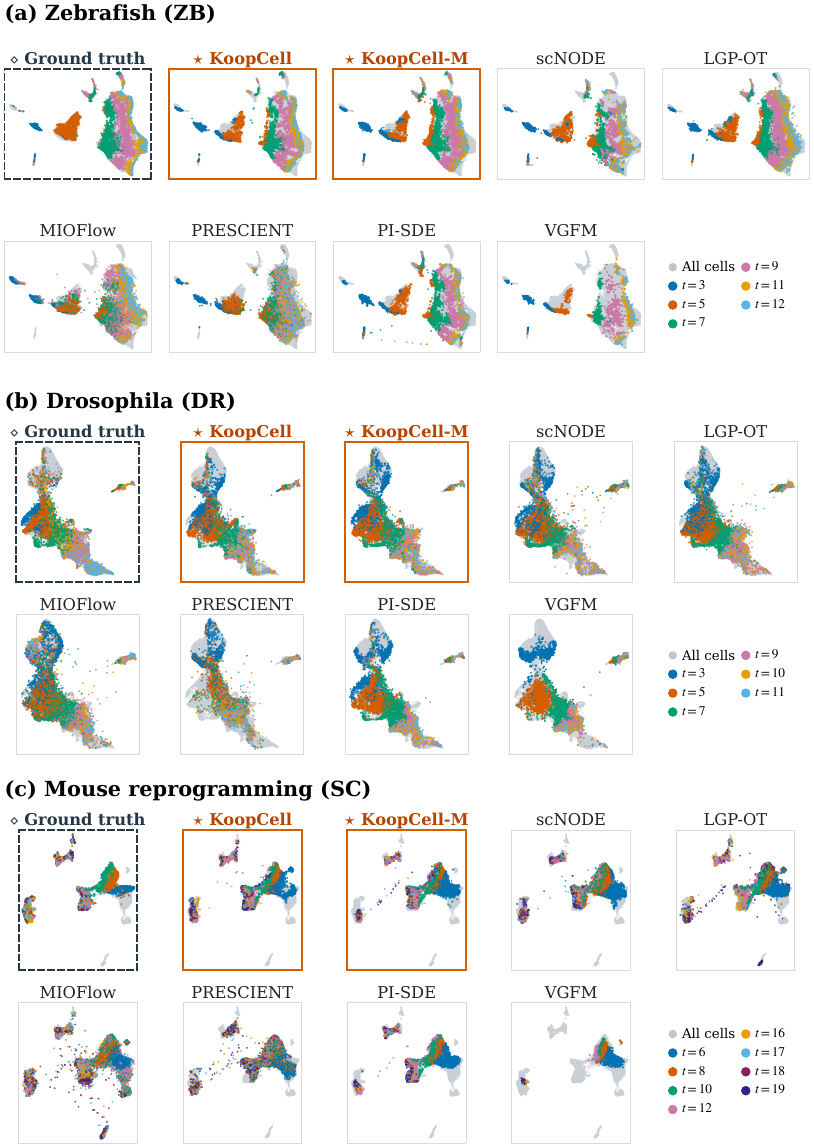}
  \caption{Hard-task predictions for ZB, DR, and SC, with all held-out
  interpolation and extrapolation time points overlaid in each panel.
  Colors indicate time points; gray points show all observed cells.
  Diamonds and dark dashed frames identify ground truth; stars and orange
  frames mark KoopCell and KoopCell-M. All methods within each dataset
  share the UMAP embedding used in Figure~\ref{fig:single-cell-datasets}
  and the same axis ranges.}
  \label{fig:hard-populations}
\end{figure}

Figures~\ref{fig:easy-populations}--\ref{fig:hard-populations}
compare the predicted populations from individual training runs.
All methods share the saved PCA--UMAP projection for each dataset and task.
We display up to 2,000 cells per time point, resampling VGFM particles
according to their normalized growth weights.

\clearpage

\subsubsection{Visualization of the learned latent representations}
\label{app:latent-representations}

\begin{figure}[!htbp]
  \centering
  \includegraphics[width=\textwidth]{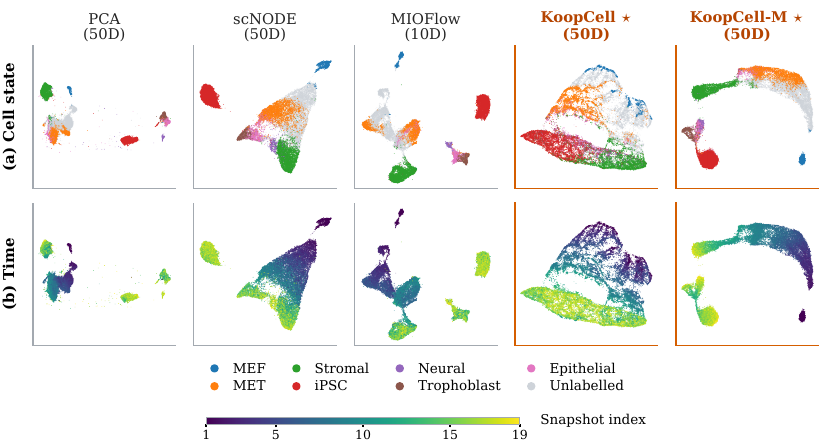}
  \caption{Latent representations of SC cells under the hard-task setting, colored by
  (a) reference cell state and (b) snapshot time index. All five methods show
  the same observed cells. Each method uses its own UMAP coordinates,
  shared between the two rows. The KoopCell and KoopCell-M columns are
  also shown in Figure~\ref{fig:latent-representations}.}
  \label{fig:latent-representations-full}
\end{figure}

Figure~\ref{fig:latent-representations-full} compares the representations of
all 23,619 observed SC cells under the hard-task setting. Each method has
its own UMAP projection; both rows use the same coordinates for that
method. Reference states are obtained by matching cell identifiers to
the published Waddington-OT cell sets and labeling the initial population
as MEF. Unmatched or ambiguous cells remain gray; classifier predictions
are not used to color these panels. 

\subsection{Biological Downstream Analysis}
\label{app:biological-downstream}

We investigate how the population dynamics learned by KoopCell-M relate
to lineage-associated genes and developmental expression programs.
Using KoopCell-M trained on all ZB snapshots, we focus on Presomitic
Mesoderm (PSM) and Hindbrain development, following \citet{zhang2024scnode}.

\paragraph{Lineage-associated gene identification.}
We first estimate the probability that each cell contributes to a terminal
cell type at 12 hpf. For a cell observed at time $t_i$, we encode
$\bm{z}_i=\bm{m}_{\bm{\theta}}(\bm{x}_i,t_i)$ and infer its path probabilities and conditional
memory distribution from the generative model.
We then construct a directed graph between successive snapshots by propagating
augmented states $(\bm{Z},\bm{H})$ with the Gaussian transition in
\Eqref{eq:memory-transition} and assigning the propagated mass to neighboring
states. Distances are scaled by the corresponding model covariance, and
the path identity is retained across transitions.
Reference annotations of all terminal cell types define the endpoints.
Let $\bm{P}_{\ell,r}$ denote the transition matrix from snapshot $\ell$ to
$\ell+1$ along path $r$, and let $\bm{F}_{\ell,r}(j,L)$ denote the probability
of reaching terminal cell type $L$ from augmented state $j$ at snapshot
$\ell$, conditional on path $r$. The probability matrices satisfy
\begin{equation}
  \bm{F}_{\ell,r}=\bm{P}_{\ell,r}\bm{F}_{\ell+1,r},
  \qquad \bm{F}_{K,r}(j,L)=\mathbf{1}(c_j=L),
  \label{eq:biology-fates}
\end{equation}
where $K$ indexes the terminal snapshot and $c_j$ is the reference label
of terminal cell $j$. Averaging over the inferred memory and path
distributions gives the cell-level fate probability $f_{iL}$.

Following the lineage-driver statistic of CellRank~\citep{weiler2024cellrank},
we rank genes by their Pearson correlation with the predicted probability
of reaching lineage $L$:
\begin{equation}
  r_{gL}=\operatorname{Corr}_{i:t_i<T}(x_{ig},f_{iL}),
  \label{eq:biology-drivers}
\end{equation}
where $x_{ig}$ is observed log-expression and $T$ denotes the terminal
time. To examine whether these fate-associated genes also distinguish
the observed terminal populations, we compare log-normalized expression
between 540 PSM and 511 Hindbrain cells at 12 hpf.
Table~\ref{tab:biological-markers} reports the mean expression difference
$\Delta_g$ (PSM minus Hindbrain) for the four highest-ranked genes per
lineage, selected by $r_{gL}$. Positive and negative values indicate higher
expression in PSM and Hindbrain, respectively.

\begin{figure}[!htbp]
  \centering
  \includegraphics[width=\textwidth]{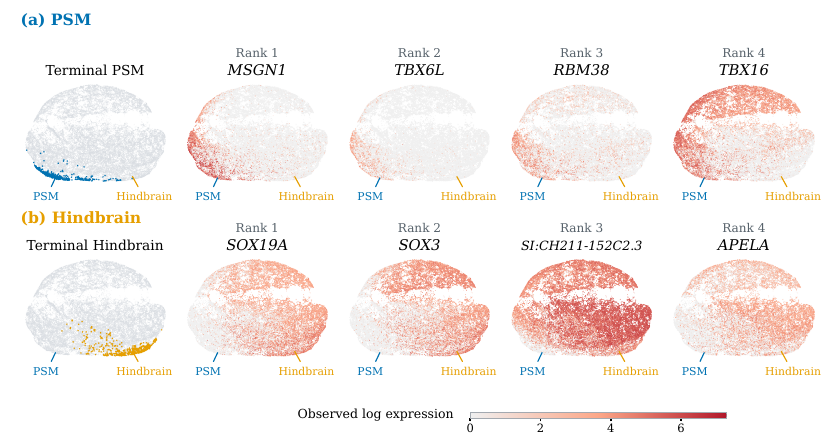}
  \caption{Observed expression of lineage-associated genes on a UMAP
  projection of the latent representations learned by KoopCell-M for ZB cells.
  Rows show the four highest-ranked genes for
  (a) PSM and (b) Hindbrain. The left panels highlight the corresponding
  annotated populations at 12 hpf; the remaining panels show observed
  log-expression across all cells using a common color scale.
  All panels share the same UMAP coordinates. Arrows indicate the terminal populations.}
  \label{fig:biological-markers}
\end{figure}

\paragraph{Recovery of developmental markers.}
\textit{MSGN1} and \textit{TBX16} rank first and fourth for PSM, with
correlations of $0.665$ and $0.500$, respectively
(Table~\ref{tab:biological-markers}). Both genes participate in zebrafish
mesodermal morphogenesis and differentiation~\citep{manning2015tbx16}.
For Hindbrain, \textit{SOX19A} and \textit{SOX3} rank first and second
($r=0.568$ and $0.540$), consistent with their expression in the developing
neuroectoderm~\citep{rossi2009rohon}.
All eight genes show higher terminal expression in their corresponding lineage.
\textit{APELA} shows a modest terminal expression difference
($\Delta_g=-0.243$).
In the UMAP projection, the observed expression patterns of these genes
align with the annotated PSM and Hindbrain populations
(Figure~\ref{fig:biological-markers}).

\begin{table}[!htbp]
  \centering
  \small
  \caption{Lineage-associated genes and terminal expression differences.
  The four highest-ranked genes are shown for each lineage.
  $r_{gL}$ is the preterminal expression--fate correlation used for ranking;
  $\Delta_g$ is mean observed log-expression in PSM minus Hindbrain at
  12 hpf.}
  \label{tab:biological-markers}
  \begin{tabular*}{\textwidth}{@{\extracolsep{\fill}}llrrr}
    \toprule
    Lineage & Gene & Rank & $r_{gL}$ & $\Delta_g$ \\
    \midrule
    PSM & \textit{MSGN1} & 1 & 0.665 & $4.388$ \\
    PSM & \textit{TBX6L} & 2 & 0.559 & $1.434$ \\
    PSM & \textit{RBM38} & 3 & 0.513 & $2.160$ \\
    PSM & \textit{TBX16} & 4 & 0.500 & $3.844$ \\
    \midrule
    Hindbrain & \textit{SOX19A} & 1 & 0.568 & $-3.670$ \\
    Hindbrain & \textit{SOX3} & 2 & 0.540 & $-3.552$ \\
    Hindbrain & \textit{SI:CH211-152C2.3} & 3 & 0.472 & $-3.770$ \\
    Hindbrain & \textit{APELA} & 4 & 0.371 & $-0.243$ \\
    \bottomrule
  \end{tabular*}
\end{table}

\paragraph{Dynamic gene-expression programs.}
We next identify groups of genes with similar predicted expression-rate
profiles along developmental trajectories. Starting from cells observed at 8 hpf, we infer their path
and memory distributions and propagate the augmented dynamics to 12 hpf.
The decoder maps each trajectory to log-normalized gene-expression space. Because noise
enters only the memory variables, the predicted expression rate for gene
$g$ follows the chain rule:
\begin{equation}
  u_g(t)=\nabla_{\bm{z}}\operatorname{Dec}_{\bm{\phi},g}(\bm{Z}_t)^\top
                (\bm{A}\bm{Z}_t+\bm{B}\bm{H}_t+\bm{b}_z).
  \label{eq:biology-expression-rate}
\end{equation}
We compute this rate using Jacobian--vector products and average it
separately for each lineage. We first give each source cell equal total
weight, distributed across paths according to their inferred probabilities
and equally among trajectory samples along each path. Let $\alpha_a$
denote the resulting weight of trajectory $a$, whose path is $r_a$.
To determine its contribution to lineage $L$, we compare its endpoint
$\bm{z}_T^a$ at $T=12$ hpf with annotated terminal cells. We define
$b_L(\bm{z}_T^a,r_a)$ as the distance-weighted fraction of neighboring cells
labeled $L$, with distances scaled by the terminal covariance for path $r_a$.
Normalizing the products $\alpha_a b_L(\bm{z}_T^a,r_a)$ gives the lineage-specific
weights and mean expression rates:
\begin{equation}
  w_{aL}=\frac{\alpha_a b_L(\bm{z}_T^a,r_a)}
                   {\sum_{a'}\alpha_{a'}b_L(\bm{z}_T^{a'},r_{a'})},
  \qquad
  \bar u_{gL}(t)=\sum_a w_{aL}u_g^a(t).
  \label{eq:biology-rate-average}
\end{equation}

For each lineage, we select the 100 genes with the largest mean expression
rates over 8--12 hpf, emphasizing net upregulation over this interval.
We standardize each rate profile across time and
apply Ward clustering to obtain three modules, ordered by the temporal
center of their above-average rates.
To identify the biological functions associated with these programs,
we test enrichment of Gene Ontology (GO) biological-process terms~\citep{go2026knowledgebase}
in each lineage's top-100 genes and in each of its three modules.
We map input gene symbols to Zebrafish Information Network (ZFIN)
identifiers using official symbols and aliases~\citep{ruzicka2019zfin}.
The background comprises the 1,549 uniquely mapped input genes with
GO biological-process annotations; each tested gene set is restricted to
this same background. A one-sided hypergeometric test measures whether
a term annotates more genes in a tested set than expected from its frequency
in the background~\citep{boyle2004termfinder}.
Benjamini--Hochberg correction~\citep{benjamini1995controlling} is applied
jointly across the tested terms and gene sets in both lineages, with
$q<0.05$ as the significance threshold.

\begin{figure}[!htbp]
  \centering
  \includegraphics[width=\textwidth]{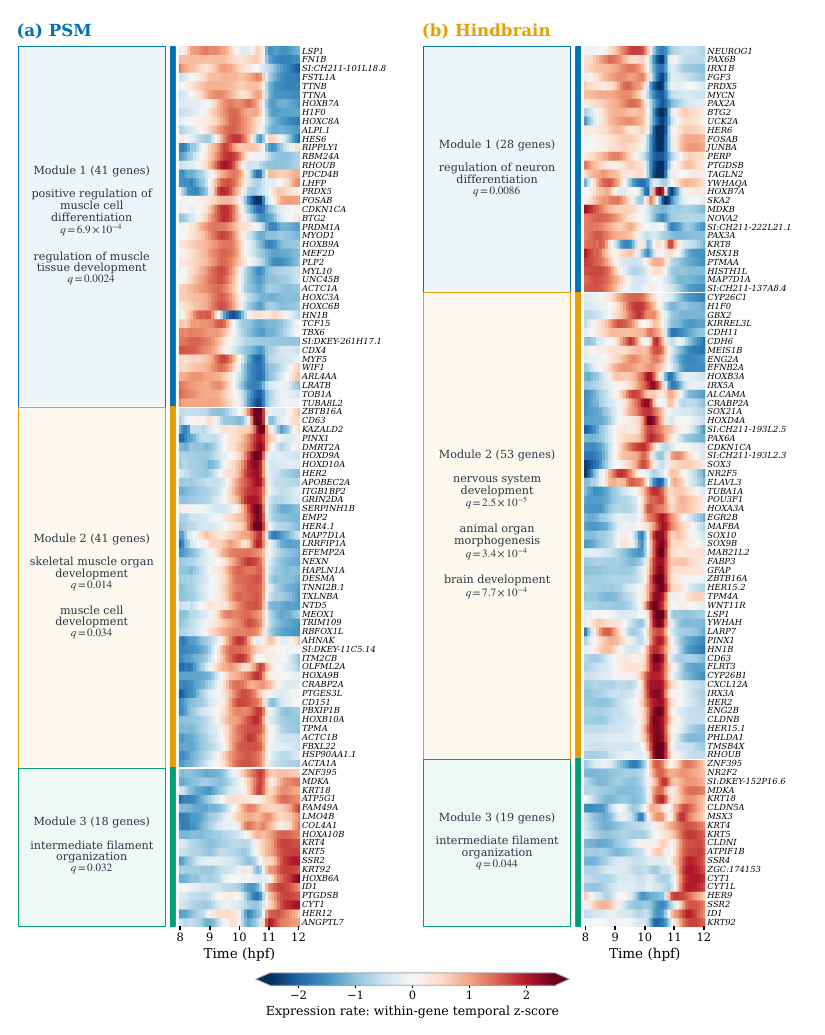}
  \caption{Dynamic gene-expression programs along KoopCell-M trajectories
  over 8--12 hpf. Each lineage contains the 100 genes with the largest mean
  predicted rates of change in log-expression, grouped into three Ward modules.
  Colors show within-gene temporal z-scores of expression rates, with red
  indicating rates above the gene's temporal mean and blue indicating
  rates below it. Boxes report representative GO biological-process
  terms and adjusted $q$-values. PSM programs are enriched for muscle
  development and Hindbrain programs for neural development; both
  include a module containing keratin genes.}
  \label{fig:biological-dynamic-modules}
\end{figure}

\paragraph{Developmental functions of the expression programs.}
The PSM modules are enriched for genes involved in muscle differentiation and development
(Figure~\ref{fig:biological-dynamic-modules}).
The first module contains \textit{MYF5}, \textit{MYOD1},
\textit{PRDM1A}, and \textit{RBM24A}, and is enriched for positive
regulation of muscle cell differentiation ($q=6.91\times10^{-4}$).
This is consistent with the role of \textit{MYF5} and \textit{MYOD1}
in initiating zebrafish myogenesis~\citep{osborn2020myogenesis}.
The second includes \textit{ACTA1A}, \textit{DESMA}, and \textit{DMRT2A},
with enrichment for skeletal muscle organ development
($q=1.43\times10^{-2}$).
In Hindbrain, the first module includes \textit{FGF3}, \textit{HER6},
and \textit{NEUROG1}, with enrichment for regulation of neuron
differentiation ($q=8.57\times10^{-3}$).
The second contains \textit{SOX3}, \textit{SOX21A}, \textit{PAX6A},
and \textit{GBX2}, and is enriched for nervous system development
($q=2.47\times10^{-5}$) and brain development ($q=7.68\times10^{-4}$).
These enrichments are consistent with the neural progenitor and differentiation
programs described in the developing zebrafish hindbrain~\citep{tambalo2020hindbrain}.
The third module in each lineage is enriched for intermediate
filament organization and includes the keratin genes
\textit{KRT4} and \textit{KRT5}~\citep{cokus2019skin}.
Together, the recovered markers and lineage-associated functional enrichments
support the biological interpretability of the learned population dynamics.

\clearpage

\end{document}

%% file: math_commands.tex
\usepackage{amsmath,amsfonts,bm}

\def\eqref#1{equation~\textup{(\ref{#1})}}
\def\Eqref#1{Equation~\textup{(\ref{#1})}}

\def\1{\bm{1}}

\DeclareMathAlphabet{\mathsfit}{\encodingdefault}{\sfdefault}{m}{sl}
\SetMathAlphabet{\mathsfit}{bold}{\encodingdefault}{\sfdefault}{bx}{n}



%% file: figures/branching/table_main.tex
\begin{tabular*}{\linewidth}{@{\extracolsep{\fill}}lcc@{}}
\toprule
Model & SWD $\downarrow$ & Branch $W_1\downarrow$ \\
\midrule
KoopCell & $0.1557\pm0.0146$ & $0.2245\pm0.0069$ \\
KoopCell-M & $\mathbf{0.1124\pm0.0046}$ & $\mathbf{0.0866\pm0.0026}$ \\
\bottomrule
\end{tabular*}

%% file: figures/branching/table_detail.tex
\begin{tabular}{ccccccc}
\toprule
Times & Model & SWD & Energy $\times10^3$ & MMD$^2\!\times10^3$ & Branch $W_1$ & Mass error \\
\midrule
\multirow[c]{2}{*}{Training times} & KoopCell & $0.0882\pm0.0047$ & $2.281\pm0.561$ & $0.831\pm0.225$ & $0.068\pm0.003$ & $0.0264\pm0.0007$ \\
 & KoopCell-M & $\mathbf{0.0798\pm0.0004}$ & $\mathbf{1.273\pm0.022}$ & $\mathbf{0.410\pm0.011}$ & $\mathbf{0.044\pm0.002}$ & $\mathbf{0.0122\pm0.0017}$ \\
\midrule
\multirow[c]{2}{*}{Interpolation} & KoopCell & $0.0994\pm0.0083$ & $2.804\pm0.960$ & $1.058\pm0.383$ & $0.105\pm0.004$ & $0.0519\pm0.0006$ \\
 & KoopCell-M & $\mathbf{0.0848\pm0.0003}$ & $\mathbf{1.258\pm0.055}$ & $\mathbf{0.437\pm0.018}$ & $\mathbf{0.067\pm0.003}$ & $\mathbf{0.0201\pm0.0013}$ \\
\midrule
\multirow[c]{2}{*}{Extrapolation} & KoopCell & $0.2120\pm0.0212$ & $10.790\pm2.713$ & $4.359\pm0.952$ & $0.344\pm0.010$ & $0.0951\pm0.0049$ \\
 & KoopCell-M & $\mathbf{0.1400\pm0.0089}$ & $\mathbf{2.393\pm0.568}$ & $\mathbf{1.012\pm0.267}$ & $\mathbf{0.106\pm0.006}$ & $\mathbf{0.0016\pm0.0002}$ \\
\bottomrule
\end{tabular}

%% file: tables/model_configurations.tex
\begingroup
\setlength{\intextsep}{5pt}
\setlength{\abovecaptionskip}{4pt}
\setlength{\belowcaptionskip}{3pt}
\begin{table}[H]
  \centering
  \caption{Settings shared across the three datasets and prediction tasks.
  Stage 1 and Stage 2 denote VAE pretraining and dynamics/representation alternating training stages, respectively.}
  \label{tab:model-common-configurations}
  \small
  \setlength{\tabcolsep}{5pt}
  \renewcommand{\arraystretch}{1.05}
  \begin{tabular*}{\textwidth}{@{\extracolsep{\fill}}lcc@{}}
    \toprule
    \textbf{Parameter} & \textbf{KoopCell} & \textbf{KoopCell-M} \\
    \midrule
    Visible latent dimension $D$ & 50 & 50 \\
    Encoder/decoder MLP activation & ReLU & ReLU \\
    Training epochs (Stage 1 / Stage 2) & 500 / 4,000 & 500 / 4,000 \\
    Batch size (Stage 1 / Stage 2) & 256 / 128 & 256 / 128 \\
    Variational coefficient $\beta$ & 0.01 & 0.01 \\
    Number of Fourier test functions & 2,048 & 2,048 \\
    Sinkhorn batch size & 512 & 256 \\
    State-classifier hidden width & -- & 128 \\
    Path-loss coefficient $\lambda_R$ & -- & 0.1 \\
    \bottomrule
  \end{tabular*}
\end{table}

\begin{table}[H]
  \centering
  \caption{Selected MLP architectures and loss weights.
  Encoder and decoder entries list hidden-layer widths;
  $d_h$ is the memory dimension.}
  \label{tab:model-task-configurations}
  \small
  \setlength{\tabcolsep}{2.5pt}
  \renewcommand{\arraystretch}{1.05}
  \begin{tabular*}{\textwidth}{@{\extracolsep{\fill}}llcccccc@{}}
    \toprule
    \rowcolor{algpretrain!10}
    \multicolumn{8}{l}{\textbf{KoopCell}} \\
    \textbf{Task} & \textbf{Data} & \textbf{Encoder MLP} & \textbf{Decoder MLP}
      & $d_h$ & $\lambda_z$ & $\lambda_x$ & $\lambda_{\mathrm{weak}}$ \\
    \midrule
    \multirow{3}{*}{Easy} & ZB & 256--256--256 & 256--256--256 & -- & 1 & 1.5 & 1 \\
      & DR & 512--256--256 & 256--256--512 & -- & 1 & 5 & 1 \\
      & SC & 512--256--256 & 256--256--512 & -- & 1 & 3 & 1 \\
    \cmidrule(lr){1-8}
    \multirow{3}{*}{Medium} & ZB & 256--256--256 & 256--256--256 & -- & 1 & 1.5 & 1 \\
      & DR & 256--256--256 & 256--256--256 & -- & 1 & 5 & 1 \\
      & SC & 256--256--256 & 256--256--256 & -- & 1 & 3 & 1 \\
    \cmidrule(lr){1-8}
    \multirow{3}{*}{Hard} & ZB & 256--256--256 & 256--256--256 & -- & 1 & 1.5 & 1 \\
      & DR & 256--256--256 & 256--256--256 & -- & 1 & 5 & 1 \\
      & SC & 256--256--256 & 256--256--256 & -- & 1 & 3 & 1 \\
    \midrule
    \rowcolor{algalternating!10}
    \multicolumn{8}{l}{\textbf{KoopCell-M}} \\
    \textbf{Task} & \textbf{Data} & \textbf{Encoder MLP} & \textbf{Decoder MLP}
      & $d_h$ & $\lambda_z$ & $\lambda_x$ & $\lambda_{\mathrm{weak}}$ \\
    \midrule
    \multirow{3}{*}{Easy} & ZB & 512--256--256 & 256--256--512 & 16 & 5 & 10 & $10^5$ \\
      & DR & 512--256--256 & 256--256--512 & 32 & 2 & 10 & $10^4$ \\
      & SC & 512--256--256 & 256--256--512 & 16 & 10 & 10 & 7,500 \\
    \cmidrule(lr){1-8}
    \multirow{3}{*}{Medium} & ZB & 512--256--256 & 256--256--512 & 32 & 5 & 10 & $10^5$ \\
      & DR & 512--256--256 & 256--256--512 & 50 & 5 & 10 & $10^4$ \\
      & SC & 512--256--256 & 256--256--512 & 32 & 5 & 5 & 7,500 \\
    \cmidrule(lr){1-8}
    \multirow{3}{*}{Hard} & ZB & 256--256--256 & 256--256--256 & 16 & 5 & 10 & $10^5$ \\
      & DR & 256--256--256 & 256--256--256 & 32 & 2 & 10 & $10^4$ \\
      & SC & 256--256--256 & 256--256--256 & 16 & 1 & 3 & 7,500 \\
    \bottomrule
  \end{tabular*}
\end{table}
\endgroup